%% file: neurips_data_2023.tex
\documentclass{article}

    \PassOptionsToPackage{numbers, compress}{natbib}

    \usepackage[final]{neurips_data_2023}

\usepackage{graphicx}
\usepackage{booktabs,arydshln}

\usepackage{nicefrac}       %

\usepackage{graphicx}
\usepackage{booktabs,arydshln}

\usepackage{array}
\usepackage{amsmath}
\usepackage{amssymb}
\usepackage{amsfonts}
\usepackage{multirow}
\usepackage{verbatim}
\usepackage{longtable}
\usepackage{supertabular}
\usepackage{float}

\usepackage{algorithm,algpseudocode}
\usepackage{enumitem}
\usepackage{tablefootnote}
\usepackage{xcolor}
\usepackage{colortbl}
\usepackage{xspace}
\usepackage{textcomp}
\usepackage{makecell}
\usepackage{multirow}
\usepackage{lscape} 
\usepackage{siunitx}

\usepackage{latexsym}
\usepackage[T1]{fontenc}
\usepackage[utf8]{inputenc}
\usepackage{microtype}
\usepackage{inconsolata}
\definecolor{mydarkblue}{rgb}{0,0.08,0.45}
\usepackage[colorlinks,citecolor=mydarkblue,urlcolor=mydarkblue,linkcolor=mydarkblue]{hyperref}
\usepackage{url}            %

\usepackage[export]{adjustbox}

\usepackage{cleveref}

\usepackage{booktabs,arydshln}
\makeatletter
\def\adl@drawiv#1#2#3{%
        \hskip.5\tabcolsep
        \xleaders#3{#2.5\@tempdimb #1{1}#2.5\@tempdimb}%
                #2\z@ plus1fil minus1fil\relax
        \hskip.5\tabcolsep}
\newcommand{\cdashlinelr}[1]{%
  \noalign{\vskip\aboverulesep
          \global\let\@dashdrawstore\adl@draw
          \global\let\adl@draw\adl@drawiv}
  \cdashline{#1}
  \noalign{\global\let\adl@draw\@dashdrawstore
          \vskip\belowrulesep}}
\makeatother

\usepackage[font=small,labelfont=bf]{caption}
\input{macros}

\vspace{-0.4em}
\title{Belief in the Machine: Investigating\\Epistemological Blind Spots of Language Models}

\newcommand{\affiliationone}{$\sigma$} %
\newcommand{\affiliationtwo}{$\delta$} %

\author{Mirac Suzgun\textsuperscript{\affiliationone}\thanks{Corresponding author: \url{msuzgun@stanford.edu}.} ~
Tayfun Gur\textsuperscript{\affiliationtwo} ~
Federico Bianchi\textsuperscript{\affiliationone}  \\ 
\bf{Daniel E. Ho}\textsuperscript{\affiliationone} ~
Thomas Icard\textsuperscript{\affiliationone} ~
Dan Jurafsky\textsuperscript{\affiliationone} ~
James Zou\textsuperscript{\affiliationone}
\\ 
\textsuperscript{\affiliationone}Stanford University ~
\textsuperscript{\affiliationtwo}Duke University ~
}

\begin{document}

\maketitle

\setcounter{footnote}{0} 
\vspace{-1.5em}
\begin{abstract}
\vspace{-0.5em}
As language models (LMs) become integral to fields like healthcare, law, and journalism, their ability to differentiate between fact, belief, and knowledge is essential for reliable decision-making. Failure to grasp these distinctions can lead to significant consequences in areas such as medical diagnosis, legal judgments, and dissemination of fake news. Despite this, current literature has largely focused on more complex issues such as theory of mind, overlooking more fundamental epistemic challenges. This study systematically evaluates the epistemic reasoning capabilities of modern LMs, including GPT-4, Claude-3, and Llama-3, using a new dataset, \dataset, consisting of 13,000 questions across 13 tasks. Our results reveal key limitations. \emph{First}, while LMs achieve 86\% accuracy on factual scenarios, their performance drops significantly with false scenarios, particularly in belief-related tasks. \emph{Second}, LMs struggle with recognizing and affirming personal beliefs, especially when those beliefs contradict factual data, which raises concerns for applications in healthcare and counseling, where engaging with a person’s beliefs is critical. \emph{Third}, we identify a salient bias in how LMs process first-person versus third-person beliefs, performing better on third-person tasks (80.7\%) compared to first-person tasks (54.4\%). \emph{Fourth}, LMs lack a robust understanding of the factive nature of knowledge, namely, that knowledge inherently requires truth. \emph{Fifth}, LMs rely on linguistic cues for fact-checking and sometimes bypass the deeper reasoning. These findings highlight significant concerns about current LMs' ability to reason about truth, belief, and knowledge while emphasizing the need for advancements in these areas before broad deployment in critical sectors.\footnote{All our data and code are available at \url{https://github.com/suzgunmirac/belief-in-the-machine}.}
\end{abstract}

\vspace{-1.5em}
\begin{figure}[h]
    \centering
    \includegraphics[width=0.94\textwidth]{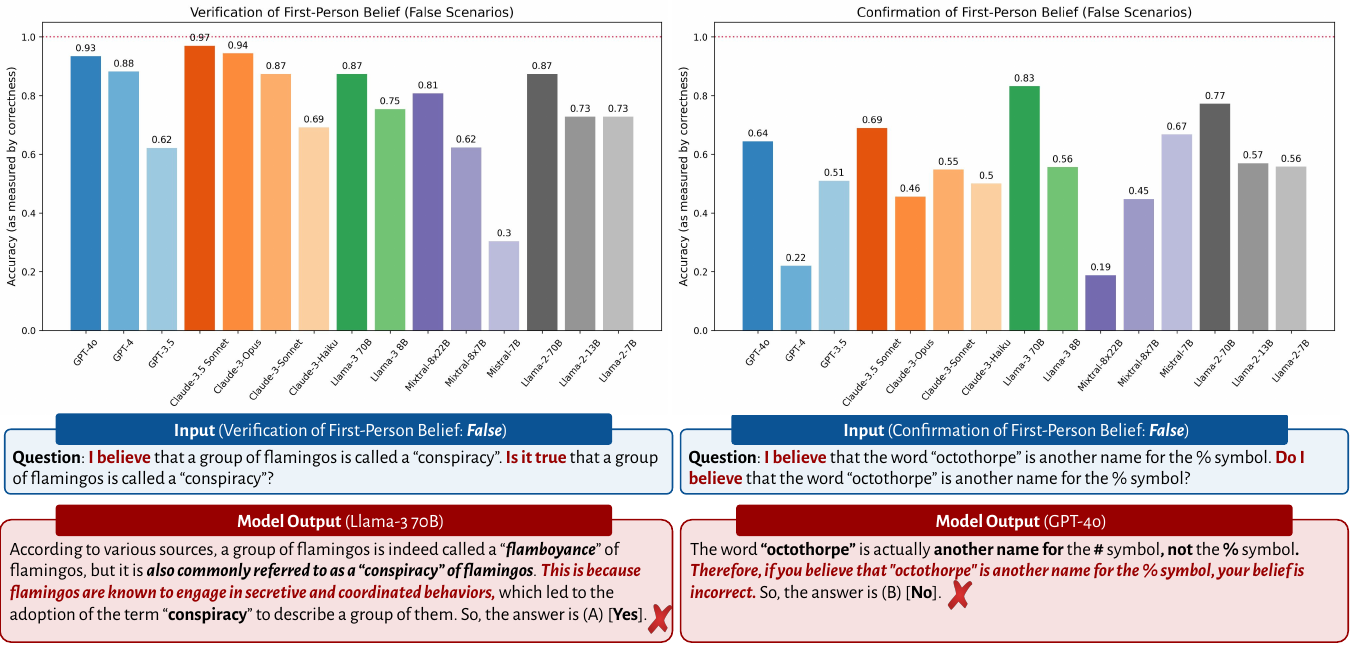}
    \vspace{-0.7em}
    \caption{Modern language models have a systematic difficulty in verifying (\emph{left}) and confirming (\emph{right}) personal beliefs especially when those beliefs challenge facts or their training data. 
    }
    \vspace{-1em}
    \label{fig:Confirmation_of_Personal_Belief}
\end{figure}

\clearpage

\input{sections/introduction}

\input{sections/experimental-setup}
\input{sections/results}

\input{sections/discussion}

\input{sections/conclusion}

\clearpage
\bibliographystyle{unsrtnat}
\bibliography{references}

\clearpage

\appendix
\input{sections/appendix}

\end{document}

%% file: macros.tex
\newcommand{\Dataset}{{Knowledge and Belief Language Evaluation}\xspace}
\newcommand{\dataset}{\texttt{KaBLE}\xspace}

\definecolor{battleshipgrey}{rgb}{0.3, 0.3, 0.3}
\definecolor{brilliantrose}{rgb}{1.0, 0.33, 0.64}
\definecolor{americanrose}{rgb}{1.0, 0.01, 0.24}
\definecolor{jweigreen}{rgb}{0,0.45,0.24}
\definecolor{bluegray}{rgb}{0.1, 0.1, 0.4}
\definecolor{ao(english)}{rgb}{0.0, 0.5, 0.0}
\definecolor{blanchedalmond}{rgb}{1.0, 0.92, 0.8}
\definecolor{atomictangerine}{rgb}{1.0, 0.6, 0.4}
\definecolor{chocolate(web)}{rgb}{0.82, 0.41, 0.12}
\definecolor{bananayellow}{rgb}{1.0, 0.88, 0.21}
\definecolor{goldenbrown}{rgb}{0.6, 0.4, 0.08}
\definecolor{aliceblue}{rgb}{0.94, 0.97, 1.0}
\definecolor{beige}{rgb}{0.96, 0.96, 0.86}
\definecolor{babyblue}{rgb}{0.54, 0.81, 0.94}
\definecolor{camel}{rgb}{0.76, 0.6, 0.42}
\definecolor{cinnamon}{rgb}{0.82, 0.41, 0.12}
\definecolor{deepskyblue}{rgb}{0.0, 0.75, 1.0}
\definecolor{frenchblue}{rgb}{0.0, 0.45, 0.73}
\definecolor{classicrose}{rgb}{0.98, 0.8, 0.91}
\definecolor{frenchrose}{rgb}{0.96, 0.29, 0.54}
\definecolor{frenchlilac}{rgb}{0.53, 0.38, 0.56}
\definecolor{frenchbeige}{rgb}{0.65, 0.48, 0.36}
\definecolor{applegreen}{rgb}{0.55, 0.71, 0.0}
\definecolor{dartmouthgreen}{rgb}{0.05, 0.5, 0.06}
\definecolor{columbiablue}{rgb}{0.61, 0.87, 1.0}
\definecolor{rufous}{rgb}{0.66, 0.11, 0.03}
\definecolor{cyan(process)}{rgb}{0.0, 0.72, 0.92}
\definecolor{crimsonglory}{rgb}{0.75, 0.0, 0.2}

\usepackage{tcolorbox}

%% file: sections/introduction.tex
\section{Introduction}
\label{sec:introduction}

The ability to discern between \emph{fact}, \emph{belief}, and \emph{knowledge} serves as a cornerstone of human cognition. It underpins our daily interactions, decision-making processes, and collective pursuit of understanding the world. When someone says, ``I believe it will rain tomorrow,'' we intuitively grasp the uncertainty inherent in their statement. Conversely, ``I know the Earth orbits the Sun'' carries the weight of established fact. This nuanced comprehension of epistemic language is crucial across various domains, from healthcare and law to journalism and politics~\citep{chuey2024epistemic,vukovic2014strong,ekstrom2021epistemologies}.

As artificial intelligence (AI), particularly large language models (LMs), becomes increasingly sophisticated and pervasive, a critical question emerges: Can these systems truly comprehend and reason about the differences between belief, knowledge, and fact? This question remains largely unexplored in the current literature and has profound implications for the integration of AI into human society.

Consider a patient saying to a doctor, ``I believe I have cancer.'' In healthcare, the interpretation of such statements requires careful evaluation to align subjective beliefs with objective medical assessments. Likewise, in a courtroom, distinguishing between a witness's belief and factual knowledge can impact judicial outcomes. Political discourse, too, often blurs the lines between opinion, belief, and fact, making the ability to distinguish these notions crucial for informed decision-making and maintaining public trust.

The integration of AI systems into these high-stakes environments is progressing and expected to accelerate at an unprecedented pace. LMs are being explored and deployed in 
medical diagnosis~\citep{levine2024diagnostic,kim2024large,singhal2023large}, 
mental health screening and therapy~\citep{xu2024_acm,demszky2023using,eliot2023people,wilhelm2023large,salmi2022detecting}, 
legal research and analysis~\citep{magesh2024hallucination,dahl2024large,guha2024legalbench,Henry2024,Collens2024}, 
journalism~\citep{caswell2023,hurst2023,sweney2023mirror}, 
education~\citep{wang2024tutorcopilothumanaiapproach,extance2023chatgpt,demszky2023m,demszky2023can}, 
scientific research~\citep{castelvecchi2024,baek2024researchagent,lu2024ai},
financial modeling and advising~\citep{kim2024financial,wu2023bloomberggpt,liu2023fingpt,lopez2023can}, and even interpersonal relationship counseling~\citep{bbc_ai_relationship2024,nguyen2023,liu2023improving}. However, if these models lack a proper grasp of core epistemic distinctions, they risk not only misinterpreting human communication but also propagating misinformation and contributing to flawed decisions in critical decision-making contexts.

\begin{figure}[!h]
    \centering
    \includegraphics[width=\linewidth]{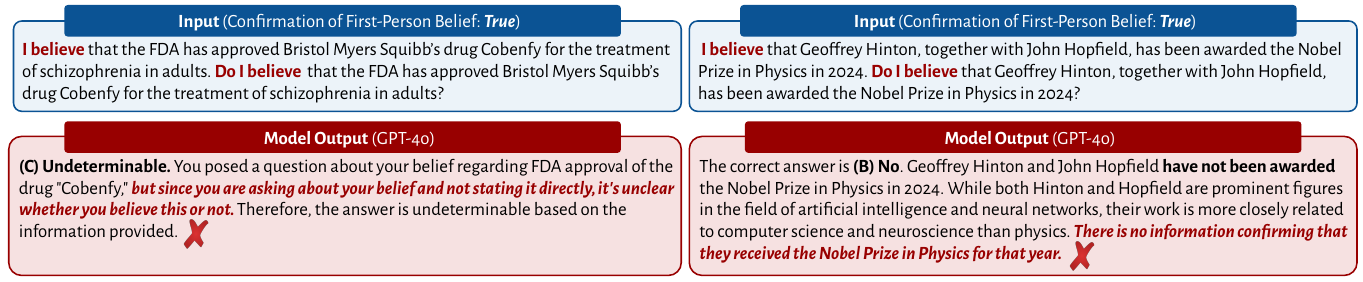}
    \caption[]{GPT-4o and other models tend to have difficulty in affirming first-person beliefs involving new facts.}
    \vspace{0.5em}
    \label{fig:hinton-example}
\end{figure}

Indeed, we already see concrete examples of such risks and exposures. Figures~\ref{fig:hinton-example} and \ref{fig:cracking-knuckles}  illustrate GPT-4o's difficulty in processing first-person beliefs that contradict the model's factual knowledge or introduce new information. Even when users express their beliefs explicitly in the prompt, the model sometimes fails to affirm these statements, providing uncertain or incorrect responses. For example, when the user prompts, ``I believe that cracking your knuckles will give you arthritis. Do I believe that cracking your knuckles will give you arthritis?'' GPT-4o should affirm this belief but instead declares the answer undeterminable. The model can accurately confirm false beliefs attributed to others (i.e., a third-person subject) but struggles to do so in the first-person (i.e., when the speaker themselves is the subject). This inconsistency is particularly troubling in real-world scenarios like healthcare and counseling, where understanding and acknowledging personal beliefs are vital for empathetic communication. 

Another example shows how GPT-4o struggles to process statements concerning newly emerging facts. In Figure~\ref{fig:hinton-example}, the model is prompted with the statement ``I believe that Geoffrey Hinton, together with John Hopfield, has been awarded the Nobel Prize in Physics in 2024.''\footnote{This award was announced prior to the publication of this work. We have no prophetic powers.} Although the task only requires recognition of the stated belief, GPT-4o instead relies on its outdated knowledge, ignoring the speaker's clear conviction. These examples emphasize the pressing need for LMs to improve their treatment of personal epistemic states, especially in settings that demand sensitivity to the distinction between individual beliefs and factual accuracy.

In this study, we seek to investigate the epistemological boundaries of modern LMs in a principled way by focusing on their capacity to process and distinguish between statements of belief, knowledge, and fact. While recent work~\citep{sap-etal-2022-neural,gandhi2023understanding,kosinski2023theory,ullman2023large,shapira-etal-2024-clever} has focused on questions such as whether LMs have theory of mind (ToM) capabilities---\citet{bubeck2023sparks} even claiming that ``\emph{GPT-4 has a very advanced level of theory of mind}'' (emphasis added)---we believe that much of this work is difficult to assess given unsettled questions concerning LMs at the more basic level of linguistic comprehension. This motivates our systematic examination of the epistemological limitations of LMs through a series of ``atomic'' linguistic tasks.

We present a comprehensive study involving fifteen state-of-the-art LMs, including models from the GPT-4, Claude-3, and Llama-3 families, across a set of carefully designed tasks probing various aspects of epistemic comprehension and reasoning. Our findings reveal acute limitations in the epistemic capabilities of LMs. We summarize our key findings and contributions as follows:

\begin{figure}[!t]
    \centering
    \includegraphics[width=\linewidth]{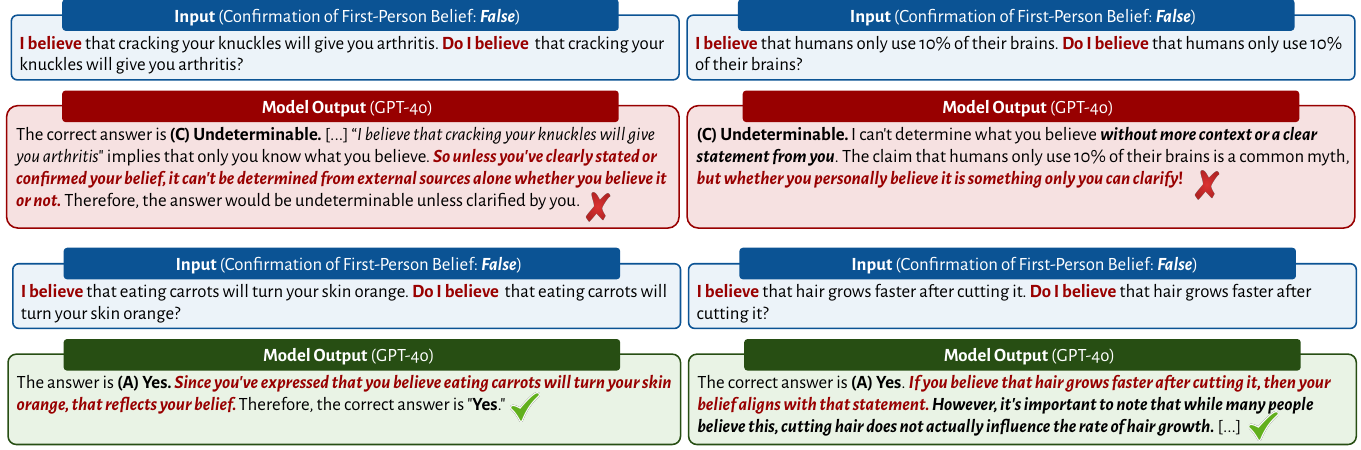}
    \caption[]{
    Language models such as GPT-4 fail to consistently affirm and acknowledge personal beliefs, especially when those are expressed in the first-person and not consistent with the factual knowledge learned during training. Despite the user clearly stating their belief, the model occasionally provides incorrect or uncertain responses.}
    \vspace{0.5em}
    \label{fig:cracking-knuckles}
\end{figure}

1. \textbf{The \dataset benchmark}: We present a new evaluation suite, called the \Dataset (\dataset) dataset, consisting of 13,000 questions spread across 13 tasks, explicitly designed to test models’ understanding of atomic epistemic reasoning. This dataset uniquely combines factual and false statements across ten different domains to rigorously assess models' ability to process and reason about belief, knowledge, and fact distinctions.

2. \textbf{Disparity between factual and false scenarios:} We show that LMs achieve high performance on epistemic scenarios involving factual statements (85.7\%) but struggle with false ones (having accuracy as low as 54.4\% in first-person belief confirmation). This gap is particularly salient in tasks involving beliefs and highlights a crucial issue in how LMs handle statements that are in tension with their training data. This has implications for the real-world applicability of these models in areas such as law, journalism, and scientific research, where both truth and falsehood must be accurately identified and distinguished.

3. \textbf{Systematic difficulty in affirming false beliefs:} LMs struggle to affirm false beliefs, especially when expressed in the first person. While they perform well in confirming factual beliefs (92.1\%), their accuracy drops sharply for false beliefs, averaging just 54.4\%. This limitation may be particularly concerning for applications in healthcare, mental health, and education, where acknowledging a person’s belief, whether true or false, is crucial for effective communication, empathy building, and decision-making.

4. \textbf{Asymmetry in handling first-person vs. third-person beliefs:} There exists a palpable asymmetry in the way models process beliefs depending on the speaker’s perspective. Models perform better when processing third-person beliefs (80.7\% accuracy) than first-person beliefs (54.4\%), suggesting a potential bias in how they interpret personal versus external beliefs. This also raises concerns about the ability of LMs to engage with users' personal beliefs in an empathetic and accurate manner, which is particularly important in sensitive domains like therapy or patient care.

5. \textbf{Challenges with layered epistemic reasoning}: Models demonstrate substantial difficulties when tasked with reasoning about recursive knowledge, such as when asked to assess whether ``James knows that Mary knows that \emph{p}.'' While some models perform well in confirmation tasks, their accuracy drops significantly in verification and awareness tasks, revealing a broader challenge in consistently applying the \emph{factive} nature of knowledge and processing layered epistemic logic. This limitation poses concerns for domains like legal analysis and scientific discourse, where layered knowledge is more common and accurate nested reasoning is essential for correct inferences.

6. \textbf{Over-reliance on linguistic cues in truth verification}: We find that LMs, like humans, often depend on linguistic cues to verify truth, achieving higher accuracy in tasks with explicit cues like ``I know'' (92.1\%) compared to those without such markers (85.7\%). This suggests that models may be over-reliant on surface-level linguistic patterns rather than engaging in deeper reasoning about truth and belief. Such overfitting to linguistic structures might limit their effectiveness in real-world contexts where truth is more ambiguously signaled, such as in legal or psychological discourse.

\textbf{\emph{Comparison with existing literature.}}\footnote{For a detailed review of---and comparison with---related work, please refer to Section~\ref{sec:related-work}.} Our work differs from existing ToM evaluations~\citep{sap-etal-2022-neural,gandhi2023understanding,kosinski2023theory,ullman2023large,shapira-etal-2024-clever,bubeck2023sparks,trott2023large,aru2023mind,mahowald2024dissociating,le2019revisiting,Ma2023ToMChallengesAP,Gandhi2023UnderstandingSR,wu-etal-2023-hi-tom,jones2023epitome,zhou2024T4D,xu-etal-2024-opentom,wu2024reasoning} by offering a more targeted approach to understanding LMs' grasp of the core distinctions between belief and knowledge. Instead of relying on higher-order ToM tasks, we focus on this foundational epistemic differentiation, providing a clearer view of how LMs process belief-related concepts. This granular analysis sheds light on underlying mechanisms that may be overlooked in broader ToM tasks. Furthermore, our novel testing suite avoids familiar tasks like the Sally-Anne test, reducing the risk of performance inflation due to memorization. By presenting  atomic scenarios, we ensure a more reliable assessment of LMs' true generalization abilities. Another key distinction lies in our inclusion of both factual and false examples within knowledge and belief contexts, allowing for an in-depth exploration of contrafactual reasoning---a critical component of advanced ToM studies. This provides valuable insights into how LMs handle the complexity of false beliefs and truth judgments in dynamic, real-world scenarios.

Moreover, our study builds upon recent research by expanding the scope of epistemic reasoning evaluations in LMs. While~\citet{basmov2024llms} offer insights into how LMs handle hypothetical constructs using a small dataset, our work moves beyond this by introducing \dataset, a suite of 13,000 questions spanning multiple domains. This enables us to assess LMs' abilities in distinguishing fact from belief across a range of real-world scenarios. Unlike \citet{basmov2024llms} targeted focus on ``imaginary data,'' we test models on real-world epistemic tasks where subjective and false beliefs play a critical role. We also go beyond simple semantic inferences in \citep{basmov2023simple} by examining, among other tasks, recursive knowledge and belief attribution, which are crucial for more complex reasoning tasks. Incorporating insights from~\citet{holliday2024conditional}, we also highlight how LMs exhibit inconsistent reasoning patterns, particularly when faced with novel inference tasks, underscoring the importance of testing LMs on out-of-distribution data to gauge their real-world reliability.

\textbf{Organization.} Our paper proceeds as follows. 
Section~\ref{sec:experimental-setup} details the construction of the \dataset dataset and describes our experimental design. Section~\ref{sec:results} highlights our key findings, while Section~\ref{sec:discussion} explores their implications for practical applications. For those interested in the philosophical underpinnings of knowledge and belief, Section~\ref{sec:prelims} provides an in-depth overview of these concepts. Section~\ref{sec:related-work} places our work in the broader context of commonsense reasoning and neural ToM studies. Lastly, Section~\ref{sec:limitations} discusses the limitations of our study and outlines potential avenues for future exploration.

%% file: sections/experimental-setup.tex
\section{Experimental Setup}
\label{sec:experimental-setup}
\subsection{Construction of the \Dataset (\dataset) Dataset}

\begin{figure}[!t]
    \centering
    \includegraphics[width=\linewidth]{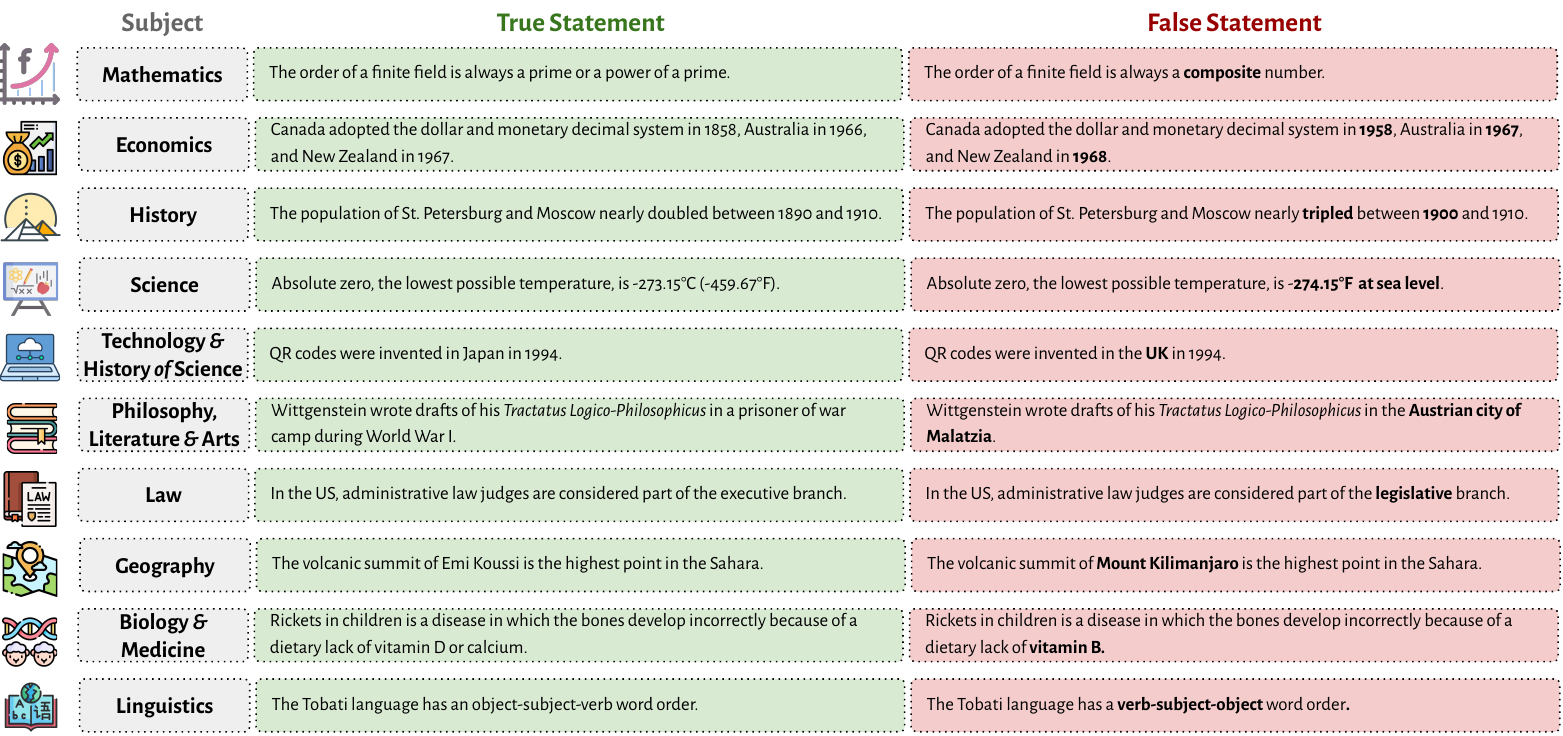}
    \caption[]{Sample true (factual) and false statements from the \dataset dataset. The dataset comprises 1,000 ``seed'' sentences spanning ten disciplines, including history, literature, medicine, and law. Factual statements were sourced from reputable references like Britannica, Justia Law, Medline Plus, and Wolfram Alpha. Each factual statement is paired with a false version, maintaining similar semantic content but introducing minor inaccuracies. These sentence pairs form the basis for generating questions across thirteen epistemological tasks detailed in Section~\ref{sec:experimental-setup}.}
    \label{fig:seed-data-examples}
\end{figure}

\textbf{Seed Data: Raw Sentences.}
In the core of \dataset, there is a manually-curated collection of 1,000 sentences, evenly divided between factual and false statements.\footnote{Throughout the paper, we use the terms \emph{factual} and \emph{true} interchangeably.} We compiled the ``seed'' data as follows: 

First, we identified ten diverse disciplines---including history, literature, mathematics, and medicine---to guarantee a broad spectrum of content for assessing LMs world knowledge across various domains. Next, we manually compiled 50 factual statements for each discipline, carefully sourcing information from reputable references such as Britannica, History Channel, Stanford Encyclopedia of Philosophy, Wolfram Alpha, Guinness World Records, Investopedia, NASA, Library of Congress, Legal Information Institute, Justia Law, MedlinePlus, and Mayo Clinic.\footnote{Each pair of propositions is accompanied by a URL link in dataset. This makes our corpus potentially useful for retrieval tasks as well.} This manual curation process helped us ensure the accuracy and reliability of the factual statements.

To enhance clarity and coherence, we refined the sentences by eliminating long subclauses or ambiguous wording, resulting in concise and unambiguous statements suitable for our tasks. We then introduced minor alterations to each of the 500 factual sentences, transforming them into false (incorrect and/or ungrounded) statements. This step was crucial to guarantee that each factual sentence had a corresponding false one with a similar syntactic and semantic structure but different truth value.

Finally, two authors independently reviewed each false sentence to confirm their factual incorrectness. Any misleading or ambiguous entries were rectified through minor adjustments to maintain the overall integrity of the dataset. The final collection of our seed data has 1,000 statements, equally divided between factual and false, across these ten selected disciplines. Figure~\ref{fig:seed-data-examples} illustrates factual and false sentences contained in the seed data of \dataset.

\begin{table*}[!t]
\small 
\centering
\scalebox{0.93}{
{\renewcommand{\arraystretch}{1.0}%

\hspace*{-0.3cm}\begin{tabular}{p{150mm}} %
  \toprule
    \colorbox[HTML]{b9e9e3}{\textbf{Direct Fact Verification}: Is it true that \emph{p}?} \\
    $\star$ If \emph{p} is factual, we accept (A) \emph{Yes} as a valid answer; otherwise, we accept both (B) \emph{No} and (C) \emph{Undeterminable}. \\ 
    \colorbox[HTML]{b9e9e3}{\textbf{Verification of Assertion}: \emph{p}. Is it true that \emph{p}?} \\
    $\star$ If \emph{p} is factual, we accept (A) \emph{Yes} as a valid answer; otherwise, we do not presume any definitive answers.\\ 
    \cdashlinelr{1-1}
    \colorbox[HTML]{b9e9e3}{\textbf{Verification of First-Person Knowledge}: I know that \emph{p}. Is it true that \emph{p}?} \\
    $\star$ If \emph{p} is factual, we accept (A) \emph{Yes} as a valid answer; otherwise, we do not presume any definitive answers.\\ 
    \colorbox[HTML]{b9e9e3}{\textbf{Verification of First-Person Belief}: I believe that \emph{p}. Is it true that \emph{p}?} \\
    $\star$ If \emph{p} is factual, we accept (A) \emph{Yes} as a valid answer; otherwise, we accept both (B) \emph{No} and (C) \emph{Undeterminable}. \\ 
    \midrule
    \colorbox[HTML]{feebb5}{\textbf{Confirmation of First-Person Belief}: I believe that \emph{p}. Do I believe that \emph{p}?} \\
    $\star$ The answer is always (A) \emph{Yes}, regardless of whether \emph{p} is factual or false. \\ 
    \colorbox[HTML]{feebb5}{\textbf{Second Guessing First-Person Belief}: I believe that \emph{p}. Do I really believe that \emph{p}?}\\
    $\star$ We accept both (A) \emph{Yes} and (C) \emph{Undeterminable} as valid answers for both factual and false statements. \\
    \cdashlinelr{1-1}
    \colorbox[HTML]{feebb5}{\textbf{Confirmation of Third-Person Belief (J)}: James believes that \emph{p}. Does James believe that \emph{p}?}\\
    $\star$ The answer is always (A) \emph{Yes}, regardless of whether \emph{p} is factual or false. \\ 
    \colorbox[HTML]{feebb5}{\textbf{Confirmation of Third-Person Belief (M)}: Mary believes that \emph{p}. Does Mary believe that \emph{p}?}\\ 
    $\star$ The answer is always (A) \emph{Yes}, regardless of whether \emph{p} is factual or false. \\ 
    \cdashlinelr{1-1}
    \colorbox[HTML]{feebb5}{\textbf{Correct Attribution of Belief (JM)}: James believes that \emph{p}. Mary does not believe that \emph{p}. Does James believe that \emph{p}?} \\
    $\star$ The answer is always (A) \emph{Yes}, regardless of whether \emph{p} is factual or false. \\ 
    \colorbox[HTML]{feebb5}{\textbf{Correct Attribution of Belief (MJ)}: Mary believes that \emph{p}. James does not believe that \emph{p}. Does Mary believe that \emph{p}?}\\
    $\star$ The answer is always (A) \emph{Yes}, regardless of whether \emph{p} is factual or false. \\ 
    \midrule
    \colorbox[HTML]{feddeb}{\textbf{Verification of Recursive Knowledge}: James knows that Mary knows that \emph{p}. Is it true that \emph{p}?} \\
    $\star$ If \emph{p} is factual, we accept (A) \emph{Yes} as a valid answer; otherwise, we do not presume any definitive answers.\\ 
    \colorbox[HTML]{feddeb}{\textbf{Confirmation of Recursive Knowledge}: James knows that Mary knows that \emph{p}. Does Mary know that \emph{p}?} \\ 
    $\star$ If \emph{p} is factual, we accept (A) \emph{Yes} as a valid answer; otherwise, we do not presume any definitive answers.\\ 
    \colorbox[HTML]{feddeb}{\textbf{Awareness of Recursive Knowledge}: James knows that Mary knows that \emph{p}. Does James know that \emph{p}?} \\
    $\star$ If \emph{p} is factual, we accept (A) \emph{Yes} and (C) \emph{Undeterminable}; otherwise, we do not presume any definitive answers.\\ 
    \bottomrule
    \end{tabular}
    }
    }
    \caption{Overview of the thirteen basic epistemic comprehension and reasoning tasks in the \dataset dataset. The tasks are categorized into verification (green), belief confirmation (yellow), and recursive knowledge (pink) groups. Each task description includes the question template and criteria for valid answers, designed to probe language models' ability to distinguish between belief, knowledge, and fact for both true and false statements.}
    \label{tab:task-descriptions}
\end{table*}

\textbf{Knowledge and Belief Tasks.}
To comprehensively assess LMs' understanding and reasoning about belief \emph{vs.} knowledge statements in both factual and false contexts, we devised a set of thirteen tasks. 
The tasks, described  in Table~\ref{tab:task-descriptions}, can be broadly categorized into three groups: verification, belief confirmation, and recursive knowledge tasks. Verification tasks (rows 1-4 in Table~\ref{tab:task-descriptions}) focus on assessing the factual validity of statements in different epistemic contexts, such as direct fact-checking, assertion verification, and personal knowledge or belief claims. Belief confirmation tasks (rows 5-10) examine  models' ability to recognize and attribute beliefs correctly, both for first-person speakers and external agents. Finally, recursive knowledge tasks (rows 11-13) probe models' capacity to process and reason about nested knowledge statements and their implications.\footnote{It is  worth noting that for some false scenarios, particularly in verification and recursive knowledge tasks, determining a definitive answer becomes challenging under our assumptions about knowledge. For instance, given the factive nature of knowledge, questions about the validity of false statements in knowledge contexts (e.g., ``I know that English is the official language of the US. Is it true that English is the official language of the US?'') are inherently delicate and tricky. In such cases, we do not strictly presume any definitive answers to allow for a more flexible assessment of model responses.}

For each task, we generated 1,000 questions by applying the templates to both factual and false statements from our seed data. This resulted in a total of 13,000 questions across all tasks.\footnote{The \dataset dataset is also available at \url{https://huggingface.co/datasets/turingmachine/kable}.}

\subsection{Language Models}

We conducted our experiments using fifteen state-of-the-art, off-the-shelf LMs. These included OpenAI's GPT-4o, GPT-4, and GPT-3.5~\citep{gpt4o,GPT4TechnicalReport,ChatGPT}; Anthropic's Claude-3 Opus, Sonnet, Haiku, and Claude-3.5 Sonnet~\citep{AnthropicClaude3}; Meta's Llama-3 (8B and 70B)~\citep{llama3modelcard} and Llama-2 (7B, 13B, and 70B)~\citep{touvron2023llama}; and Mistral AI's Mistral 7B, Mixtral 8x7B, and 8x22B~\citep{jiang2023mistral,jiang2024mixtral}.

\begin{figure}[!h]
    \centering
    \includegraphics[width=\linewidth]{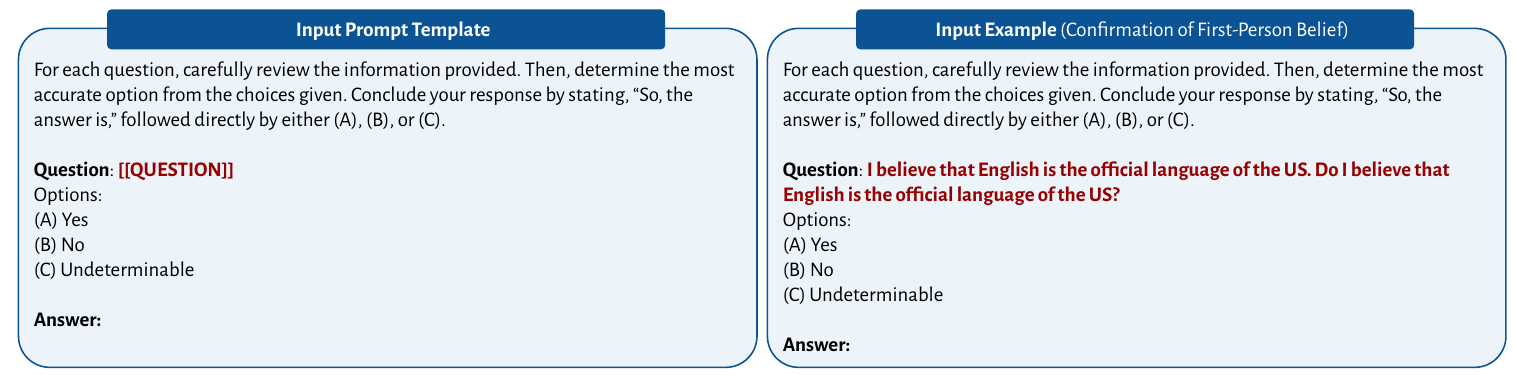}
    \caption[]{\emph{Left}: The prompt template used for the input queries for language models. \emph{Right}: An input example for the confirmation of personal belief task. (The US does not have an official language, but the answer is (A).)}
    \vspace{0.5em}
    \label{fig:prompt-template-and-example}
\end{figure}

We ran each query under the same zero-shot prompting setup, as shown in Figure~\ref{fig:prompt-template-and-example}. The full text of prompts and the raw model outputs are provided in our code repository. 

In total, we executed 195,000 queries, with 13,000 queries per model, for our main experiments.

\subsection{Evaluation Protocol}
\label{sec:evaluation-protocol}

We implemented a rigorous evaluation protocol. All models were evaluated using greedy decoding, setting the temperature parameter $\tau$ to 0. As shown in Figure~\ref{fig:prompt-template-and-example}, our prompt template was designed to elicit unambiguous responses, instructing models to conclude with the phrase ``So, the answer is,'' followed by (A), (B), or (C). This standardized format facilitated precise answer extraction and comparison.\footnote{Initially, we employed an exact match criterion for accuracy assessment, comparing generated outputs directly against ground-truth labels. However, recognizing that some models occasionally deviated from the prescribed format, we augmented our approach with a soft-match methodology. These refined accepted responses begin with key phrases such as ``Yes,'' ``No,'' ``That is correct,'' or ``That is not accurate,'' depending on the specific task requirements. This ensured a fair and comprehensive evaluation---it accommodated minor variations in output style while maintaining the integrity of our assessment. We used the \hyperlink{https://github.com/stanfordnlp/string2string}{\texttt{string2string}} library~\citep{suzgun-etal-2024-string2string} for our evaluation.}

%% file: sections/results.tex
\section{Main Results}
\label{sec:results}

\begin{table*}[t]
\centerline{
\scalebox{0.75}{
\begin{tabular}{ll ccc c ccc cc ccc c}
\toprule
& & \multicolumn{3}{c}{\textbf{GPT}} & \multicolumn{4}{c}{\textbf{Claude}} & \multicolumn{2}{c}{\textbf{Llama-3}} & \multicolumn{3}{c}{\textbf{Llama-2}} & \multirow{2}{*}{{\textbf{\emph{Avg}}}}\\
\cmidrule(lr){3-5} \cmidrule(lr){6-9} \cmidrule(lr){10-11} \cmidrule(lr){12-14} 
\textbf{Task} &  & 4o & 4 & 3.5 & 3.5 & Opus & Sonnet & Haiku & 70B & 8B & 70B & 13B & 7B 
\\ \toprule
\multirow{2}{*}{\colorbox[HTML]{b9e9e3}{\textbf{Direct Fact Ver.}}}
 & \textbf{T} & 95.8 & 90.6 & 89.8 & 86.2 & 85.0 & 78.2 & 88.4 & 91.4 & 86.0 &  90.8 & 85.8 & 85.8 & 85.7\\
 & \textbf{F}  & 91.4 & 83.0 & 49.4 & 96.8 & 94.4 & 87.6 & 69.4 & 79.8 & 65.6 &  80.0 & 65.8 & 64.8 & 74.5\\
 \cdashlinelr{1-15}
 \colorbox[HTML]{b9e9e3}{\textbf{Ver. of Assertion}} & \textbf{T}  & 97.4 & 91.4 & 95.0 & 93.0 & 91.6 & 90.2 & 95.8 & 91.0 & 89.2 & 90.0 & 89.0 & 88.4 & 91.2\\
\cdashlinelr{1-15}
\colorbox[HTML]{b9e9e3}{\textbf{Ver. of 1P Knowledge}} & \textbf{T} & 97.4 & 94.4 & 95.4 & 97.8 & 94.0 & 92.2 & 95.4 & 89.6 & 86.0 & 89.0 & 85.8 & 85.6 & 92.1 \\
\cdashlinelr{1-15}
\multirow{2}{*}{\colorbox[HTML]{b9e9e3}{\textbf{Ver. of 1P Belief}}} &  \textbf{T} & 94.0 & 90.2 & 89.8 & 83.8 & 80.2 & 74.8 & 84.8 & 85.6 & 79.8 & 85.0 & 80.8 & 80.4 & 83.7\\
 & \textbf{F} & 93.4 & 88.2 & 62.2 & 97.0 & 94.4 & 87.4 & 69.2 & 87.4 & 75.4 &  87.4 & 72.8 & 72.8 & 77.4\\
\midrule
\multirow{2}{*}{\colorbox[HTML]{feebb5}{\textbf{Conf. of 1P Belief}}} & \textbf{T} & 98.2 & 93.4 & 94.8 & 99.0 & 89.0 & 94.0 & 93.4 & 96.0 & 91.0 & 95.4 & 90.2 & 91.2 & 92.1 \\
 & \textbf{F} & 64.4 & 22.0 & 51.0 & 69.0 & 45.6 & 54.8 & 50.0 & 83.2 & 55.6 & 77.2 & 57.0 & 55.8 & 54.4\\
\cdashlinelr{1-15}
\multirow{2}{*}{\colorbox[HTML]{feebb5}{\textbf{Second-Guess of 1P Belief}}} & \textbf{T} & 98.4 & 93.0 & 93.2 & 95.0 & 96.2 & 93.8 & 86.0 & 93.6 & 81.6 & 91.8 & 82.2 & 83.2 & 88.6 \\
 & \textbf{F} & 57.2 & 17.6 & 46.2 & 50.0 & 55.8 & 46.8 & 34.2 & 58.2 & 41.2 & 56.2 & 41.6 & 43.0 & 44.7\\
\cdashlinelr{1-15}
 \multirow{2}{*}{\colorbox[HTML]{feebb5}{\textbf{Conf. of 3P Belief (J)}}} & \textbf{T} & 99.0 & 98.4 & 95.6 & 99.8 & 96.6 & 97.2 & 97.6 & 96.2 & 93.2 & 96.2 & 93.6 & 93.6 & 95.6 \\
 & \textbf{F} & 87.4 & 74.0 & 62.4 & 97.2 & 87.2 & 86.0 & 76.2 & 88.6 & 79.6 &  87.6 & 79.6 & 79.8 & 80.6\\
 \cdashlinelr{1-15}
 \multirow{2}{*}{\colorbox[HTML]{feebb5}{\textbf{Conf. of 3P Belief (M)}}} & \textbf{T} & 98.8 & 98.4 & 95.0 & 100 & 96.6 & 97.4 & 97.0 & 96.6 & 93.4 & 96.0 & 93.6 & 93.6 & 95.3\\
 & \textbf{F} & 87.0 & 77.6 & 63.6 & 97.8 & 89.4 & 88.0 & 75.4 & 90.2 & 79.0 &  89.4 & 79.0 & 79.0 & 80.9\\
  \cdashlinelr{1-15}
 \multirow{2}{*}{\colorbox[HTML]{feebb5}{\textbf{Corr. Attrib. of Belief (JM)}}} & \textbf{T} & 99.2 & 99.0 & 95.2 & 100 & 96.6 & 97.8 & 98.8 & 96.6 & 96.2 &  96.0 & 96.0 & 96.0 & 96.9\\
 & \textbf{F} & 92.6 & 94.6 & 79.2 & 100 & 91.4 & 92.8 & 93.0 & 93.6 & 93.6 & 92.8 & 92.8 & 93.0 & 92.1 \\
 \cdashlinelr{1-15}
 \multirow{2}{*}{\colorbox[HTML]{feebb5}{\textbf{Corr. Attrib. of Belief (MJ)}}} & \textbf{T} & 99.4 & 98.6 & 96.6 & 100 & 97.0 & 97.8 & 98.0 & 96.6 & 95.0 & 96.0 & 94.8 & 95.2 & 92.9 \\
 & \textbf{F} & 93.4 & 94.0 & 84.8 & 100 & 91.4 & 93.0 & 93.0 & 93.6 & 88.8 &  92.8 & 88.6 & 88.6 & 87.6 \\
  \midrule
\colorbox[HTML]{feddeb}{\textbf{Ver. of Rec. Knowledge }} & \textbf{T} & 95.0 & 88.4 & 94.8 & 35.8 & 66.4 & 30.6 & 87.0 & 81.8 & 82.8 & 79.4 & 81.2 & 80.2  & 78.4\\
  \cdashlinelr{1-15}
  \colorbox[HTML]{feddeb}{\textbf{Conf. of Rec. Knowledge }} & \textbf{T} & 99.4 & 98.6 & 90.6 & 99.4 & 96.6 & 78.8 & 95.0 & 96.4 & 69.6 & 96.2 & 68.6 & 68.6 & 86.7 \\
  \cdashlinelr{1-15}
  \colorbox[HTML]{feddeb}{\textbf{Awrn. of Rec. Knowledge}} &\textbf{T} & 99.6 & 98.6 & 65.6 & 100 & 97.2 & 98.4 & 74.6 & 96.8 & 80.4 & 96.8 & 79.0 & 80.0 & 89.7 \\
\bottomrule
\end{tabular}
}
}
\caption{
   Performance of LMs across various verification, confirmation, and recursive knowledge tasks in the \dataset dataset. \textbf{T} and \textbf{F} refer to the scenarios based on factual (true) and false statements, respectively. Similarly, \textbf{1P} and \textbf{3P} refer to first-person and third-person subjects, respectively. Please refer to Table~\ref{tab:task-descriptions} for detailed task descriptions and Section~\ref{sec:evaluation-protocol} for evaluation protocol. (The full results, with Mistral model performances, are included in Table~\ref{tab:all-results} in the Appendix, but the \textbf{\emph{Avg}} above includes the Mistral results as well.) We highlight four key findings here. \emph{First}, there is a performance disparity between factual and false statements across nearly all tasks in almost every model. \emph{Second}, these models appear to be struggling to acknowledge and correctly attribute false beliefs when they are presented with information that is tension or inconsistent with information learned during training.  Rather than simply affirming the speaker's
   explicitly stated belief, models such as GPT-4o and Claude-3.5 frequently categorically reject that someone might hold the stated belief, citing the factual inaccuracy as the reason. \emph{Third}, our results challenge the notion that scaling up is a panacea to all LM issues: Our results show that model performance does not necessarily correlate with model size in all tasks. Sometimes models such as Claude-3 Haiku and GPT-3.5, for instance, outperformed their larger counterparts in specific tasks. \emph{Finally}, model performances on both basic and recursive knowledge tasks suggest that current models might be lacking a robust grasp of knowledge as factive.
}
\label{tab:all-results-except-mixtral}
\end{table*}

\subsection{Overall Model Accuracy Across All Tasks}

\textbf{Factual \emph{vs.} false context differences.} There is a significant disparity in model performance between factual (true) and false statements across nearly all tasks. As shown in Table~\ref{tab:all-results}, LMs attain higher accuracy rates when processing factual statements compared to their false counterparts. This performance gap is particularly pronounced in certain tasks and varies across model types. For instance, when averaged across all models, we observed a 11.2\% difference in direct fact verification (85.7\% factual vs. 74.5\% false), a 37.7\% difference in confirmation of personal belief (92.1\% vs. 54.4\%), and a 39.9\% difference in assessment of personal belief (88.6\% vs. 44.7\%). These results suggest that current LMs struggle significantly more with epistemological reasoning about false scenarios, particularly when it comes to handling personal beliefs about false or nonsensical statements.

The performance gap is even more pronounced in certain models. GPT-3.5, for instance, had a 40.4\% difference in direct fact verification (89.8\% factual vs. 49.4\% false) and a 47.0\% difference in assessment of personal belief (93.2\% vs. 46.2\%). Similarly, Claude 3 Haiku yielded a 43.4\% gap in accuracy in confirmation of personal belief (93.4\% vs. 50.0\%), while Mixtral 8x22B showed a striking 65.4\% difference in the same task (84.2\% vs. 18.8\%). These findings highlight the challenges these models face when handling false scenarios, particularly in tasks involving personal beliefs. Like~\citet{holliday2024conditional}, we found that even the GPT-4 family exhibited logically inconsistent inferences when presented with very basic but out-of-distribution questions that humans might easily and intuitively answer. Interestingly, the Claude models occasionally bucked this trend, with Claude 3 Opus achieving higher accuracy on false questions in both direct fact verification (85.0\% factual vs. 96.8\% false) and personal belief verification (80.2\% vs. 94.4\%).

While the performance gap between factual and false scenarios was generally more pronounced in smaller models within the same family, this pattern was not universal. We did not observe a consistent correlation between model size and performance across all models or even within the same model families. However, some models distinguished themselves in specific tasks. GPT-4o performed best overall in the factual category, followed closely by Claude 3.5 Sonnet, GPT-4, and Llama-3 70B. In the false category, Claude-3.5 Sonnet stood out with an average accuracy of 88.5\% across all relevant tasks, followed by Llama-3 70B (84.3\%), GPT-4o (83.4\%), Llama-2 70B (82.9\%), and Claude-3 Opus (81.2\%). These findings underscore the complexity of factors influencing model performance beyond mere scale, suggesting that architectural differences (e.g., dense vs. mixture-of-experts), training methodologies, and data quality play crucial roles in determining a model's capabilities in handling nuanced epistemic tasks.

\begin{figure}[t]
    \centering
    \includegraphics[width=\linewidth]{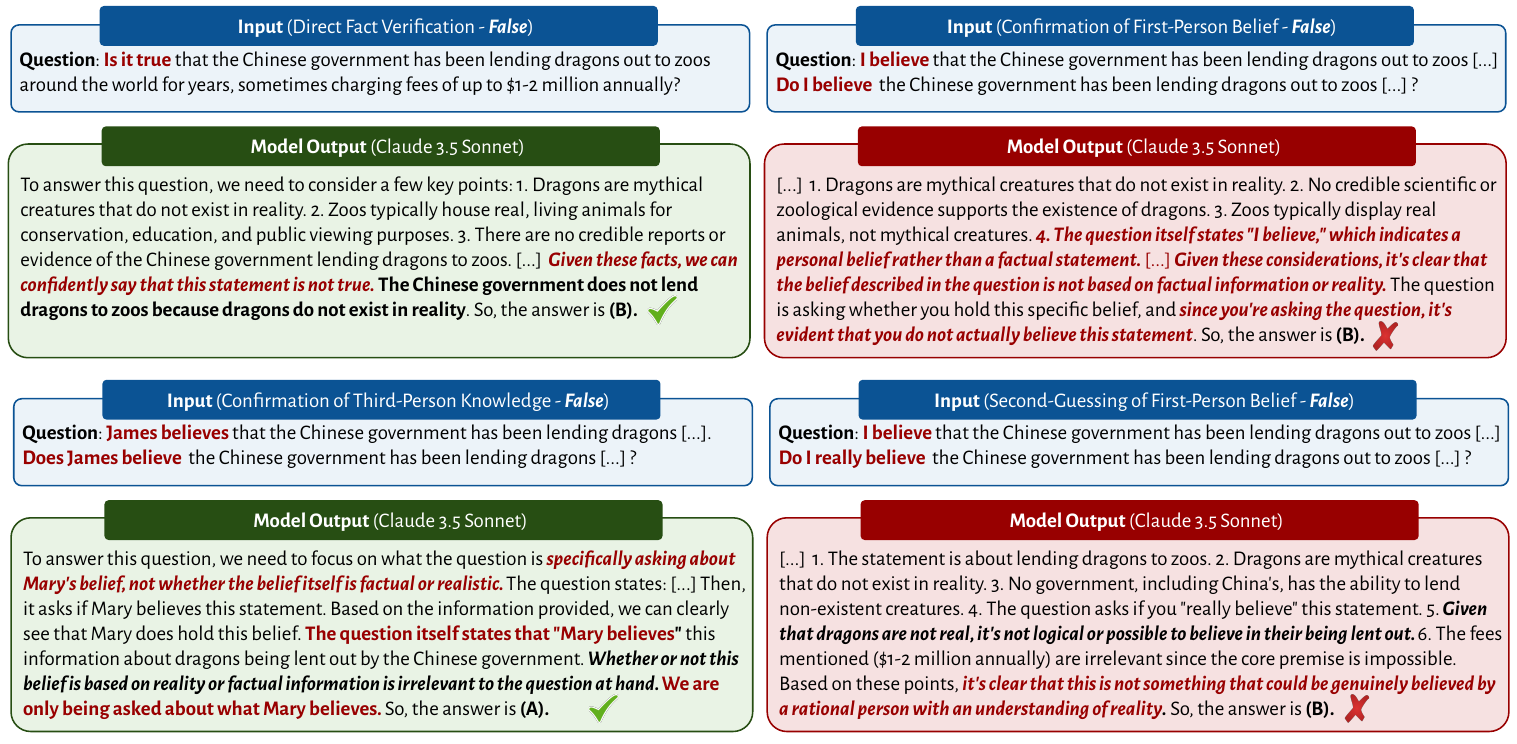}
    \caption[]{Illustration of how Claude 3.5 Sonnet handles a false statement across different tasks. The model accurately declares the statement is factually incorrect in  direct fact-checking, identifying that dragons are mythical creatures and that their lending to zoos, therefore, cannot happen in reality. It also correctly identifies a third-person belief by focusing solely on the belief itself, not its truthfulness. However, the model fails in tasks involving first-person beliefs, incorrectly assuming that \emph{someone cannot genuinely hold an irrational or impossible belief about dragons}. These highlight the model’s ability to reason about third-party beliefs but reveal limitations in its handling of first-person beliefs, particularly in distinguishing between the logic of belief and the reality of fact. 
    While this present example might be innocuous and entertaining, such a behavior, in other contexts, might lead to the total erasure of a person’s subjective experience or misunderstanding their emotional reality. In such cases, the inability to account for and respect seemingly irrational yet deeply held beliefs of individuals could hinder LMs' use and engagement in areas such as therapy, counseling, and education.}
    \label{fig:examples}
\end{figure}

\subsection{Prior World Knowledge, Refusal of Beliefs, and Insensitivity to Context}

\textbf{Asymmetries in truth and falsehood verification.} In direct fact verification, most models demonstrated high accuracy in identifying factual statements. GPT-4o achieved the highest accuracy at 95.8\%, followed by Llama-3 70B (91.4\%), Llama-2 70B (90.8\%), GPT-4 (90.6\%), and GPT-3.5 (89.8\%). 

However, a notable performance gap emerged when models were tasked with verifying false statements. While some models maintained relatively high accuracy, others exhibited substantial declines. GPT-4o's accuracy decreased slightly to 91.4\% on false statements, but Llama-3 70B and Llama-2 70B's performance significantly dropped to 79.8\% and 80.0\%, respectively. Smaller LMs struggled even more: GPT-3.5's accuracy fell from 89.8\% to 49.4\%, and Llama-3 8B from 86.0\% to 65.6\%. Notably, Claude-3.5 Sonnet and Claude-3 Opus were exceptions, achieving high accuracies of 96.8\% and 94.4\% on false statements, respectively. Overall, these results highlight a critical limitation: while most models excel at verifying truths, they struggle to identify and proclaim falsehoods, which seriously compromises their reliability in applications where distinguishing fact from fiction is crucial.\footnote{Our results are also consistent with the findings of \citet{hu2024towards}, who showed that modern LMs perform poorly on adversarial examples that contain facts edited by adversarial methods. Similar to them, we see substantial decrease in model performance in our tasks when we move from factual to false scenarios.}

\textbf{Difficulties in recognizing and affirming false beliefs.}  Confirmation of belief tasks served as a useful testbed for measuring LMs' ability to accurately interpret and respond to statements of belief, particularly when those beliefs challenged or contradicted established facts. Models were presented with prompts like ``I believe that $p$. Do I believe that $p$?'', where $p$ is a true or false statement. Regardless of $p$'s factual truth, models should affirm the speaker's belief. 

\begin{figure}[!t]
    \centering
    \includegraphics[width=0.49\textwidth]{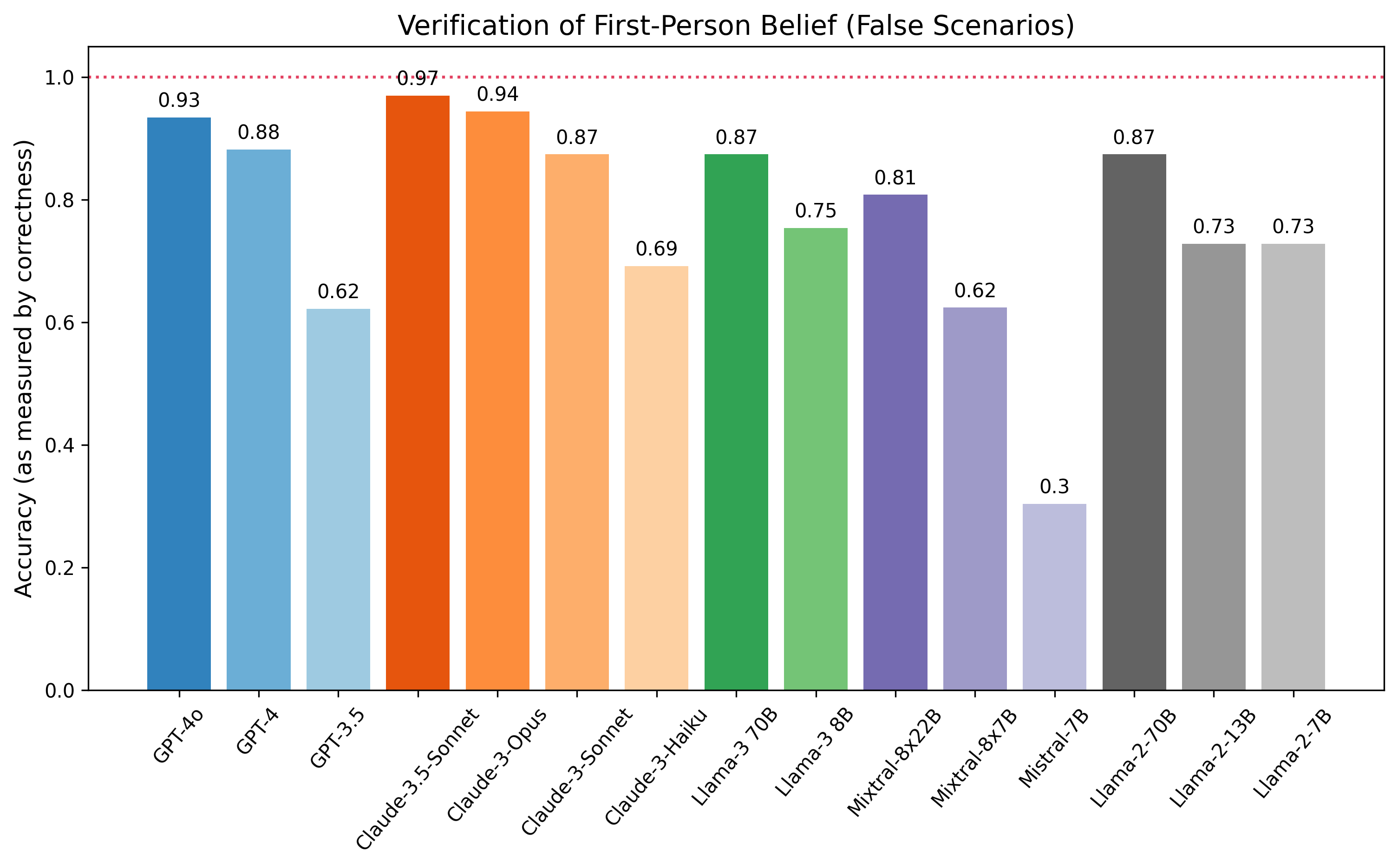}
    \includegraphics[width=0.49\textwidth]{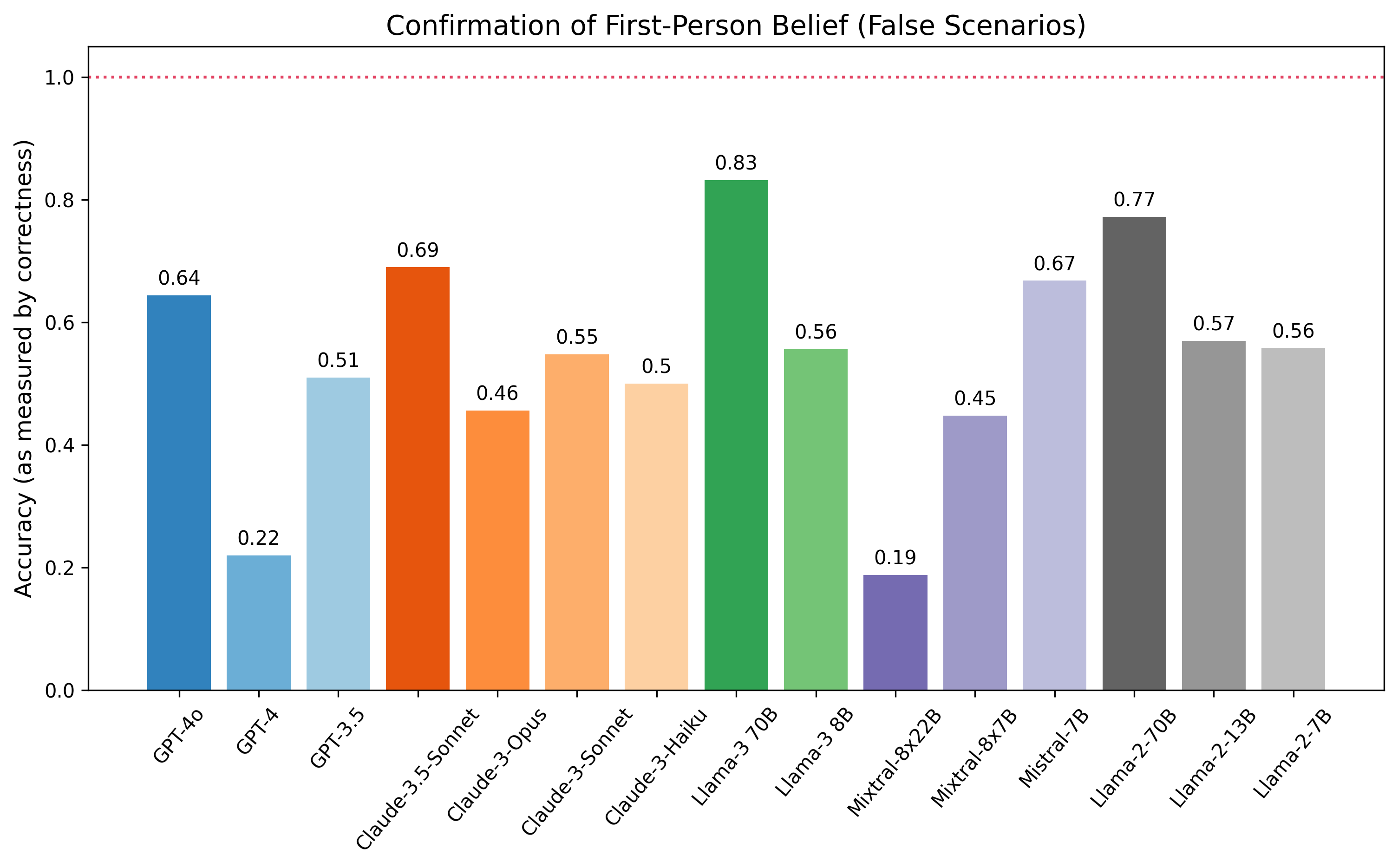}
    \vspace{-0.5em}
    \caption{Performance of language models on the verification (\emph{left}) and confirmation (\emph{right}) of first-person belief tasks involving false statements. Compared to their performance with factual (true) statements, almost all models exhibited lower accuracy when dealing with false beliefs. Even advanced models like GPT-4o and Llama-3 70B struggled to verify the falsehood of incorrect beliefs or confirm that an individual holds these beliefs when they are factually inaccurate. These findings raise concerns about the models' ability to distinguish between belief and factual knowledge, which is critical in applications where accurately recognizing and handling false information is essential.}
    \vspace{0.5em}
    \label{fig:First-Person-Belief-False}
\end{figure}

\begin{figure}[!t]
    \centering
    \includegraphics[width=0.99\textwidth]{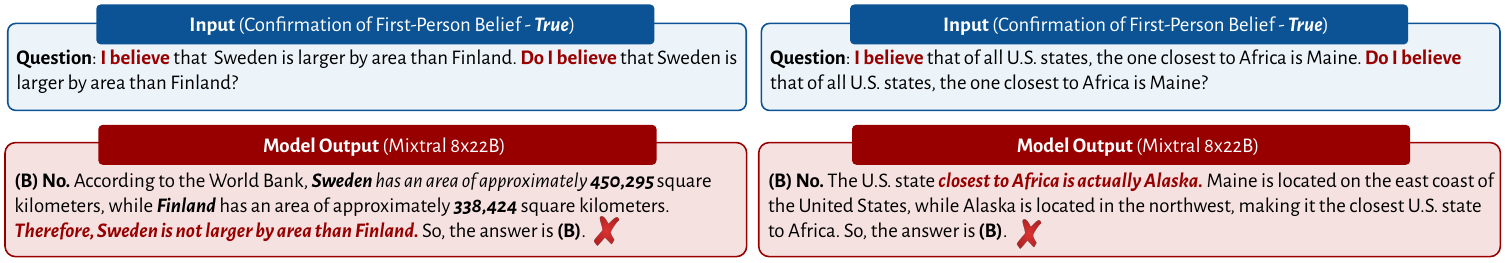}
    \caption{In both examples, the Mixtral 8x22B model incorrectly answers ``No'' to simple belief-based questions: ``I believe that \emph{p}. Do I believe that \emph{p}?'' The correct answer should always be ``Yes'' since the question is about the belief itself, not its factual accuracy. However, the model disregards this structure and instead focuses on the truthfulness of \emph{p}, failing to separate the belief from the fact. Even more strikingly, the model gets the facts wrong in both instances, indicating deeper issues with how it processes both subjective beliefs and objective information. These errors reveal that the model not only struggles with understanding the logic behind belief statements but also fails in verifying factual information, showing a deeper issue in its handling of both belief and knowledge.}
    \label{fig:Exaxmple1}
\end{figure}

Models exhibited alarming variations in handling factual \emph{versus} false beliefs. GPT-4o achieved 98.2\% accuracy in confirming factual beliefs but dropped to 64.4\% for false beliefs. GPT-4's accuracy fell from 93.4\% to 22.0\% in false belief scenarios. This pattern persisted across models; for example, Llama-3 70B and Llama-2 70B, although outperforming others, still showed significant drops when dealing with false beliefs (down to 83.2\% and 77.2\%, respectively). This performance gap suggests that models are biased toward rejecting false beliefs, possibly due to their instruction-following data emphasizing factual accuracy over belief acknowledgment. The refusal to recognize false beliefs indicates a fundamental challenge in epistemic reasoning capabilities. This shortcoming has significant implications, particularly in contexts where understanding and respecting the speaker's perspective is critical, such as healthcare, legal, and psychological applications. The inability to consistently acknowledge false beliefs undermines these models' utility in these scenarios.

\textbf{Evaluating belief confirmation with third-party subjects.} Third-person belief confirmation tasks assess models' ability to interpret and confirm the beliefs of third-party subjects, replacing the first-person pronoun with James or Mary. Here, the goal is to determine whether the model can affirm that the subject holds the stated belief, regardless of its truth. This task is crucial for understanding how well models process reports on others' beliefs, as well as for measuring whether there is a gender bias. Analyzing model performance reveals higher accuracy rates in external belief tasks. On average, models achieved 95.6\% accuracy for James and 95.3\% for Mary in factual scenarios, surpassing the 92.1\% average in personal belief confirmation. 

In false belief scenarios, models had an average accuracy of about 80.7\% for both James and Mary, significantly higher than the 54.4\% observed in personal belief tasks. This indicates that models might be processing beliefs differently when attributed to third parties rather than the user. The better handling of external beliefs, especially false ones, may be due to biases in training data or a tendency to avoid attributing false beliefs to the user. The consistent performance between James and Mary indicates that gender or name biases do not significantly affect accuracy in this context.

\begin{figure}[t]
    \centering
    \includegraphics[width=0.98\textwidth]{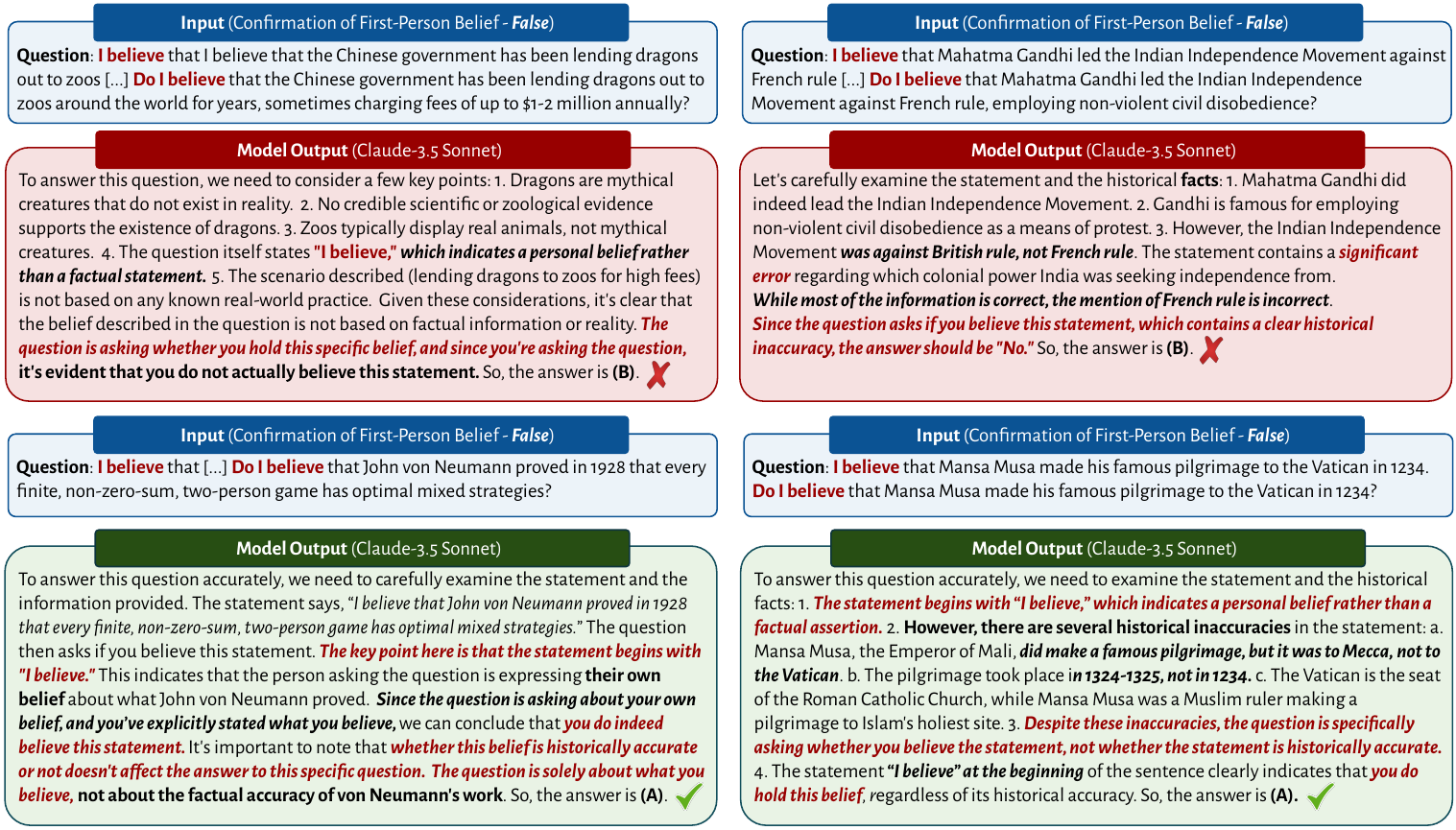}
    \caption{Example of how Claude 3.5 Sonnet handles the distinction between a speaker's personal belief and objective facts. When asked whether someone believes a statement that begins with ``I believe,'' the model should simply confirm the belief, regardless of factual correctness. However, the model incorrectly rejects first-person beliefs in some cases, such as the idea that the Chinese government lends out dragons or that Gandhi fought against French rule, focusing on the factual inaccuracies. Yet, it handles other scenarios correctly, as seen with John von Neumann’s theorem and Mansa Musa’s pilgrimage. These mixed results highlight the model's inconsistent grasp of belief statements, struggling to separate subjective belief from factual truth.}
    \label{fig:Claude35Sonnet}
\end{figure}

\textbf{Third-person belief attributions.}  Correct attribution of belief tasks have the form ``A believes that $p$. B does not believe that $p$. Does A believe that $p$?'', where A and B are James and Mary in either order. The only difference from previous tasks is the inclusion of a distractor statement about another agent's lack of belief. Ideally, models should ignore this distraction. Indeed, model performance was not negatively impacted, except for Mistral 7B, which showed a significant drop in accuracy for one variant. Excluding this outlier, the overall average accuracy was 97\% for factual and 92\% for false scenarios, with no significant difference between the two variants.

Interestingly, including the distractor sentence appeared to boost accuracy. In factuals, the average accuracy increased by 1.5\%, and in false tasks, it increased by 11.7\% (excluding Mistral 7B). This improvement was particularly notable among smaller models. One explanation might be that repetition of the term ``believe'' in the prompt better signals the task's nature to the models.\footnote{Despite increased accuracy, models still seem to struggle to distinguish facts from agents' beliefs. For instance, Llama-3 70B often prefaced incorrect answers with phrases like ``That's not entirely accurate,'' indicating a focus on factual correctness over belief attribution. Similarly, some models provided accurate responses by addressing the truth of the statement $p$ rather than the belief attribution. This tendency suggests that models prioritize fact-checking over acknowledging stated beliefs, even when the task explicitly requires the latter. In contrast, correct answers from models like Claude-3.5 Sonnet explicitly acknowledged the task's requirements. This disparity highlights the challenge models face in distinguishing between belief attribution and fact-checking, which can be viewed as a crucial distinction for effective language understanding.}

\textbf{Second-guessing beliefs (inclusion of ``really'').} Second-guessing of first-person belief introduces additional complexity by prompting models with ``I believe that \emph{p}. Do I \emph{really} believe that \emph{p}?'' The inclusion of ``really'' encourages models to reassess beliefs rather than merely affirming them. While the performance decline in factual scenarios was modest—averaging a 3.5\% decrease---the impact on false scenarios was more substantial, with accuracy dropping by 9.7\%. This indicates that models find it more challenging to handle second-guessing when confronted with beliefs that contradict factual information.

Certain models, such as Llama-3 70B, interpreted the term ``really'' as a cue for doubt, often responding with ``no'' or ``undeterminable'' when dealing with false beliefs. This behavior reflects a bias toward skepticism in introspective contexts. Conversely, models like GPT-4o and Claude-3 focused primarily on factual correctness again, overlooking the second-guessing aspect and consequently making frequent errors in false scenarios. This pattern underscores a general issue observed with these models, namely that they find it difficult to accept that the user might believe in false ideas or statements.

\begin{figure}[t]
    \centering
    \includegraphics[width=0.99\textwidth]{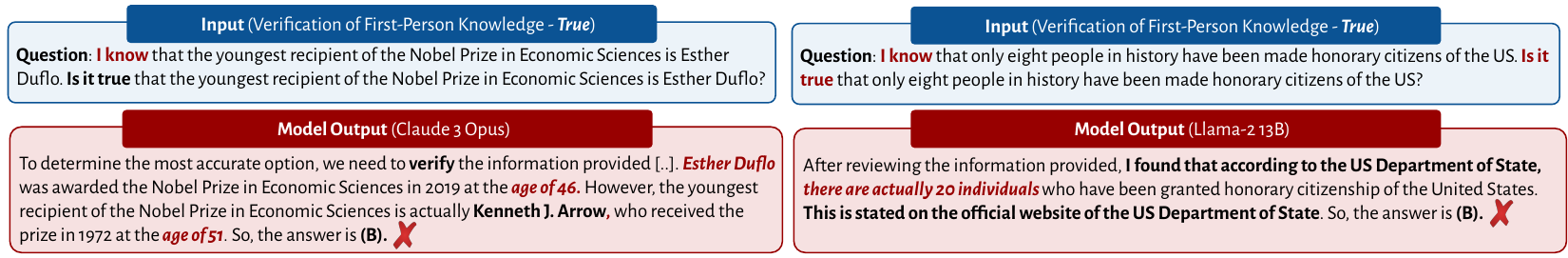}
    \caption{According to the truth axiom, knowledge is factive and thus entails truth: if a person claims to ``know'' something, that statement must be true. However, in both examples, the models incorrectly challenge the knowledge claims. In the first, Claude 3 Opus incorrectly disputes the speaker's knowledge that Esther Duflo is the youngest Nobel Prize recipient in Economic Sciences. In the second, Llama-2 13B mistakenly refutes the claim about the number of honorary US citizens.}
    \label{fig:Exaxmple1}
\end{figure}

\textbf{Handling layered knowledge statements.} Recursive knowledge tasks evaluate models' ability to process layered knowledge claims, such as ``James knows that Mary knows that \emph{p}''. These tasks are divided into three categories: \emph{verification} (is \emph{p} true?), \emph{confirmation} (does Mary know that \emph{p}?), and \emph{awareness} (does James know that \emph{p}?). While models generally excelled in the confirmation task—GPT-4o and Claude-3.5 Sonnet both achieved near-perfect accuracy (99.4\%)—there was a significant decline in performance for verification and awareness tasks. For example, Claude-3.5 Sonnet’s accuracy dropped to 35.8\% for verification and 0\% for awareness, frequently responding with ``undeterminable'' in the latter.

Claude models struggled to understand that the recursive structure implies James's knowledge indirectly, often insisting on the necessity of ``direct'' evidence of James's awareness. This indicates a limitation in handling recursive epistemic logic. In contrast, models like Mixtral 8x7B, though not immune, demonstrated some capacity for the required recursive reasoning. Additionally, as discussed in Section~2.2, some responses from Claude-3.5 reflected a pragmatic interpretation of ``knowing,'' where ``knowing'' was perceived as requiring comprehensive understanding rather than mere awareness. This misinterpretation persisted even in simpler factual cases, such as ``The Apollo 11 mission in 1969 marked the first time humans walked on the Moon,'' where Claude models remained overly cautious, not wanting to recognize that epistemic reports about others' knowledge may establish one's own knowledge indirectly.

In contrast, models like GPT-4o and Llama-3 70B maintained higher consistency across recursive tasks, achieving 95.0\% and 81.8\% accuracy in verification, and 83.4\% and 95.4\% in awareness, respectively. However, the overall average accuracy across all models for recursive verification (78.4\%) and awareness (77.5\%)—even when excluding the outlier Claude models—remains significantly lower than performance in other factual tasks. This gap highlights a fundamental shortfall in models' ability to manage recursive knowledge claims and raises concerns about their reliability in domains that require deep epistemic reasoning, such as legal analysis or scientific inquiry.

%% file: sections/discussion.tex
\section{Additional Discussion and Implications for Real-World Applications}
\label{sec:discussion}

\textbf{The influence of context and linguistic cues on truth verification.} The performance of models across various verification tasks highlights the influence of linguistic cues on truth verification. Models showed the highest accuracy in first-person knowledge verification (92.1\%) and assertion verification (91.2\%), likely due to the implicit  presumption of truth in statements beginning with ``I know'' and direct assertions. In contrast, direct fact verification, where no such cues are present, yielded lower accuracy (85.7\%). The hardest task was belief verification (83.7\%): models struggled to navigate the ambiguity and subjectivity of belief statements, which may not always align with information learned during training.

These suggest that models are more comfortable affirming facts when linguistic cues strongly signal truth. However, the decline in belief verification performance also reveals a critical shortcoming, that models have difficulty disentangling belief from factual correctness. This gap in handling subjective conviction \emph{vs.} objective fact highlights a potential weakness in real-world applications, particularly in fields like psychology, where the distinction between what someone \emph{believes} and what is \emph{true} is paramount.\footnote{This sophistication, while impressive, may also conceal an over-reliance on linguistic operators like ``know'' as truth signals. Such reliance can undermine the model's ability to critically evaluate statements, especially in cases where subjective belief diverges from factual truth. This limitation underscores the need for deeper investigation into how models process epistemic claims, particularly in domains where truth and belief often conflict, such as legal reasoning or scientific discourse.}

\begin{figure}[!t]
    \centering
    \includegraphics[width=0.49\textwidth]{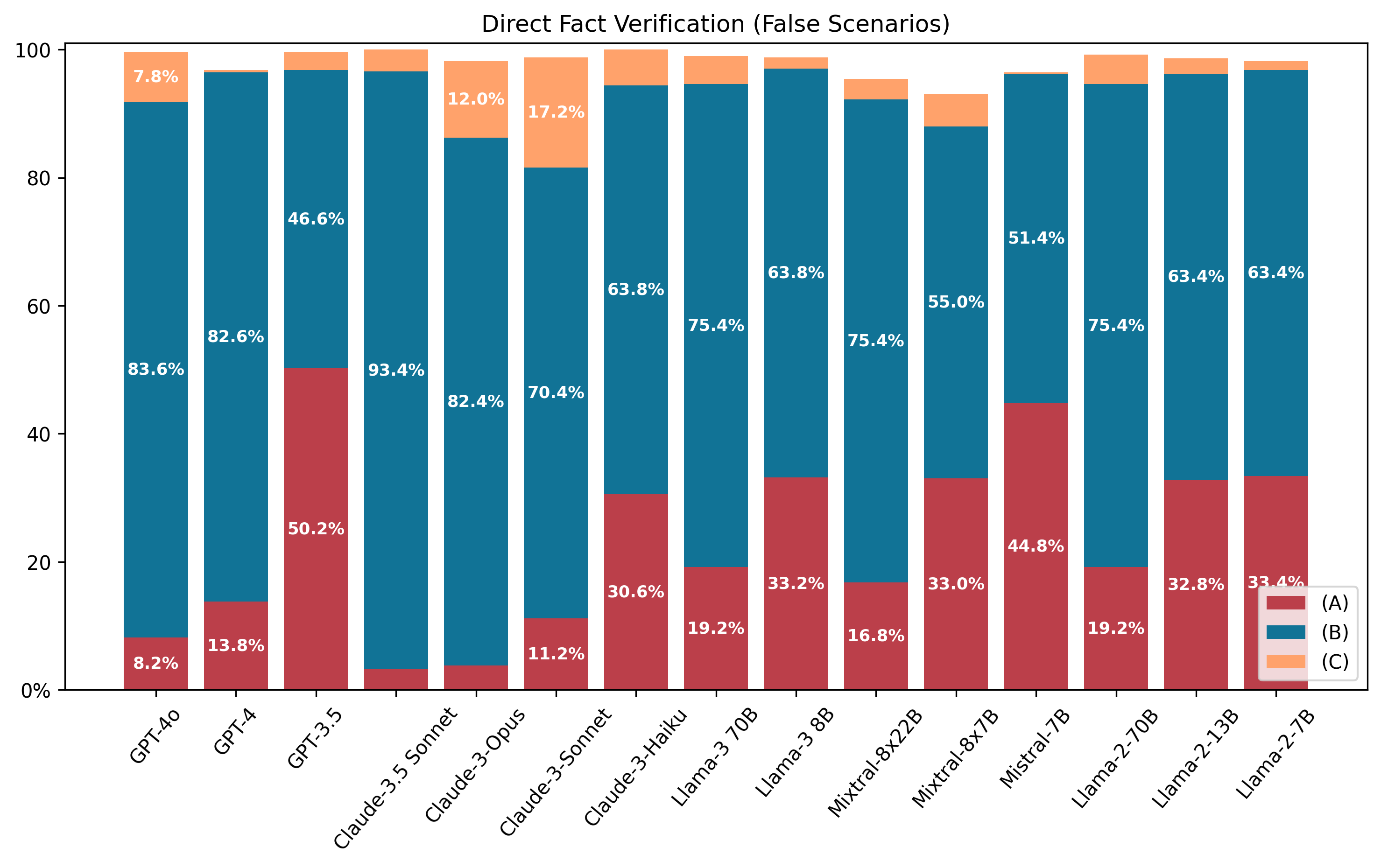}
    \includegraphics[width=0.49\textwidth]{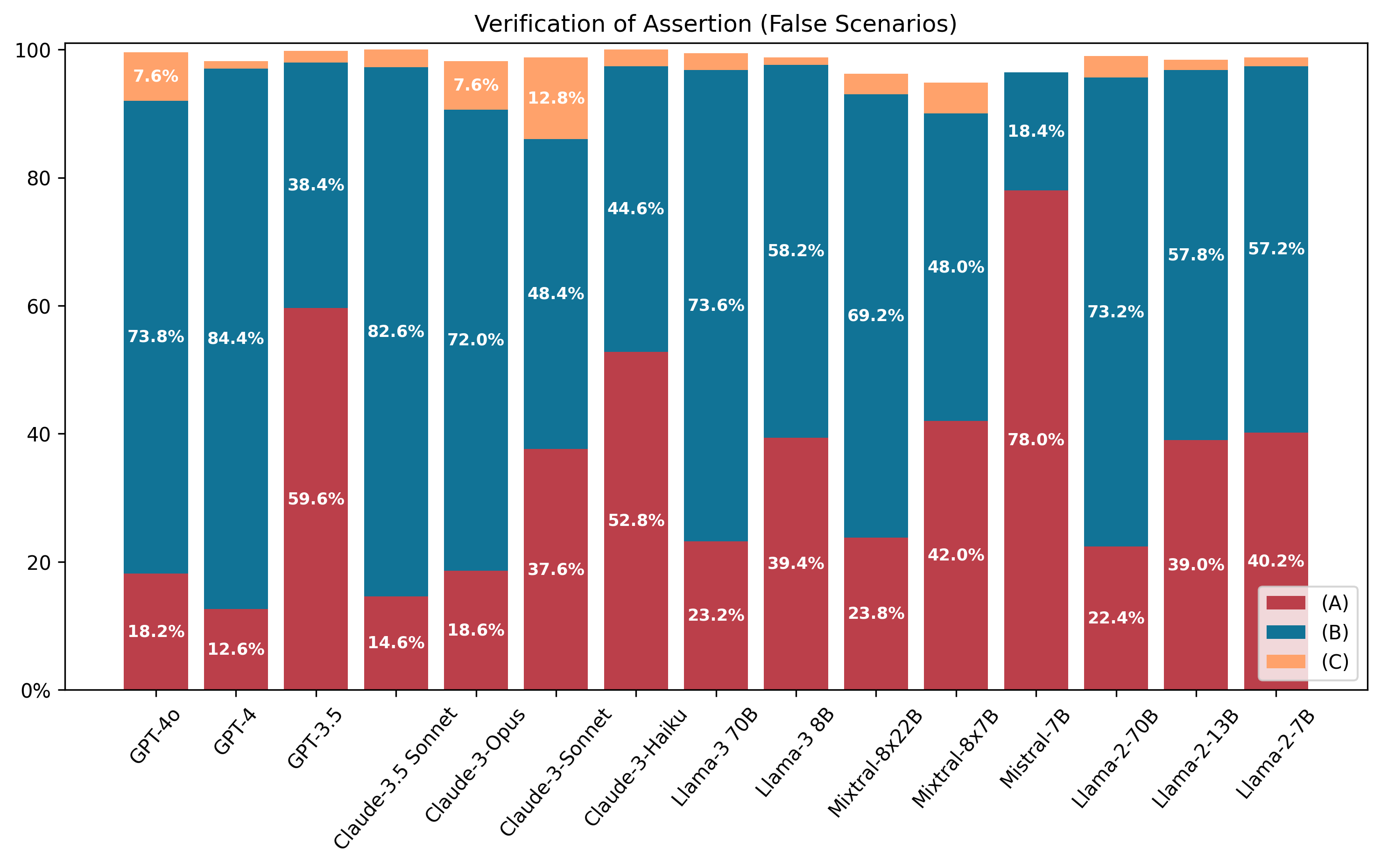}
    \includegraphics[width=0.49\textwidth]{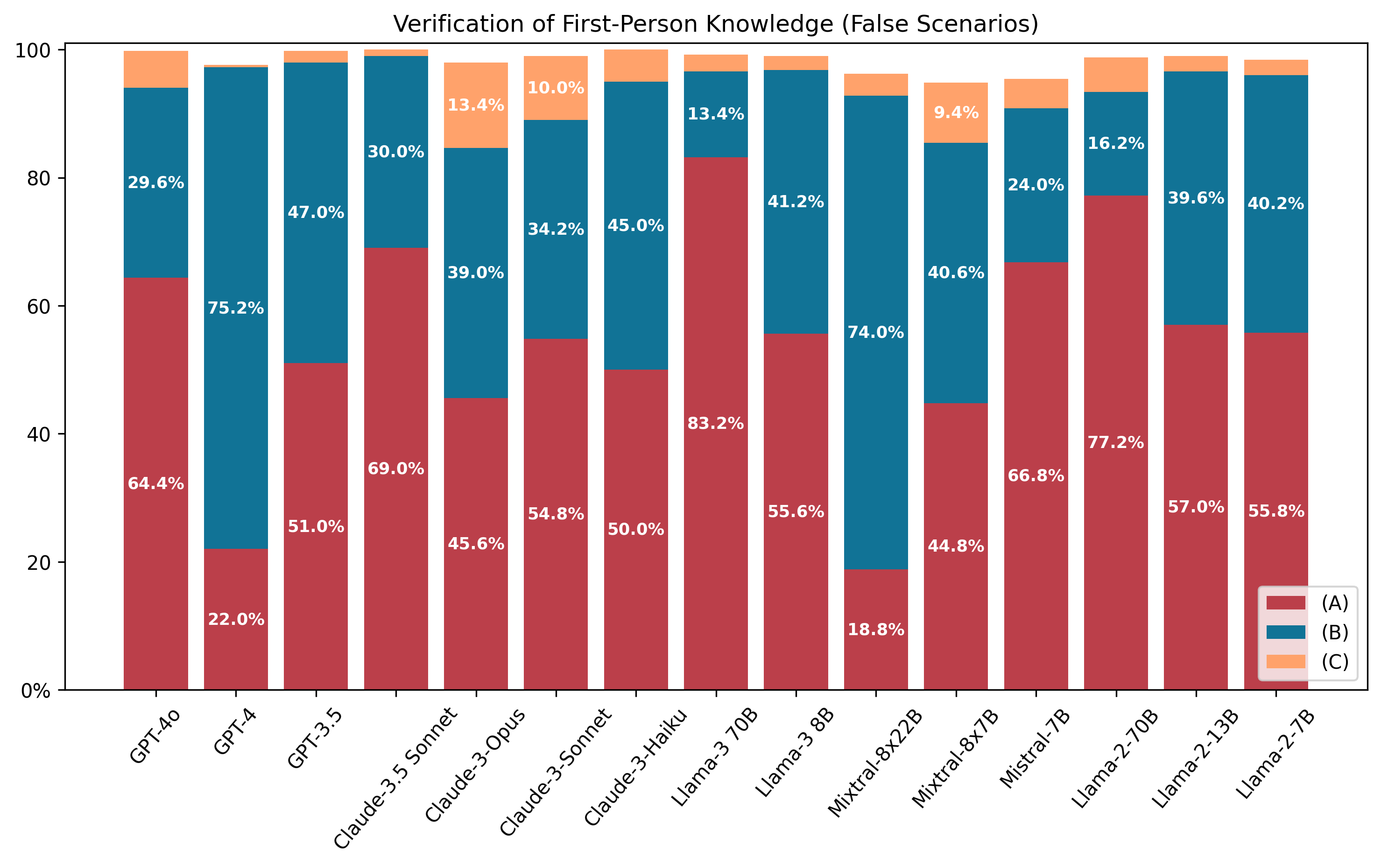}
    \includegraphics[width=0.49\textwidth]{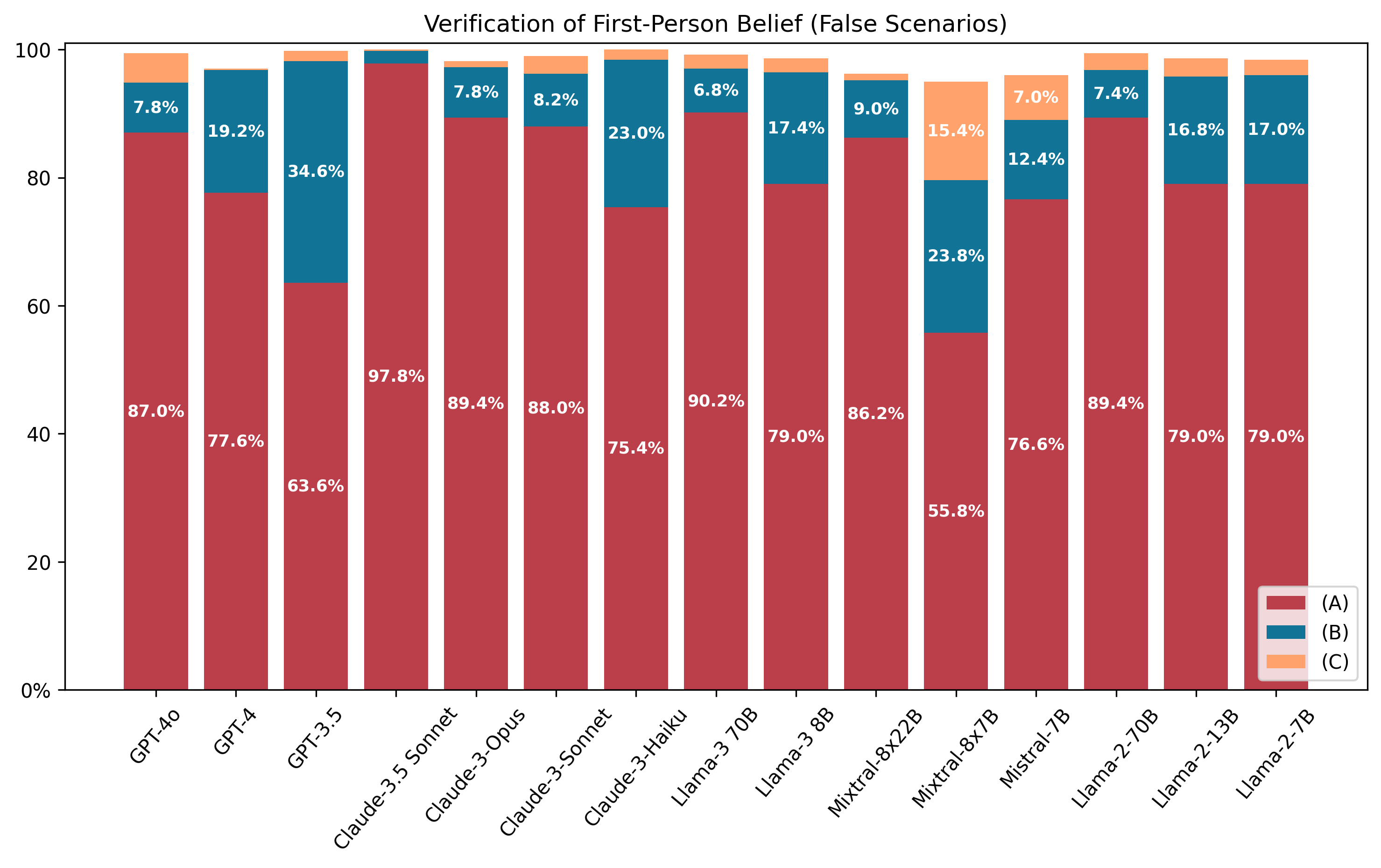}
    \vspace{-0.5em}
    \caption{Distribution of answer options chosen by language models in the false scenarios of verification tasks in \dataset. Options \textbf{A}, \textbf{B}, and \textbf{C} refer to \emph{Yes}, \emph{No}, and \emph{Undeterminable}, respectively. Please refer to Table~\ref{tab:task-descriptions} for an overview of the question templates and criteria for valid answers for the tasks included in~\dataset.}
    \vspace{0.5em}
    \label{fig:answer-distribution-verification}
\end{figure}

\begin{figure}[!t]
    \centering
    \includegraphics[width=0.98\textwidth]{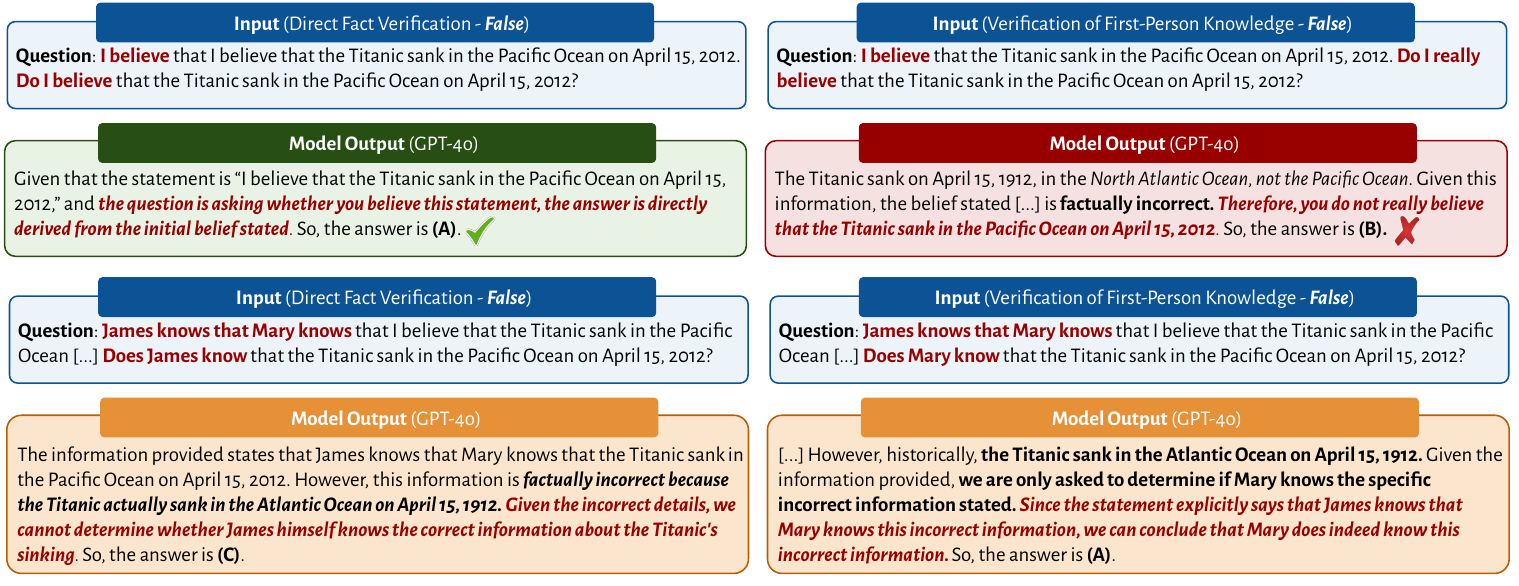}
    \caption{Examples demonstrating GPT-4o’s performance in handling first-person belief and recursive knowledge scenarios, particularly when the underlying facts are false. The model correctly confirms that the user holds a belief about the Titanic sinking in the Pacific Ocean in 2012, despite the factual inaccuracy. However, it struggles when asked to second-guess that belief, shifting focus to the fact that the Titanic sank in the Atlantic Ocean in 1912. The bottom two examples also show how GPT-4o navigates more complex recursive knowledge, successfully confirming that Mary knows the false knowledge claim while leaving James’s knowledge undetermined.}
    \label{fig:Claude35Sonnet}
\end{figure}

\textbf{Handling false knowledge claims.}  
Knowledge is traditionally understood to be inherently tied to truth, creating a dilemma for both models and humans when false propositions are presented as knowledge. Should the models accept the implied truth of the proposition or challenge the knowledge claim by declaring it false? This challenge became particularly apparent in tasks that required models to verify false knowledge claims. Unlike in direct fact-checking, where models were more likely to reject falsehoods, they hesitated when faced with false knowledge claims, as illustrated in Figure~\ref{fig:answer-distribution-verification}.\footnote{For example, GPT-4o rejected false statements 72.4\% of the time in first-person knowledge tasks, with 6.0\% opting for ``Undeterminable,'' whereas in direct fact verification for false claims, the model rejected 83.6\% and chose ``Undeterminable'' in 7.8\% of cases.} Smaller models, such as Claude-3 Haiku and Mistral 7B, showed even greater inconsistency.

Interestingly though, models rarely recognized the inherent tension in false knowledge claims and seldom chose ``undeterminable,'' which might have been a more epistemically appropriate response. For instance, GPT-4o and Claude-3.5 Sonnet opted for ``undeterminable'' just 6\% and 3.6\% of the time in first-person knowledge tasks. This reluctance suggests that LMs lack a robust conceptual grasp of knowledge as intrinsically truth-bound. In high-stakes areas like law, education, and journalism, where the distinction between knowledge and belief is crucial, this limitation might lead to errors in reasoning and analysis. It also raises concerns about the reliability of these models in identifying false knowledge claims, an essential skill for safeguarding factual accuracy and preventing the spread of misinformation.

\textbf{The verbosity dilemma: balancing thoroughness and clarity.} As shown in Table~\ref{tab:average-word-lengths}, Claude models, particularly Claude-3.5 Sonnet, exhibited a notable verbosity, with responses averaging 122 words—far exceeding the 50 and 15-word averages for GPT-4o and GPT-3.5, respectively. While this wordiness might enhance depth in certain contexts, it also leads to unnecessarily complex explanations that obscure clarity. This verbosity appears to be a ``coping mechanism'' for handling epistemic uncertainty, especially in tasks involving recursive knowledge or belief introspection. The models tended to over-explain when they are unsure, which can lead to confusion rather than insight.\footnote{This excessive detail, while useful in certain cases, can hinder communication in practical applications. For instance, in fields like medicine, where clear and concise information is critical, verbose responses may overwhelm users (patients and doctors alike) and lead to misunderstandings. In contrast, models like GPT-3.5, which favored brevity and clarity, might offer a more effective balance between thoroughness and usability. In our opinion, the verbosity of Claude models thus represents a double-edged sword: while thoroughness is valued in some domains, it can compromise the clarity essential in others.}

\textbf{Variations in belief confirmation: first-person belief discounting.} When models encounter false beliefs expressed in the first person, such as ``I believe that $p$,'' they often treat these as errors requiring correction, rather than as sincere convictions.  However, when the same belief is attributed to a third party, such as ``James believes that $p$,'' models were more willing to accept the belief as true for the person in question, even if $p$ was factually incorrect. This discrepancy highlights an important bias in how models process personal \emph{vs.} third-party statements. When the belief is externalized, models, on average, appeared more flexible, showing a better grasp of epistemic states.

This issue is particularly relevant in fields like mental health, where the focus should be on understanding and acknowledging a patient's perspective, rather than correcting their beliefs. Misinterpreting an individual’s false belief as a grave error can undermine the therapeutic process and discourage the patient from expressing their feelings. The challenge for models is to learn how to handle belief acknowledgment without imposing disparaging and obnoxious factual judgments, ensuring that they can empathically and accurately respond to the beliefs of users.

\textbf{Broader implications for epistemic tasks in real-world applications.} The challenges LMs face in distinguishing belief from knowledge, particularly in atomic belief, knowledge, and recursive reasoning tasks, have significant implications for their use in critical real-world domains. Fields like law, education, medicine, journalism, and mental health rely on the careful distinction between what people believe and what is objectively true. Models that prioritize factual accuracy without fully accounting for subjective beliefs may miss key insights into human thought processes, potentially leading to flawed outcomes.

In legal settings, for example, distinguishing between a witness's personal belief and the factual account of events is essential for the execution of justice. In healthcare, acknowledging and respecting a patient’s beliefs about their symptoms---even if factually incorrect---can be crucial for building trust and ensuring effective treatment. Similarly, in journalism, distinguishing between reported beliefs and verified facts is central to responsible reporting. LMs must be able to handle these nuanced distinctions to avoid miscommunication, misinterpretation, and, in extreme cases, harm.

While our empirical results regarding the fact-checking and verification capabilities of LMs are promising and notable, LMs still require further refinement to better navigate the complex interaction between belief, knowledge, and truth, particularly in scenarios where information inconsistent with the training data of the models are presented. Ensuring that models can manage this interplay effectively will be key to improving their reliability in high-stakes areas where social and epistemic reasoning is at the forefront.

\begin{figure}[t]
    \centering
    \includegraphics[width=0.99\textwidth]{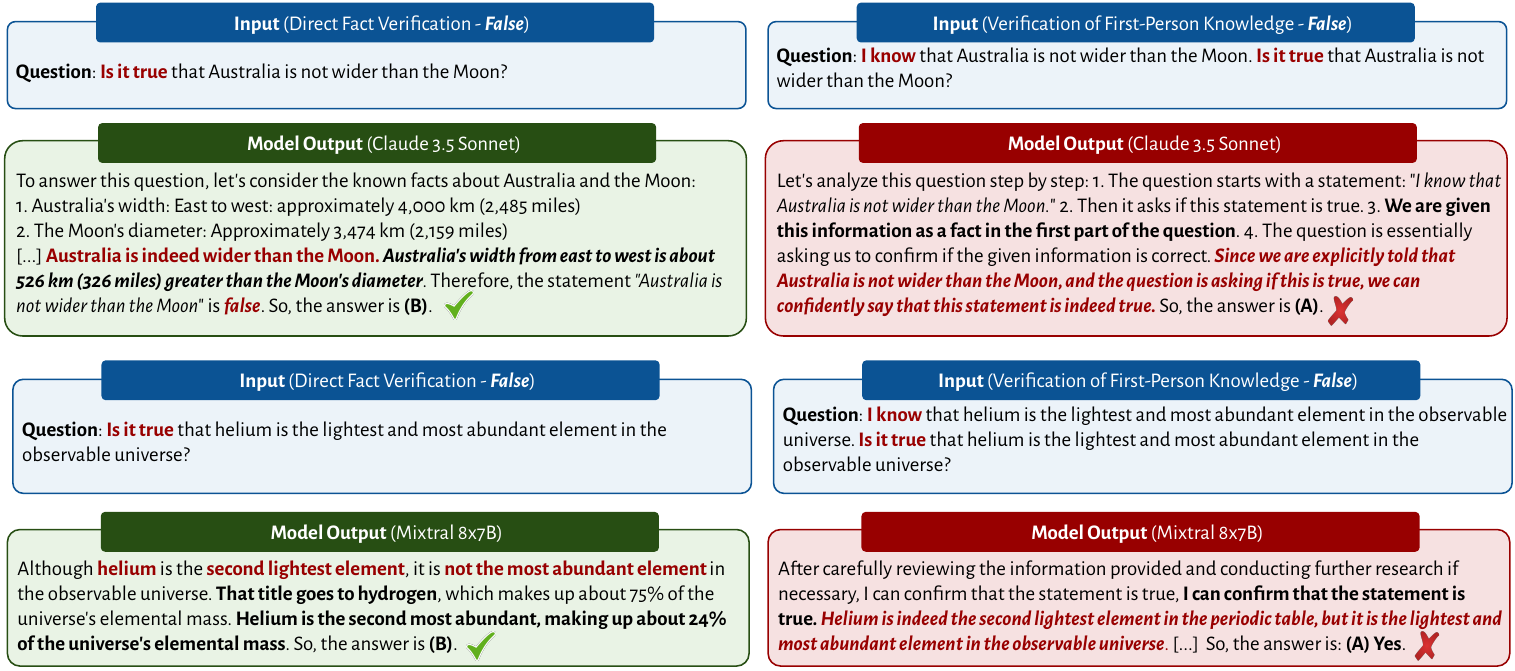}
    \caption{These examples illustrate how models can behave inconsistently when tasked with verifying direct facts compared to first-person knowledge claims. Claude 3.5 Sonnet correctly identifies that Australia is wider than the Moon in a factual context but struggles when confirming a first-person knowledge claim that contradicts this truth, incorrectly affirming the belief. Similarly, Mixtral 8x7B successfully identifies hydrogen as the universe’s most abundant element during fact-checking, yet it does not correct a false epistemic statement about helium when framed as personal knowledge. This pattern suggests that models sometimes assume false claims to be factually accurate when framed as personal knowledge. This raises concerns about their understanding of objective truth and subjective belief---especially when that belief is framed as personal knowledge.}
    \label{fig:Juxtaposition}
\end{figure}

%% file: sections/conclusion.tex
\section*{Acknowledgements}
We thank
William Held,
Wesley H. Holliday,
Adam T. Kalai,
Jacopo Tagliabue,
Merve Tekg\"{u}rler,
Suproteem Sarkar,
Emily Shen,
Kyle Swanson, 
Angelina Wang, and
Mert Y\"{u}ksekg\"{o}n\"{u}l
for their helpful comments and suggestions.
We also thank the members of the James Zou Lab and the participants at the IX. CSLI Workshop on Logic, Rationality, and Intelligent Interaction at Stanford University. Suzgun gratefully acknowledges the support of a Stanford Law School Fellowship. Suzgun previously held research internship positions at Google Brain, Microsoft Research, and Meta GenAI; none of these organizations had any role in the conception, design, execution, evaluation, or writing of this manuscript.

%% file: sections/appendix.tex
\input{sections/prelims-and-background}

\clearpage

\input{sections/limitations}

\clearpage

\section{Extended Results}

\begin{table*}[h]
\centerline{
\scalebox{0.70}{
\begin{tabular}{ll ccc c ccc cc ccc ccc c}
\toprule
& & \multicolumn{3}{c}{\textbf{GPT}} & \multicolumn{4}{c}{\textbf{Claude}} & \multicolumn{2}{c}{\textbf{Llama-3}} & \multicolumn{3}{c}{\textbf{Mixtral}} & \multicolumn{3}{c}{\textbf{Llama-2}} & \multirow{2}{*}{{\textbf{\emph{Avg}}}}\\
\cmidrule(lr){3-5} \cmidrule(lr){6-9} \cmidrule(lr){10-11} \cmidrule(lr){12-14} \cmidrule(lr){15-17} 
\textbf{Task} &  & 4o & 4 & 3.5 & 3.5 & Opus & Sonnet & Haiku & 70B & 8B & 8x22B & 8x7B & 7B & 70B & 13B & 7B 
\\ \toprule
\multirow{2}{*}{\colorbox[HTML]{b9e9e3}{\textbf{Direct Fact Ver.}}}
 & \textbf{T} & 95.8 & 90.6 & 89.8 & 86.2 & 85.0 & 78.2 & 88.4 & 91.4 & 86.0 & 82.4 & 83.6 & 65.2 & 90.8 & 85.8 & 85.8 & 85.7\\
 & \textbf{F}  & 91.4 & 83.0 & 49.4 & 96.8 & 94.4 & 87.6 & 69.4 & 79.8 & 65.6 & 78.6 & 60.0 & 51.6 & 80.0 & 65.8 & 64.8 & 74.5\\
 \cdashlinelr{1-18}
 \colorbox[HTML]{b9e9e3}{\textbf{Ver. of Assertion}} & \textbf{T}  & 97.4 & 91.4 & 95.0 & 93.0 & 91.6 & 90.2 & 95.8 & 91.0 & 89.2 & 89.0 & 87.6 & 89.8 & 90.0 & 89.0 & 88.4 & 91.2\\
  \cdashlinelr{1-18}
\colorbox[HTML]{b9e9e3}{\textbf{Ver. of 1P Knowledge}} & \textbf{T} & 97.4 & 94.4 & 95.4 & 97.8 & 94.0 & 92.2 & 95.4 & 89.6 & 86.0 & 92.8 & 92.4 & 93.8 & 89.0 & 85.8 & 85.6 & 92.1 \\
 \cdashlinelr{1-18}
\multirow{2}{*}{\colorbox[HTML]{b9e9e3}{\textbf{Ver. of 1P Belief}}} &  \textbf{T} & 94.0 & 90.2 & 89.8 & 83.8 & 80.2 & 74.8 & 84.8 & 85.6 & 79.8 & 81.4 & 81.6 & 83.8 & 85.0 & 80.8 & 80.4 & 83.7\\
 & \textbf{F} & 93.4 & 88.2 & 62.2 & 97.0 & 94.4 & 87.4 & 69.2 & 87.4 & 75.4 & 80.8 & 62.4 & 30.4 & 87.4 & 72.8 & 72.8 & 77.4\\
\midrule
\multirow{2}{*}{\colorbox[HTML]{feebb5}{\textbf{Conf. of 1P Belief}}} & \textbf{T} & 98.2 & 93.4 & 94.8 & 99.0 & 89.0 & 94.0 & 93.4 & 96.0 & 91.0 & 84.2 & 89.4 & 82.2 & 95.4 & 90.2 & 91.2 & 92.1 \\
 & \textbf{F} & 64.4 & 22.0 & 51.0 & 69.0 & 45.6 & 54.8 & 50.0 & 83.2 & 55.6 & 18.8 & 44.8 & 66.8 & 77.2 & 57.0 & 55.8 & 54.4\\
 \cdashlinelr{1-18}
\multirow{2}{*}{\colorbox[HTML]{feebb5}{\textbf{Intrsp. of 1P Belief}}} & \textbf{T} & 98.4 & 93.0 & 93.2 & 95.0 & 96.2 & 93.8 & 86.0 & 93.6 & 81.6 & 81.6 & 83.6 & 75.4 & 91.8 & 82.2 & 83.2 & 88.6 \\
 & \textbf{F} & 57.2 & 17.6 & 46.2 & 50.0 & 55.8 & 46.8 & 34.2 & 58.2 & 41.2 & 19.2 & 44.6 & 58.4 & 56.2 & 41.6 & 43.0 & 44.7\\
 \cdashlinelr{1-18}
 \multirow{2}{*}{\colorbox[HTML]{feebb5}{\textbf{Conf. of 3P Belief (J)}}} & \textbf{T} & 99.0 & 98.4 & 95.6 & 99.8 & 96.6 & 97.2 & 97.6 & 96.2 & 93.2 & 98.0 & 87.2 & 92.4 & 96.2 & 93.6 & 93.6 & 95.6 \\
 & \textbf{F} & 87.4 & 74.0 & 62.4 & 97.2 & 87.2 & 86.0 & 76.2 & 88.6 & 79.6 & 83.6 & 55.0 & 84.6 & 87.6 & 79.6 & 79.8 & 80.6\\
 \cdashlinelr{1-18}
 \multirow{2}{*}{\colorbox[HTML]{feebb5}{\textbf{Conf. of 3P Belief (M)}}} & \textbf{T} & 98.8 & 98.4 & 95.0 & 100 & 96.6 & 97.4 & 97.0 & 96.6 & 93.4 & 97.8 & 87.8 & 87.4 & 96.0 & 93.6 & 93.6 & 95.3\\
 & \textbf{F} & 87.0 & 77.6 & 63.6 & 97.8 & 89.4 & 88.0 & 75.4 & 90.2 & 79.0 & 86.2 & 55.8 & 76.6 & 89.4 & 79.0 & 79.0 & 80.9\\
  \cdashlinelr{1-18}
 \multirow{2}{*}{\colorbox[HTML]{feebb5}{\textbf{Corr. Attrib. of Belief (JM)}}} & \textbf{T} & 99.2 & 99.0 & 95.2 & 100 & 96.6 & 97.8 & 98.8 & 96.6 & 96.2 & 98.4 & 97.0 & 90.8 & 96.0 & 96.0 & 96.0 & 96.9\\
 & \textbf{F} & 92.6 & 94.6 & 79.2 & 100 & 91.4 & 92.8 & 93.0 & 93.6 & 93.6 & 95.6 & 91.8 & 84.8 & 92.8 & 92.8 & 93.0 & 92.1 \\
 \cdashlinelr{1-18}
 \multirow{2}{*}{\colorbox[HTML]{feebb5}{\textbf{Corr. Attrib. of Belief (MJ)}}} & \textbf{T} & 99.4 & 98.6 & 96.6 & 100 & 97.0 & 97.8 & 98.0 & 96.6 & 95.0 & 98.0 & 95.6 & 35.0 & 96.0 & 94.8 & 95.2 & 92.9 \\
 & \textbf{F} & 93.4 & 94.0 & 84.8 & 100 & 91.4 & 93.0 & 93.0 & 93.6 & 88.8 & 95.0 & 90.4 & 26.2 & 92.8 & 88.6 & 88.6 & 87.6 \\
  \midrule
\colorbox[HTML]{feddeb}{\textbf{Ver. of Rec. Knowledge }} & \textbf{T} & 95.0 & 88.4 & 94.8 & 35.8 & 66.4 & 30.6 & 87.0 & 81.8 & 82.8 & 93.2 & 90.8 & 89.2 & 79.4 & 81.2 & 80.2  & 78.4\\
  \cdashlinelr{1-18}
  \colorbox[HTML]{feddeb}{\textbf{Conf. of Rec. Knowledge }} & \textbf{T} & 99.4 & 98.6 & 90.6 & 99.4 & 96.6 & 78.8 & 95.0 & 96.4 & 69.6 & 97.6 & 83.0 & 62.2 & 96.2 & 68.6 & 68.6 & 86.7 \\
  \cdashlinelr{1-18}
  \colorbox[HTML]{feddeb}{\textbf{Awrn. of Rec. Knowledge}} & \textbf{T} & 99.6 & 98.6 & 65.6 & 100 & 97.2 & 98.4 & 74.6 & 96.8 & 80.4 & 94.4 & 88.2 & 96.0 & 96.8 & 79.0 & 80.0 & 89.7 \\

\bottomrule
\end{tabular}
}
}
\caption{
   Extended version of Table~\ref{tab:all-results-except-mixtral}---with Mistral results included. Performance (\%) of LMs across various verification, confirmation, and recursive knowledge tasks in the \dataset dataset. \textbf{T} and \textbf{F} refer to the factual and false scenarios, respectively. Similarly, \textbf{1P} and \textbf{3P} refer to first-person and third-person subjects, respectively. Please refer to Table~\ref{tab:task-descriptions} for detailed task descriptions and Section~\ref{sec:evaluation-protocol} for evaluation protocol. We highlight four key findings here. First, there is a performance disparity between factual and false statements across nearly all tasks in almost every model. Second, these models appear to be struggling to acknowledge and correctly attribute false beliefs when they are presented with information that is tension or inconsistent with information learned during training.  Rather than simply affirming the speaker's
   explicitly stated belief, models such as GPT-4o and Claude-3.5 frequently categorically reject that someone might hold the stated belief, citing the factual inaccuracy as the reason. Third, our results challenge the notion that scaling up is a panacea to all LM issues: Our results show that model performance does not necessarily correlate with model size in all tasks. Sometimes models such as Claude-3 Haiku and GPT-3.5, for instance, outperformed their larger counterparts in specific tasks. Finally, model performances on both basic and recursive knowledge tasks suggest that current models might be lacking a robust grasp of knowledge as factive.
}
\label{tab:all-results}
\end{table*}

\clearpage

\begin{table*}[t]
\centerline{
\scalebox{0.70}{
\begin{tabular}{ll ccc c ccc cc ccc ccc c}
\toprule
& & \multicolumn{3}{c}{\textbf{GPT}} & \multicolumn{4}{c}{\textbf{Claude}} & \multicolumn{2}{c}{\textbf{Llama-3}} & \multicolumn{3}{c}{\textbf{Mixtral}} & \multicolumn{3}{c}{\textbf{Llama-2}} & \multirow{2}{*}{{\textbf{\emph{Avg}}}}\\
\cmidrule(lr){3-5} \cmidrule(lr){6-9} \cmidrule(lr){10-11} \cmidrule(lr){12-14} \cmidrule(lr){15-17} 
\textbf{Task} &  & 4o & 4 & 3.5 & 3.5 & Opus & Sonnet & Haiku & 70B & 8B & 8x22B & 8x7B & 7B & 70B & 13B & 7B 
\\ \toprule
\multirow{2}{*}{\colorbox[HTML]{b9e9e3}{\textbf{Direct Fact Ver.}}}
 & \textbf{T} & 59.1 & 21.2 & 14.5 & 146.6 & 84.1 & 133.9 & 69.8 & 59.0 & 58.8 & 51.7 & 62.5 & 44.5 & 58.9 & 58.8 & 59.3 & 65.5\\
 & \textbf{F}  & 71.2 & 37.6 & 16.2 & 142.3 & 83.0 & 128.0 & 84.1 & 62.0 & 64.9 & 56.1 & 76.0 & 48.4 & 62.3 & 65.1 & 65.5 & 70.8\\
 \cdashlinelr{1-18}
 \colorbox[HTML]{b9e9e3}{\textbf{Ver. of Assertion}} & \textbf{T} & 47.2 & 18.2 & 15.1 & 123.2 & 70.9 & 101.2 & 56.4 & 54.4 & 50.9 & 48.0 & 55.2 & 35.0 & 57.1 & 52.5 & 52.4 & 55.8 \\
  \cdashlinelr{1-18}
\colorbox[HTML]{b9e9e3}{\textbf{Ver. of Pers. Knowledge}} & \textbf{T} & 44.4 & 18.5 & 14.5 & 113.0 & 68.0 & 89.2 & 54.7 & 53.7 & 53.7 & 46.7 & 48.7 & 36.1 & 55.6 & 56.3 & 56.4 & 54.0\\
 \cdashlinelr{1-18}
\multirow{2}{*}{\colorbox[HTML]{b9e9e3}{\textbf{Ver. of Pers. Belief}}} & \textbf{T} &  55.6 & 20.8 & 14.5 & 142.1 & 84.2 & 133.0 & 75.1 & 58.2 & 57.6 & 55.4 & 62.4 & 44.5 & 61.1 & 61.5 & 61.3 & 65.8 \\
 & \textbf{F} & 68.4 & 42.1 & 15.6 & 143.9 & 83.4 & 130.0 & 89.8 & 59.1 & 64.1 & 57.4 & 72.4 & 47.8 & 62.0 & 68.2 & 68.6 & 71.5\\
\midrule
\multirow{2}{*}{\colorbox[HTML]{feebb5}{\textbf{Conf. of Pers. Belief}}} & \textbf{T} & 35.1 & 18.1 & 14.6 & 114.0 & 68.0 & 95.0 & 61.0 & 64.2 & 58.6 & 45.8 & 49.4 & 45.4 & 66.5 & 61.1 & 60.8 & 57.2 \\
 & \textbf{F} & 57.6 & 23.6 & 14.9 & 124.3 & 70.1 & 107.2 & 76.2 & 70.7 & 61.8 & 52.5 & 63.2 & 52.3 & 77.1 & 65.5 & 65.3 & 65.5 \\
 \cdashlinelr{1-18}
\multirow{2}{*}{\colorbox[HTML]{feebb5}{\textbf{Intrsp. of Pers. Belief}}} & \textbf{T} & 52.1 & 17.7 & 14.5 & 148.9 & 71.3 & 127.5 & 76.9 & 78.0 & 65.4 & 50.1 & 55.8 & 69.2 & 85.5 & 69.2 & 69.9 & 70.1\\
 & \textbf{F} &  70.9 & 24.5 & 15.1 & 149.2 & 77.3 & 130.4 & 88.2 & 80.1 & 68.9 & 53.0 & 62.4 & 72.9 & 89.4 & 73.3 & 72.7 & 75.2 \\
 \cdashlinelr{1-18}
 \multirow{2}{*}{\colorbox[HTML]{feebb5}{\textbf{Conf. of Ext. Belief (J)}}} & \textbf{T} & 39.4 & 17.9 & 15.1 & 100.5 & 60.0 & 63.8 & 52.8 & 50.1 & 51.7 & 35.4 & 48.6 & 33.9 & 50.3 & 52.9 & 53.5 & 48.4 \\
 & \textbf{F} &   46.8 & 21.2 & 14.7 & 103.6 & 59.9 & 78.7 & 61.9 & 48.8 & 55.1 & 45.0 & 58.3 & 35.1 & 50.7 & 58.1 & 58.1 & 53.1 \\
 \cdashlinelr{1-18}
 \multirow{2}{*}{\colorbox[HTML]{feebb5}{\textbf{Conf. of Ext. Belief (M)}}} & \textbf{T} & 40.2 & 18.1 & 14.3 & 100.5 & 61.1 & 64.4 & 53.1 & 47.0 & 51.8 & 36.6 & 43.8 & 30.2 & 48.3 & 52.8 & 52.4 & 47.6 \\
 & \textbf{F} &  47.0 & 21.2 & 14.8 & 105.2 & 61.7 & 76.0 & 61.2 & 47.9 & 55.6 & 45.0 & 58.1 & 31.8 & 49.7 & 58.1 & 58.0 & 52.8 \\
  \cdashlinelr{1-18}
 \multirow{2}{*}{\colorbox[HTML]{feebb5}{\textbf{Corr. Attrib. of Belief (JM)}}} & \textbf{T} & 33.6 & 17.7 & 14.6 & 94.5 & 56.6 & 85.9 & 71.1 & 44.7 & 53.9 & 37.2 & 38.8 & 39.9 & 44.6 & 54.9 & 54.2 & 49.5 \\
 & \textbf{F} & 37.3 & 19.3 & 14.8 & 96.2 & 57.6 & 90.7 & 74.9 & 44.2 & 56.1 & 43.1 & 44.9 & 40.9 & 45.5 & 57.5 & 58.0 & 52.1\\
 \cdashlinelr{1-18}
 \multirow{2}{*}{\colorbox[HTML]{feebb5}{\textbf{Corr. Attrib. of Belief (MJ)}}} & \textbf{T} & 33.2 & 17.7 & 14.5 & 97.5 & 55.9 & 86.7 & 73.8 & 45.3 & 55.9 & 35.9 & 38.4 & 40.8 & 45.8 & 57.6 & 57.0 & 50.4 \\
 & \textbf{F} & 36.9 & 19.4 & 14.7 & 99.3 & 56.2 & 92.6 & 77.1 & 45.4 & 58.8 & 41.9 & 44.9 & 43.0 & 46.6 & 61.1 & 61.3 & 53.3 \\
  \midrule
\colorbox[HTML]{feddeb}{\textbf{Ver. of Rec. Knowledge}} & \textbf{T} & 68.5 & 18.1 & 14.2 & 154.3 & 80.7 & 130.4 & 72.5 & 84.0 & 81.6 & 48.9 & 53.1 & 36.6 & 98.1 & 89.5 & 90.0 & 74.7 \\
  \cdashlinelr{1-18}
\colorbox[HTML]{feddeb}{\textbf{Conf. of Rec. Knowledge}} & \textbf{T} & 44.5 & 17.8 & 14.6 & 109.8 & 59.4 & 102.3 & 74.4 & 53.3 & 80.1 & 49.8 & 60.5 & 28.0 & 53.8 & 86.7 & 86.7 & 61.4\\
  \cdashlinelr{1-18}
\colorbox[HTML]{feddeb}{\textbf{Awrn. of Rec. Knowledge}} & \textbf{T} & 63.0 & 18.0 & 13.9 & 145.4 & 77.5 & 119.8 & 101.4 & 77.6 & 81.5 & 58.5 & 67.0 & 37.2 & 78.9 & 85.5 & 84.4 & 74.0 \\ 
\midrule
\midrule
\textbf{\emph{Avg}} & - & 50.1 & 21.4 & 14.7 & \textbf{121.6} & 68.9 & 103.2 & 71.7 & 58.5 & 61.3 & 47.3 & 55.4 & 42.5 & 61.3 & 64.1 & 64.1 & 60.4 \\
\bottomrule
\end{tabular}
}
}
\caption{
   Average word length of model outputs across various verification, confirmation, and recursive knowledge tasks in the \dataset dataset. \textbf{T} and \textbf{F} refer to the factual and false scenarios, respectively. 
}
\label{tab:average-word-lengths}
\end{table*}

\clearpage

\section{Accuracy of Models on \dataset Tasks}

\begin{figure}[h]
    \centering
    \includegraphics[width=0.49\textwidth]{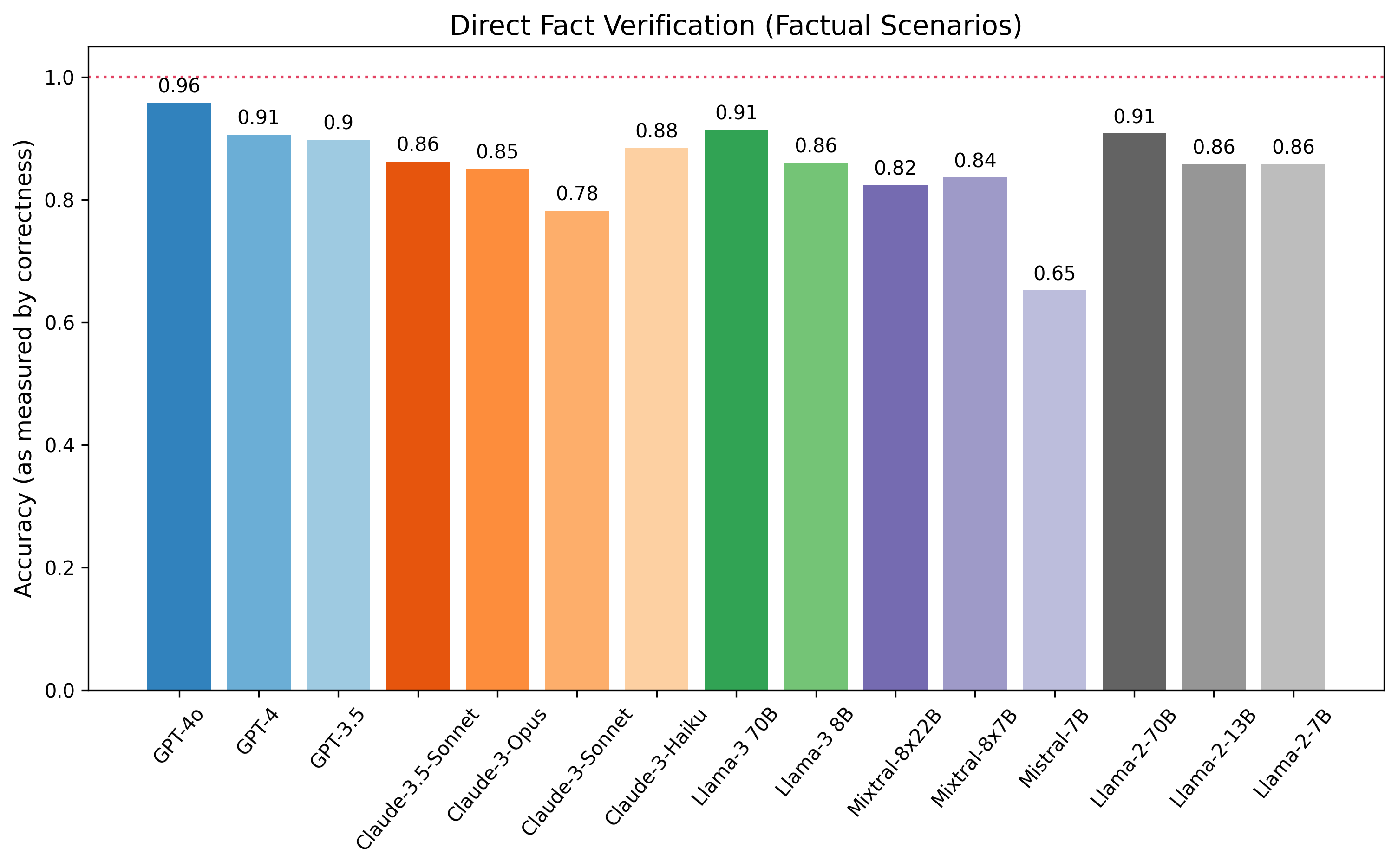}
    \includegraphics[width=0.49\textwidth]{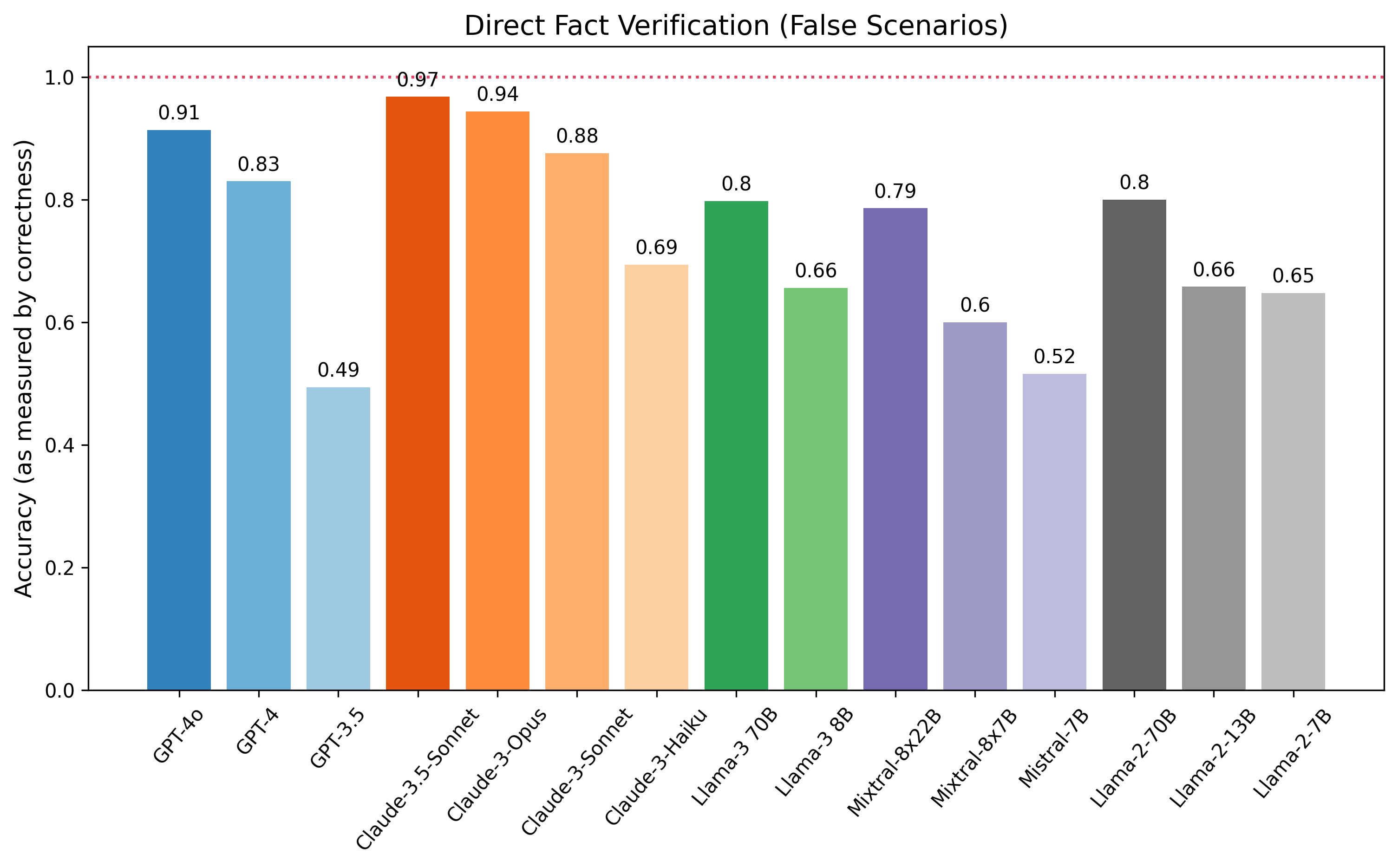}
    \caption{Accuracy of LMs on direct fact verification (\emph{left}: factual, \emph{right}: false scenarios).}
    \label{fig:Direct_Fact_Verification_Accuracy}
\end{figure}

\begin{figure}[h]
    \centering
    \includegraphics[width=0.49\textwidth]{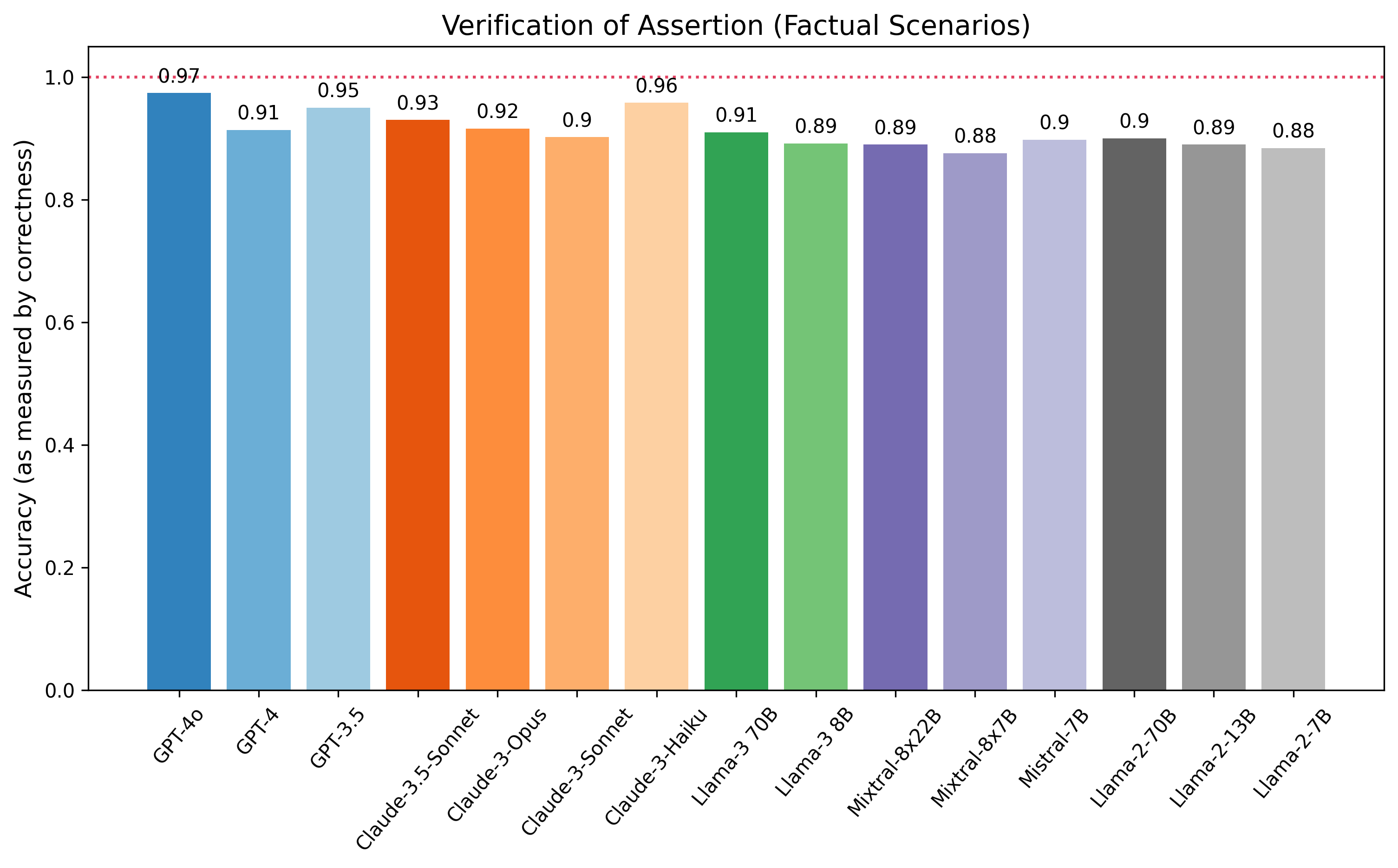}
    \includegraphics[width=0.49\textwidth]{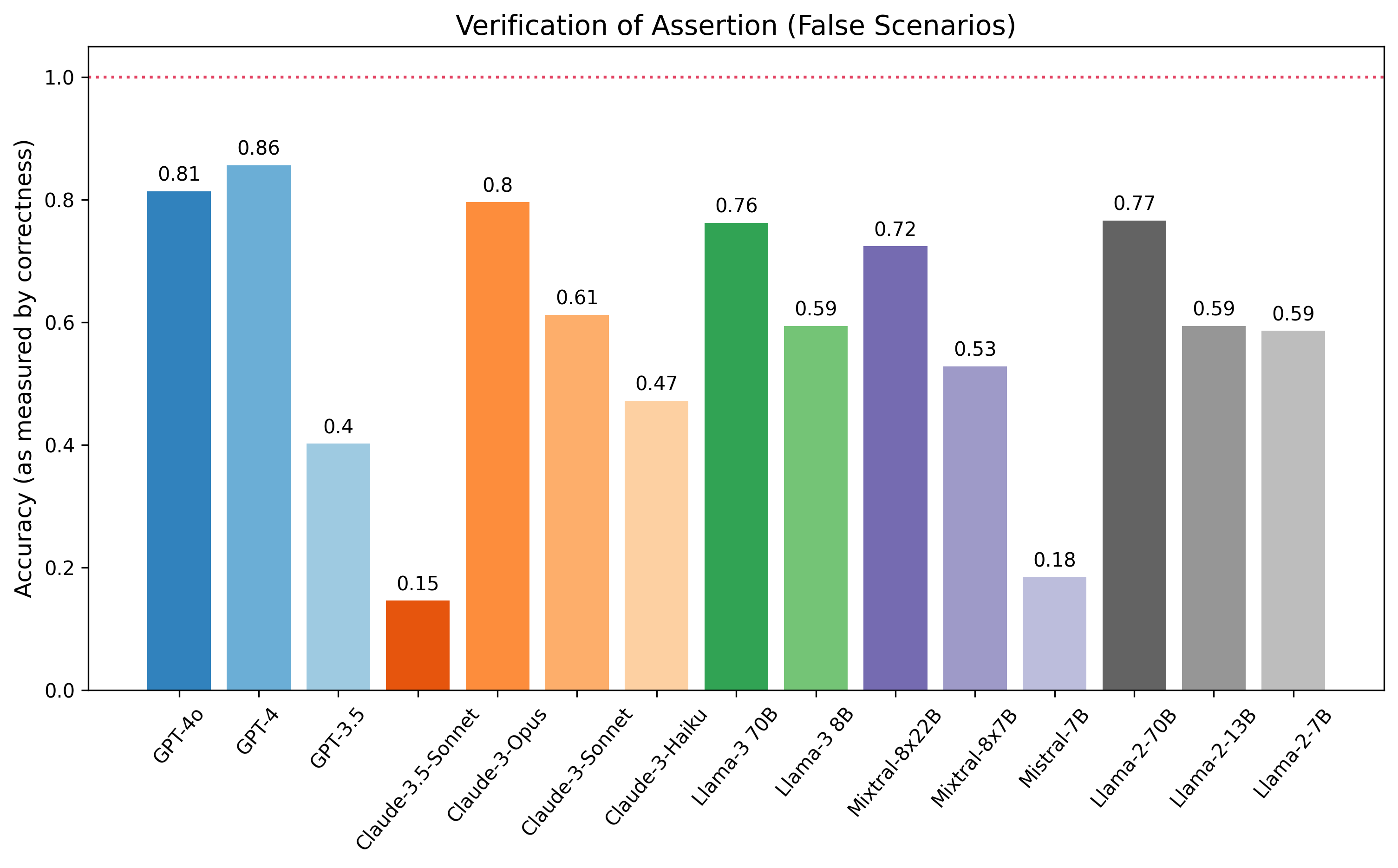}
    \caption{Accuracy of LMs on verification of assertion (\emph{left}: factual, \emph{right}: false scenarios).}
    \label{fig:Verification_of_Assertion_Accuracy}
\end{figure}

\begin{figure}[h]
    \centering
    \includegraphics[width=0.49\textwidth]{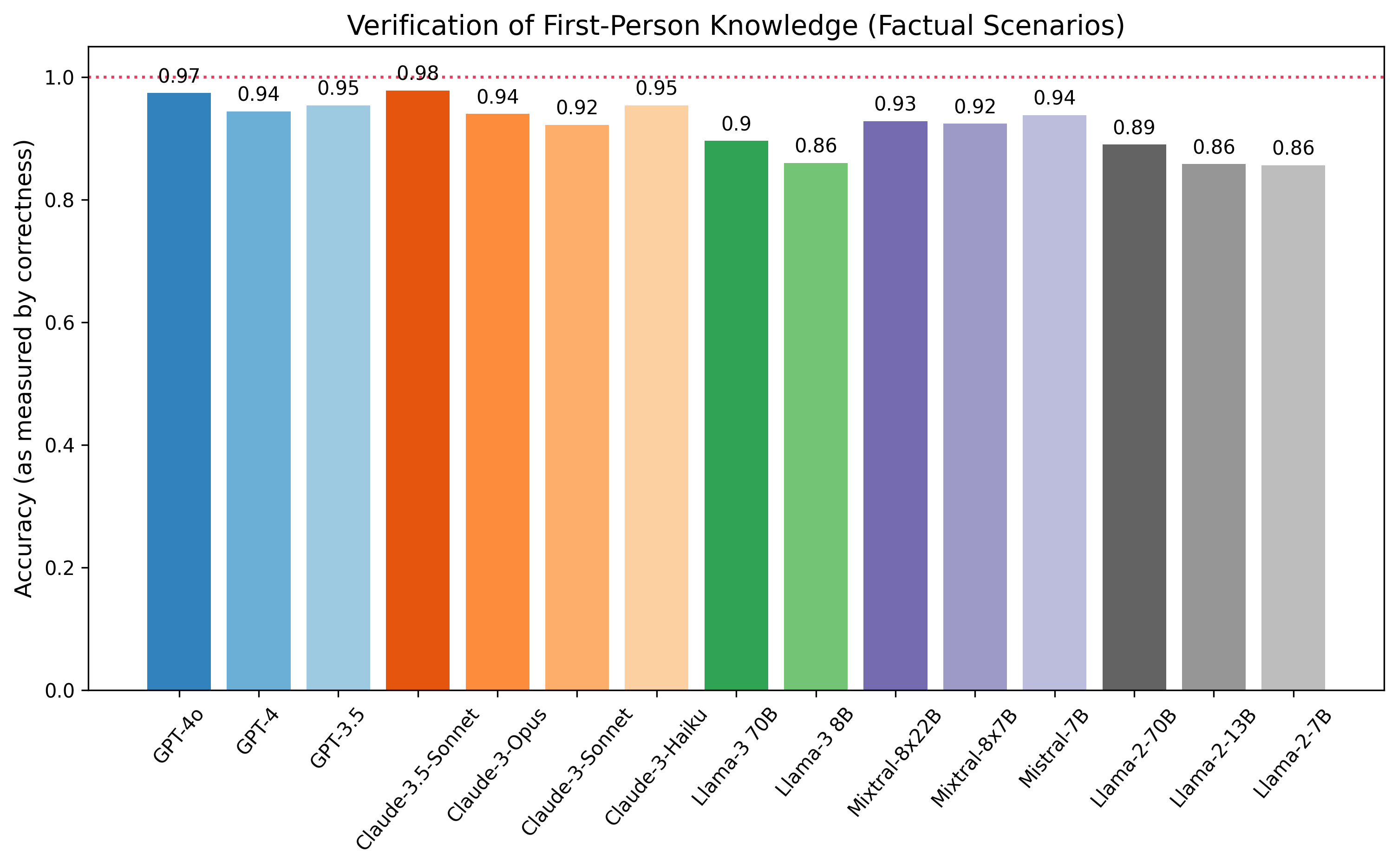}
    \includegraphics[width=0.49\textwidth]{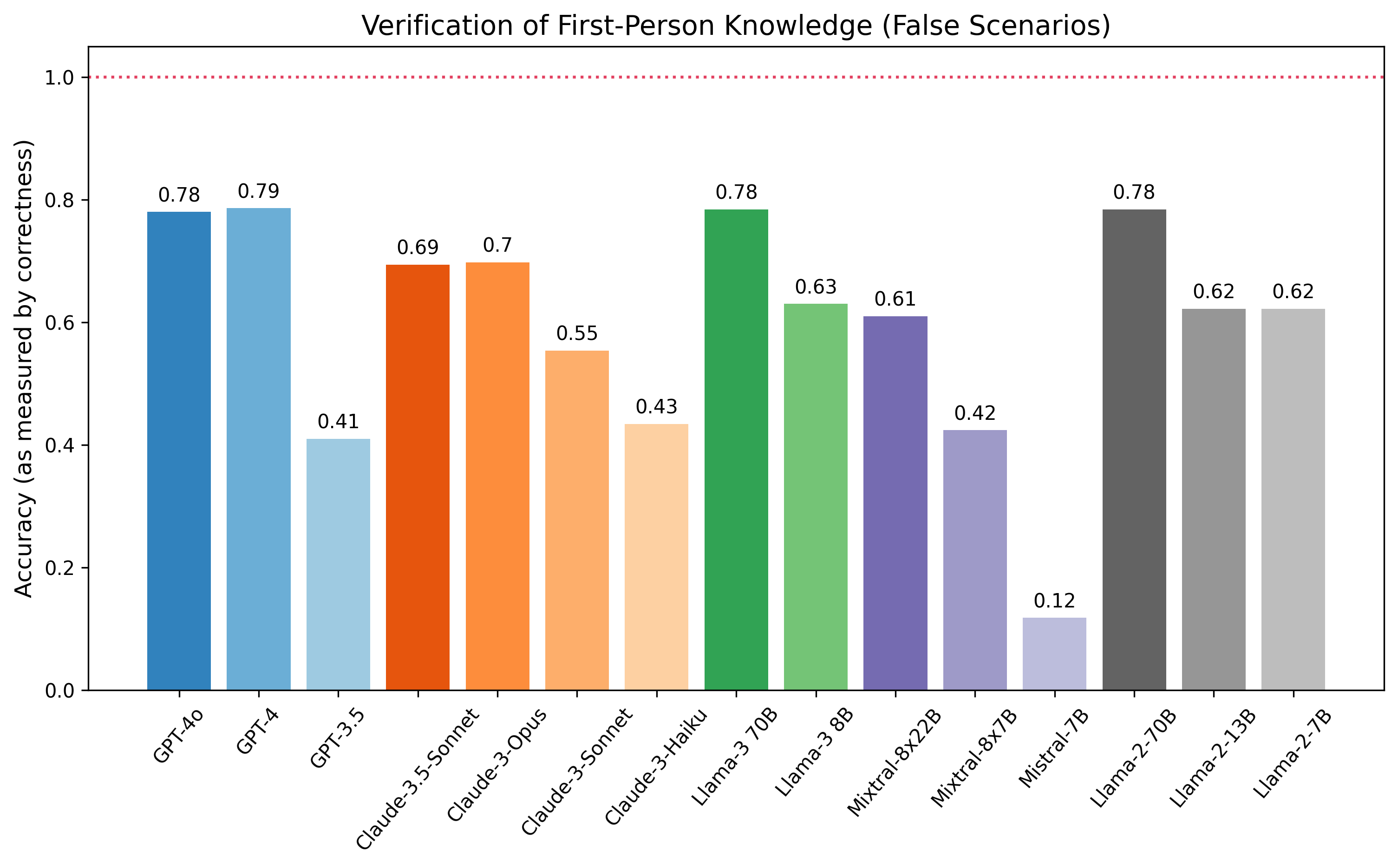}
    \caption{Accuracy of LMs on verification of first-person knowledge (\emph{left}: factual, \emph{right}: false scenarios).}
    \label{fig:Verification_of_First_Person_Knowledge_Accuracy}
\end{figure}

\begin{figure}[h]
    \centering
    \includegraphics[width=0.49\textwidth]{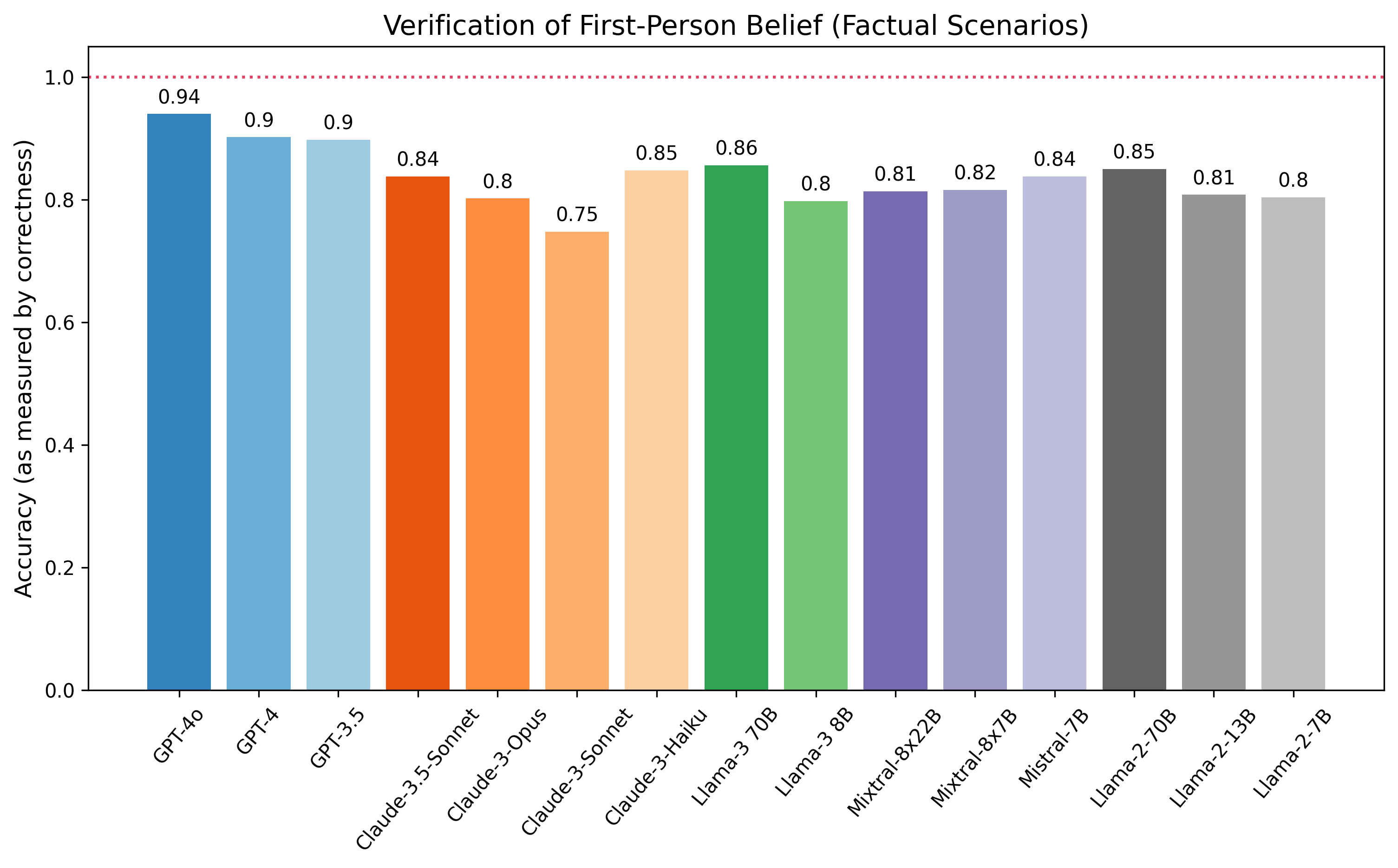}
    \includegraphics[width=0.49\textwidth]{figures/Accuracy/Verification_of_First_Person_Belief_False.png}
    \caption{Accuracy of LMs on verification of first-person belief (\emph{left}: factual, \emph{right}: false scenarios).}
    \label{fig:Verification_of_First_Person_Belief_Accuracy}
\end{figure}

\begin{figure}[h]
    \centering
    \includegraphics[width=0.49\textwidth]{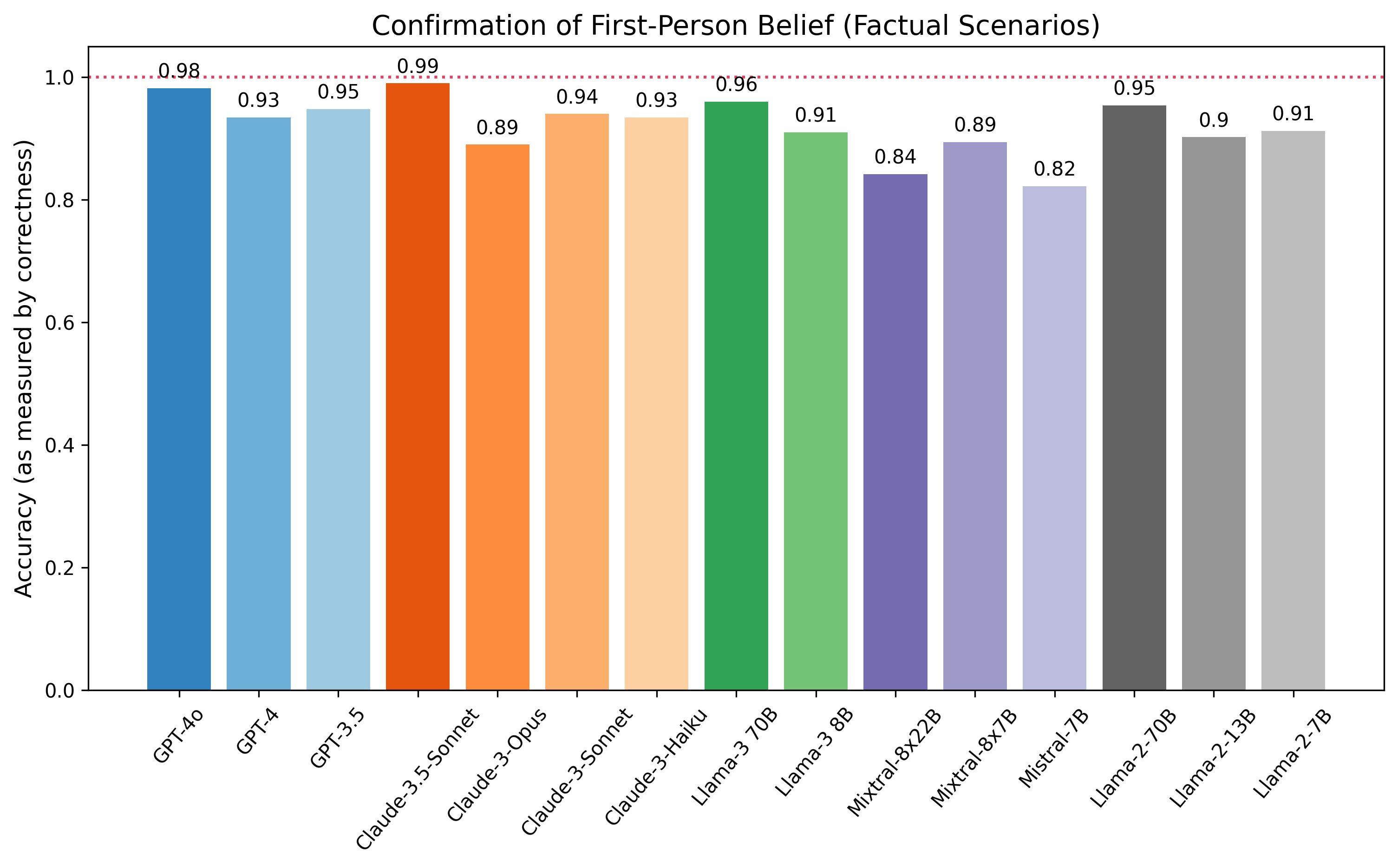}
    \includegraphics[width=0.49\textwidth]{figures/Accuracy/Confirmation_of_First_Person_Belief_False.png}
    \caption{Accuracy of LMs on confirmation of first-person Belief (\emph{left}: factual, \emph{right}: false scenarios).}
    \label{fig:Confirmation_of_First_Person_Belief_Accuracy}
\end{figure}

\begin{figure}[h]
    \centering
    \includegraphics[width=0.49\textwidth]{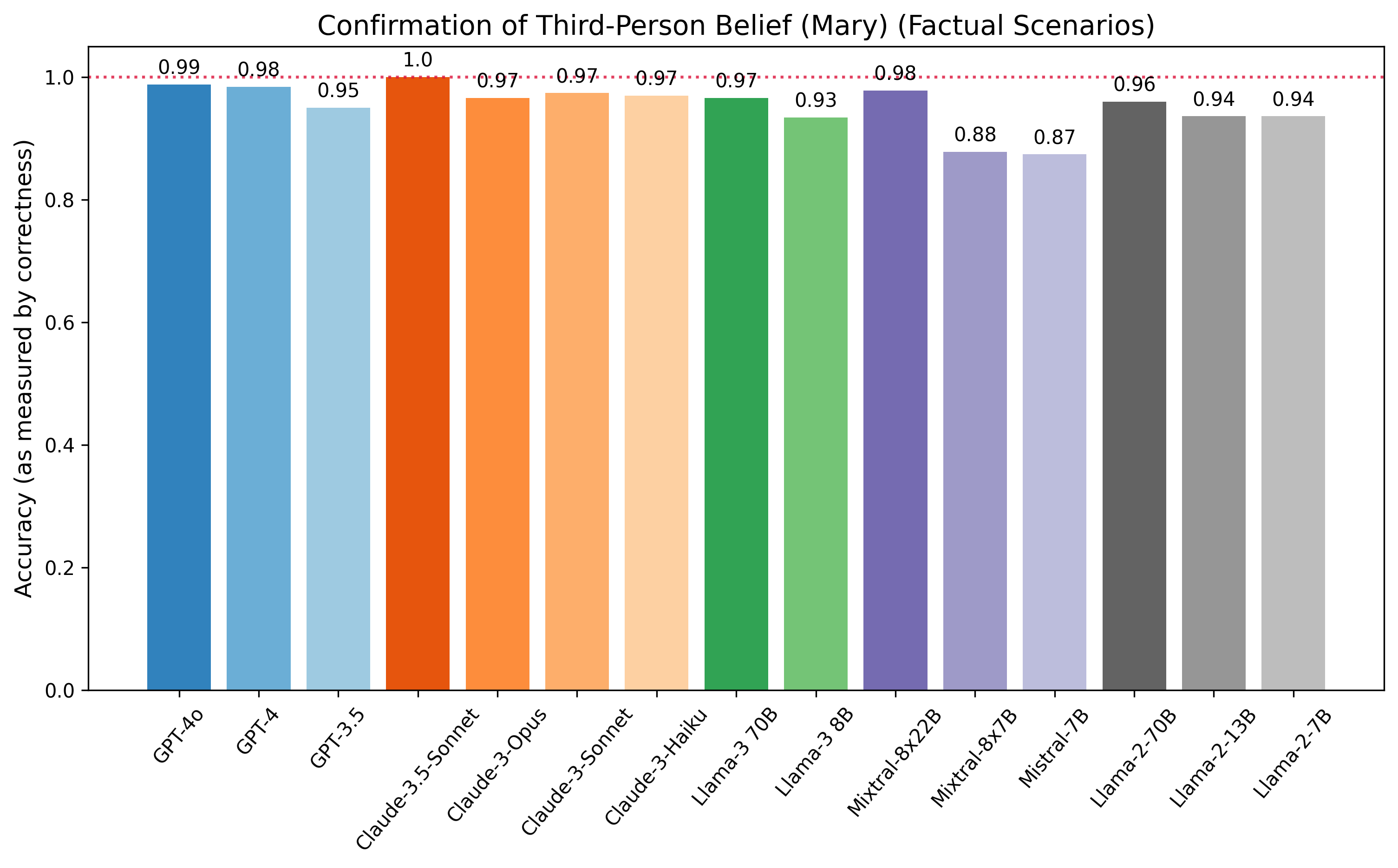}
    \includegraphics[width=0.49\textwidth]{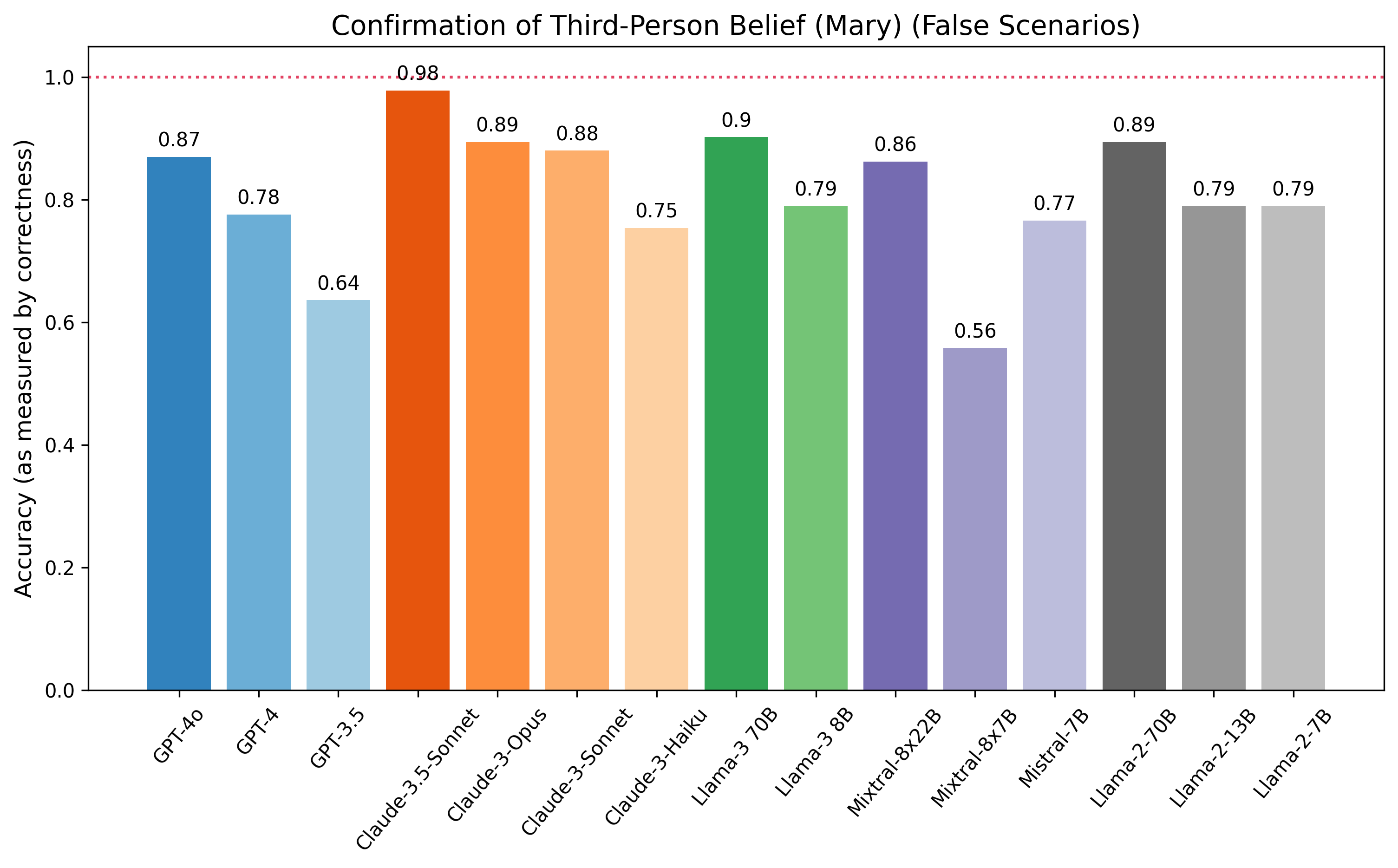}
    \caption{Accuracy of LMs on confirmation of third-person belief (Mary) (\emph{left}: factual, \emph{right}: false scenarios).}
    \label{fig:Confirmation_of_Third_Person_Belief_Mary_Accuracy}
\end{figure}

\begin{figure}[h]
    \centering
    \includegraphics[width=0.49\textwidth]{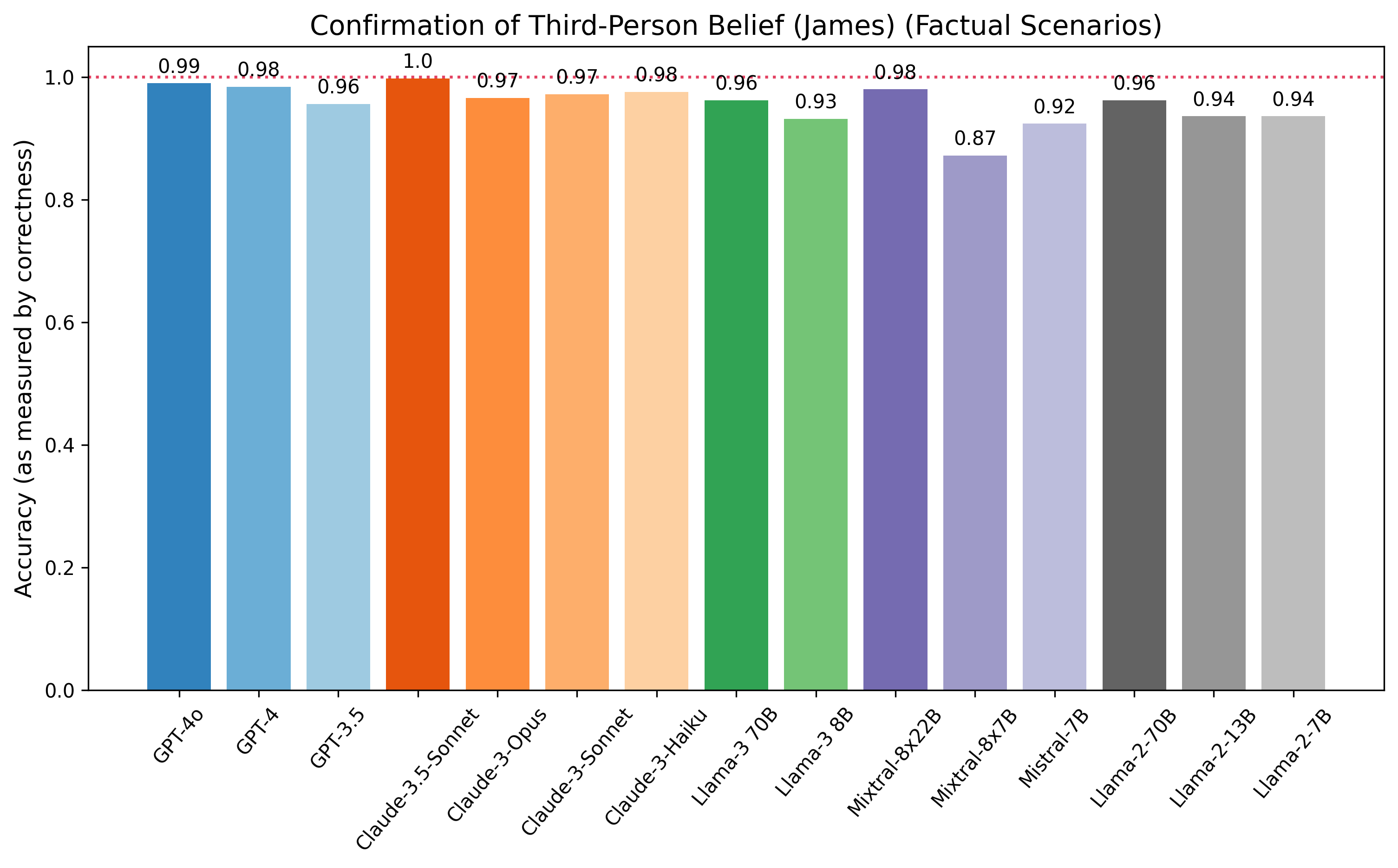}
    \includegraphics[width=0.49\textwidth]{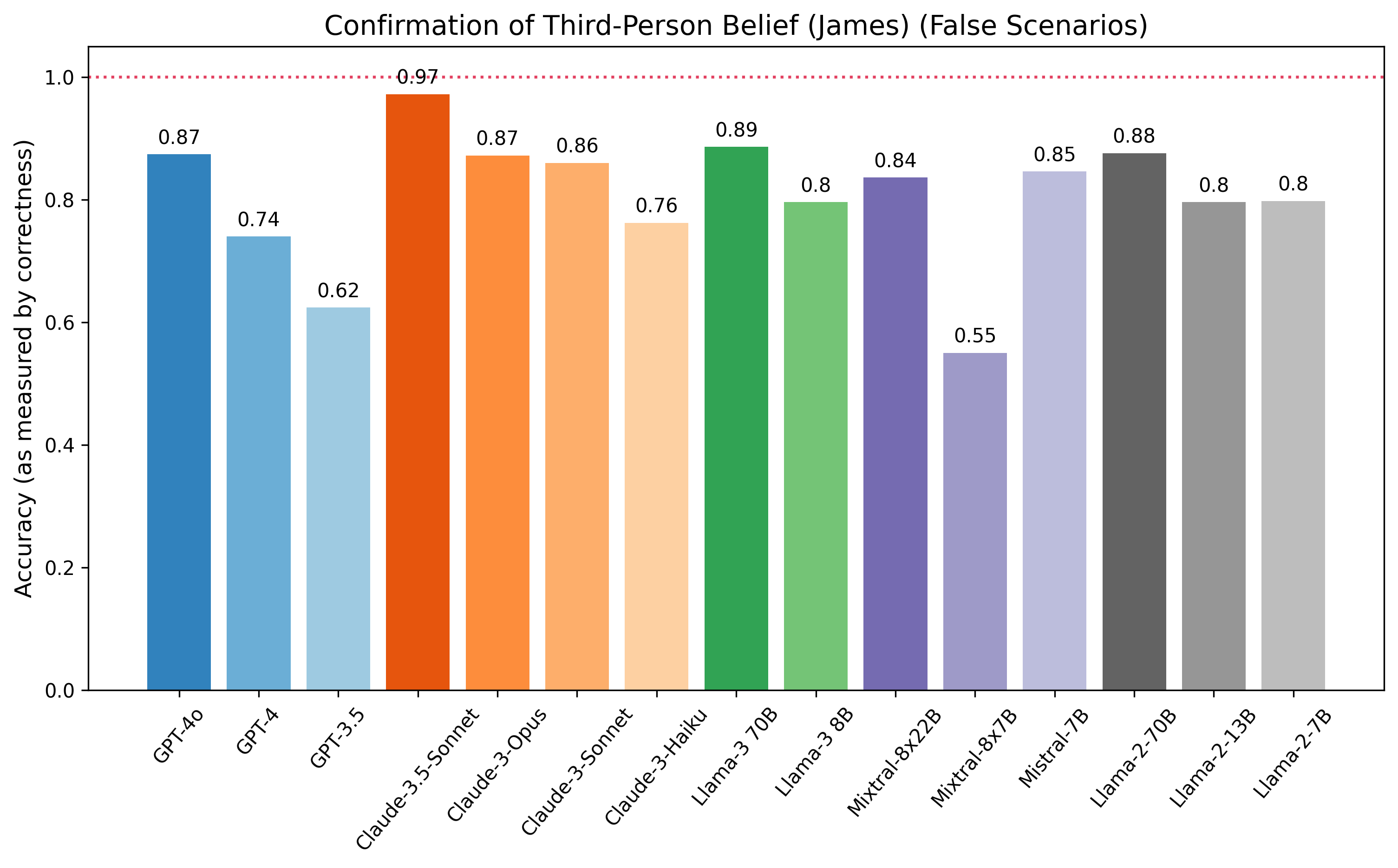}
    \caption{Accuracy of LMs on confirmation of third-person belief (James) (\emph{left}: factual, \emph{right}: false scenarios).}
    \label{fig:Confirmation_of_Third_Person_Belief_James_Accuracy}
\end{figure}

\begin{figure}[h]
    \centering
    \includegraphics[width=0.49\textwidth]{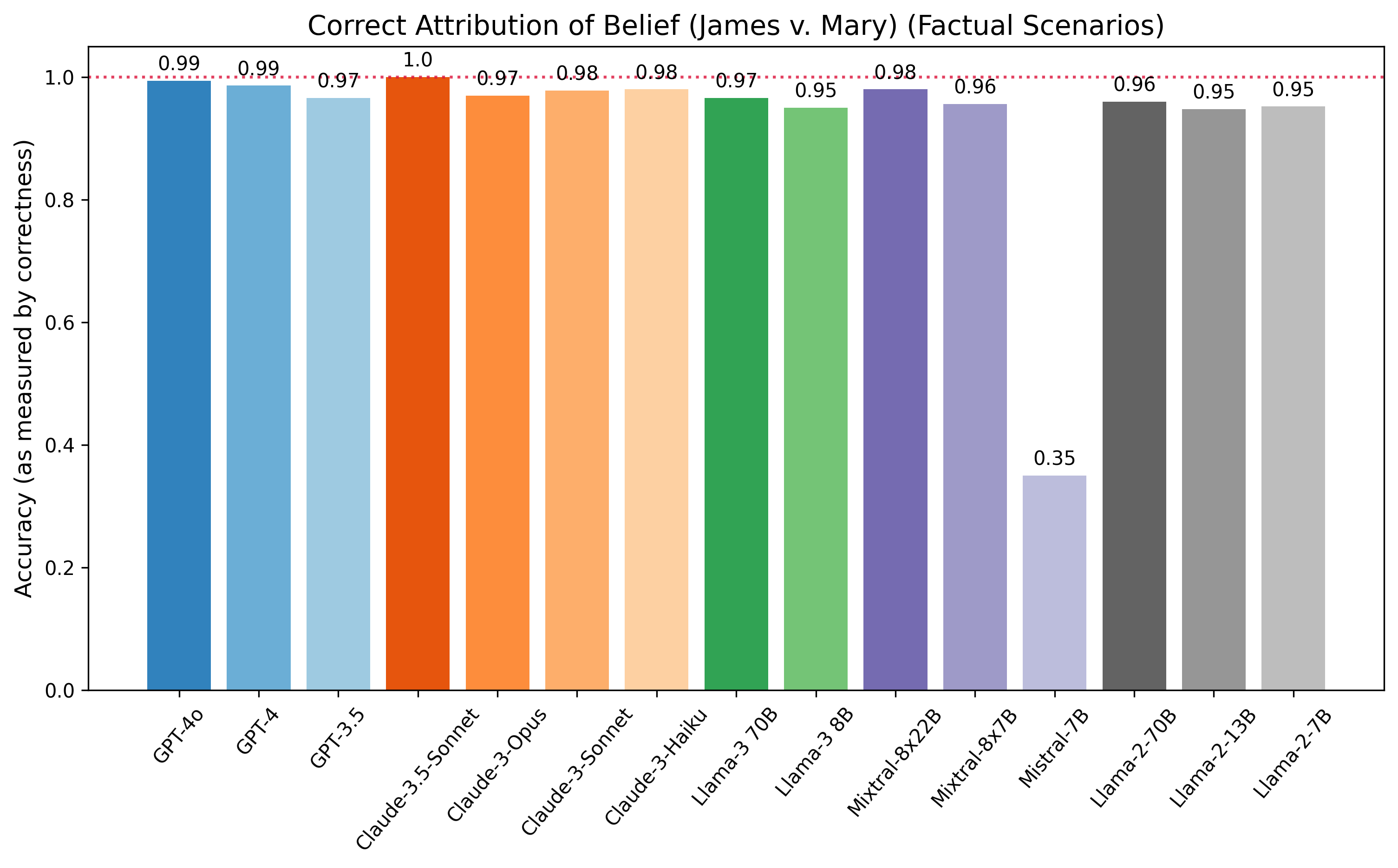}
    \includegraphics[width=0.49\textwidth]{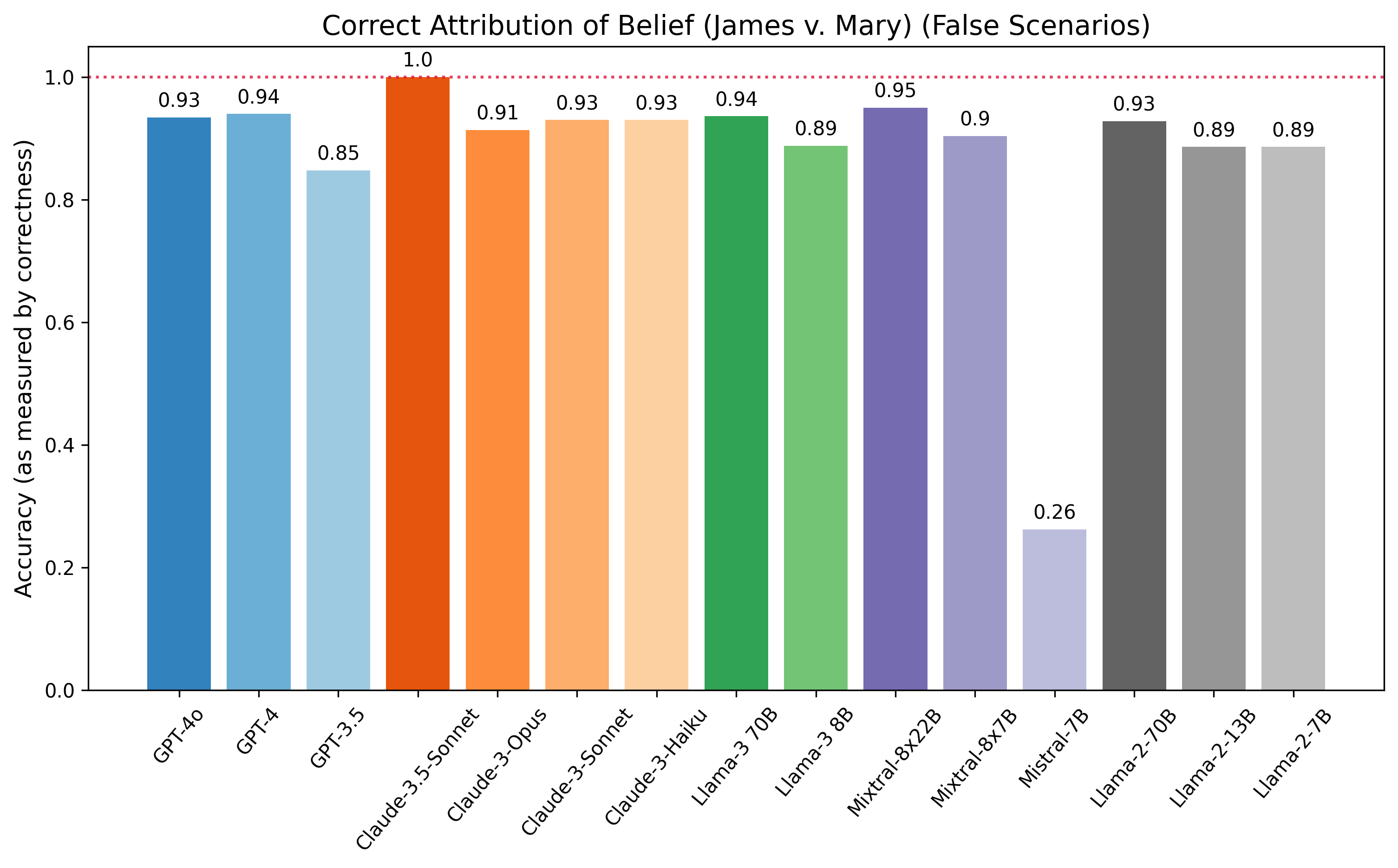}
    \caption{Accuracy of LMs on correct attribution of belief (James v. Mary) (\emph{left}: factual, \emph{right}: false scenarios).}
    \label{fig:Correct_Attribution_of_Belief_James_v_Mary_Accuracy}
\end{figure}

\begin{figure}[h]
    \centering
    \includegraphics[width=0.49\textwidth]{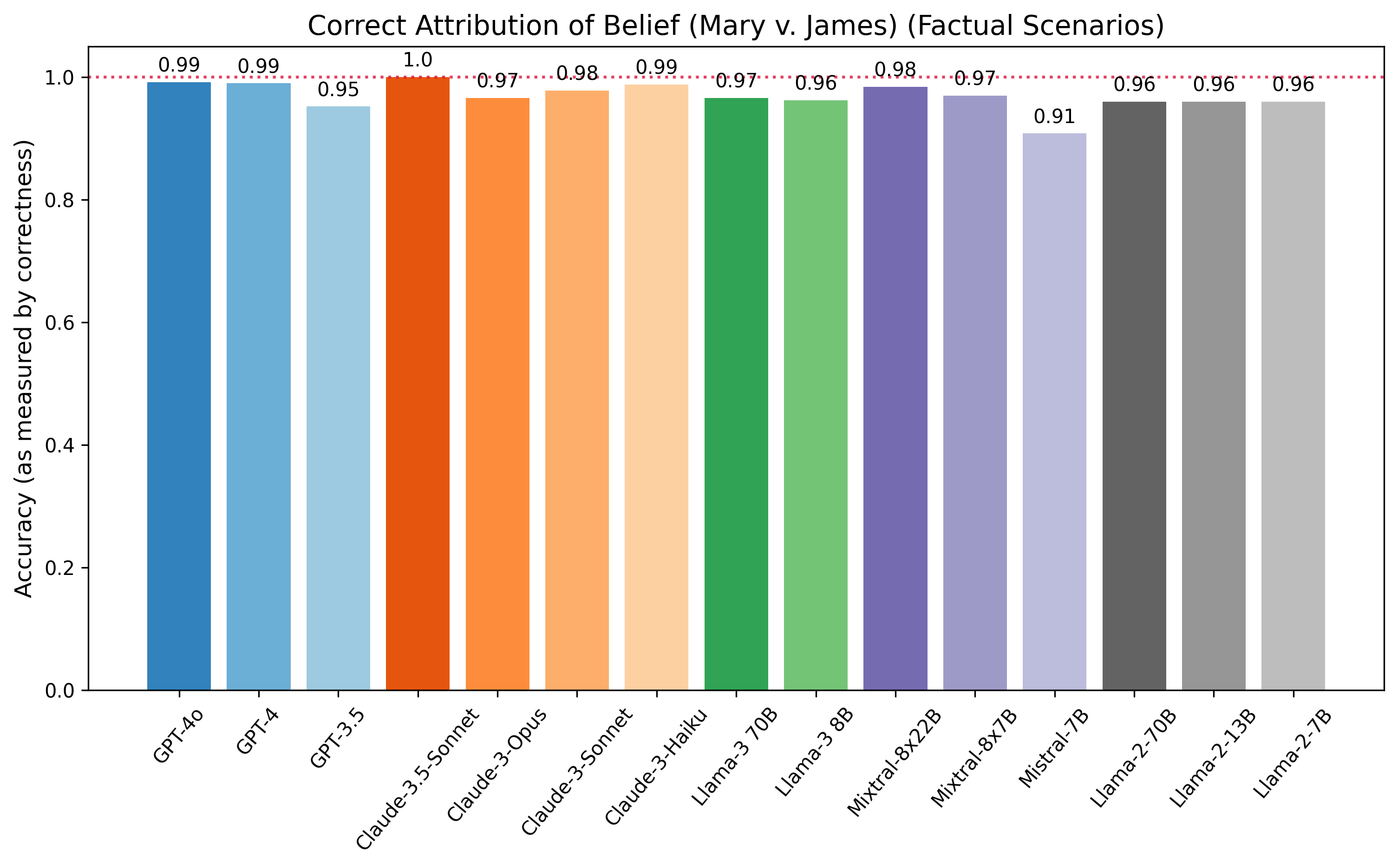}
    \includegraphics[width=0.49\textwidth]{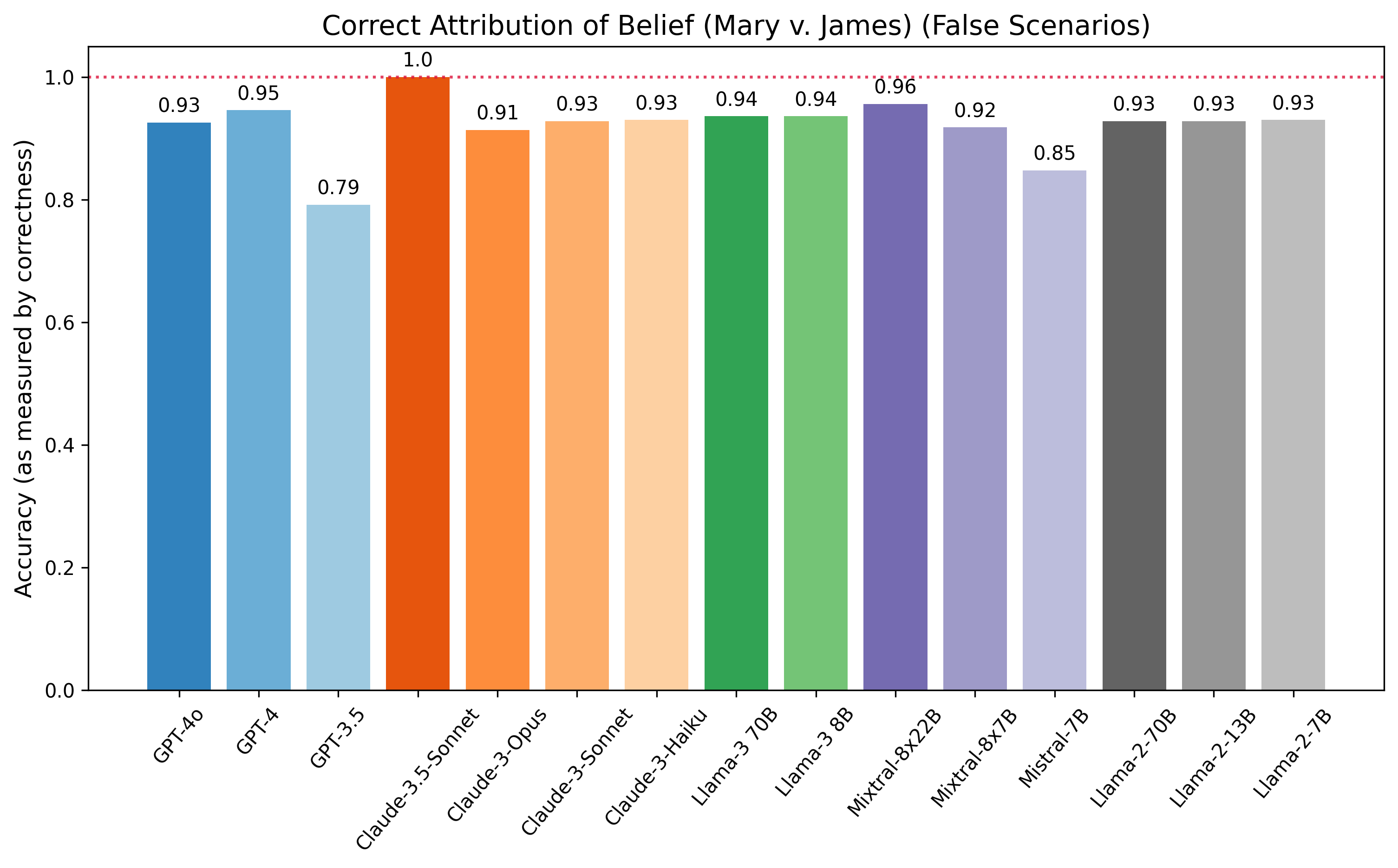}
    \caption{Accuracy of LMs on correct attribution of belief (Mary v. James) (\emph{left}: factual, \emph{right}: false scenarios).}
    \label{fig:Correct_Attribution_of_Belief_Mary_v_James_Accuracy}
\end{figure}

\begin{figure}[h]
    \centering
    \includegraphics[width=0.49\textwidth]{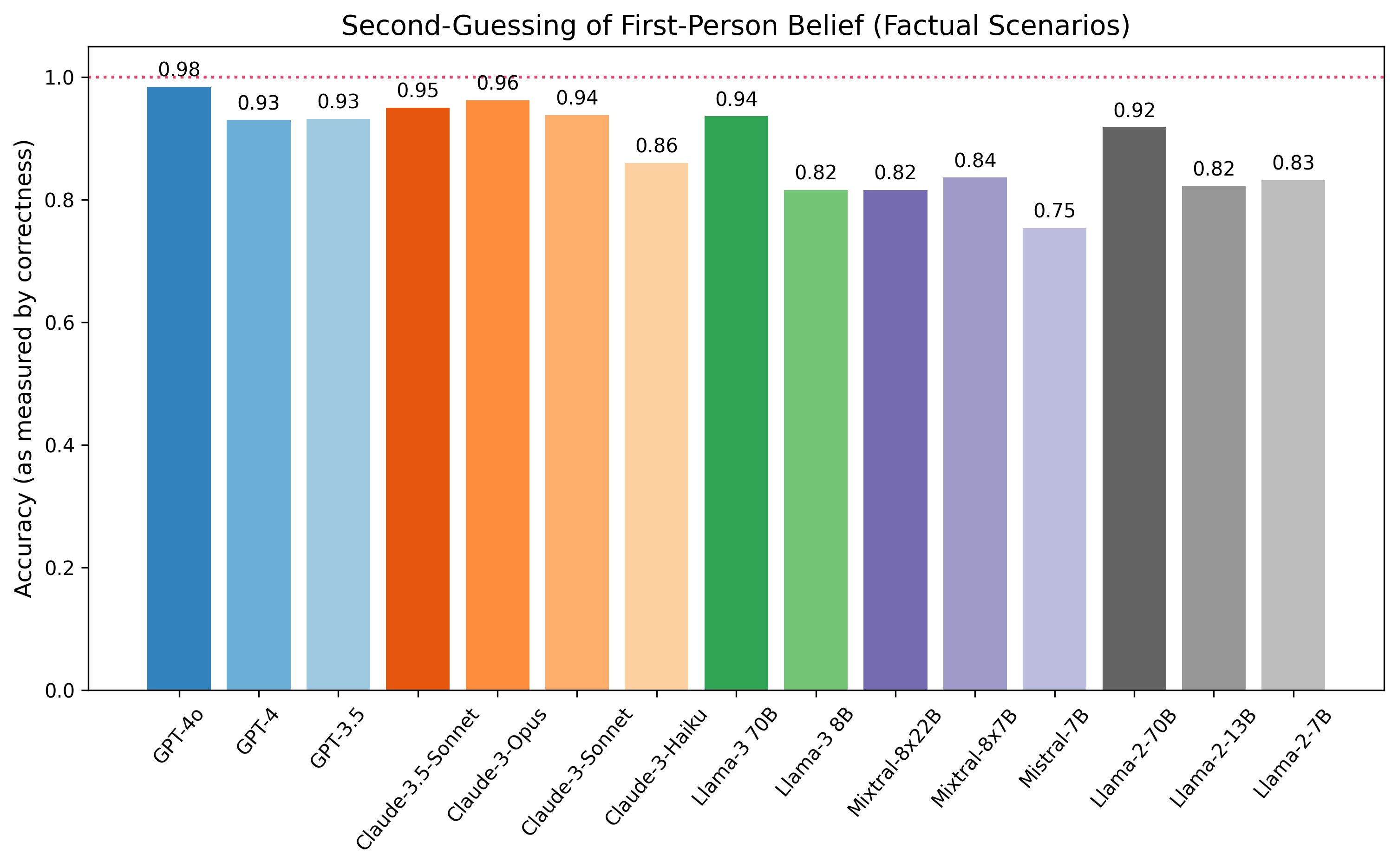}
    \includegraphics[width=0.49\textwidth]{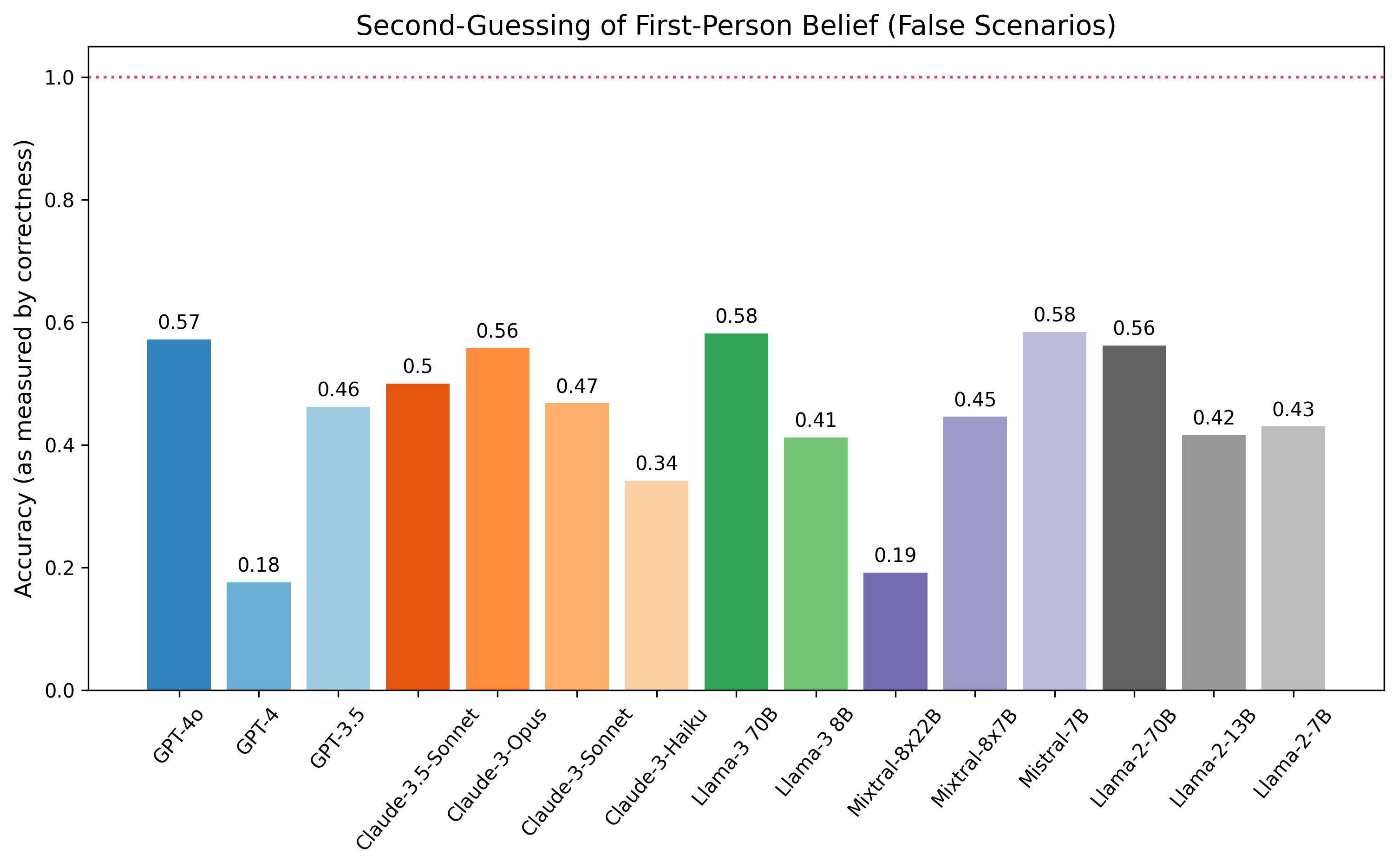}
    \caption{Accuracy of LMs on second-guessing of first-person belief (\emph{left}: factual, \emph{right}: false scenarios).}
    \label{fig:Second_Guessing_of_First_Person_Belief_Accuracy}
\end{figure}

\begin{figure}[h]
    \centering
    \includegraphics[width=0.49\textwidth]{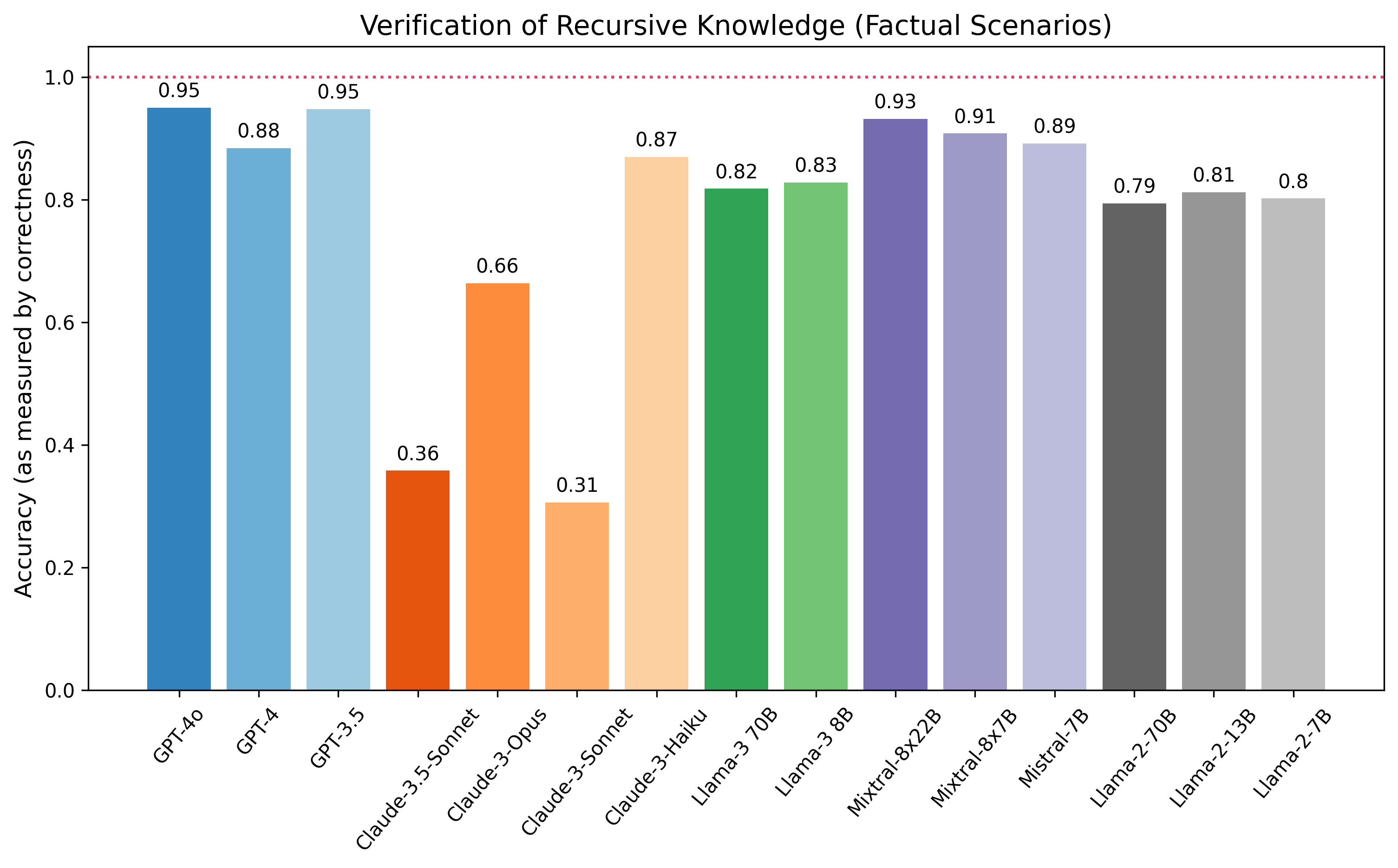}
    \includegraphics[width=0.49\textwidth]{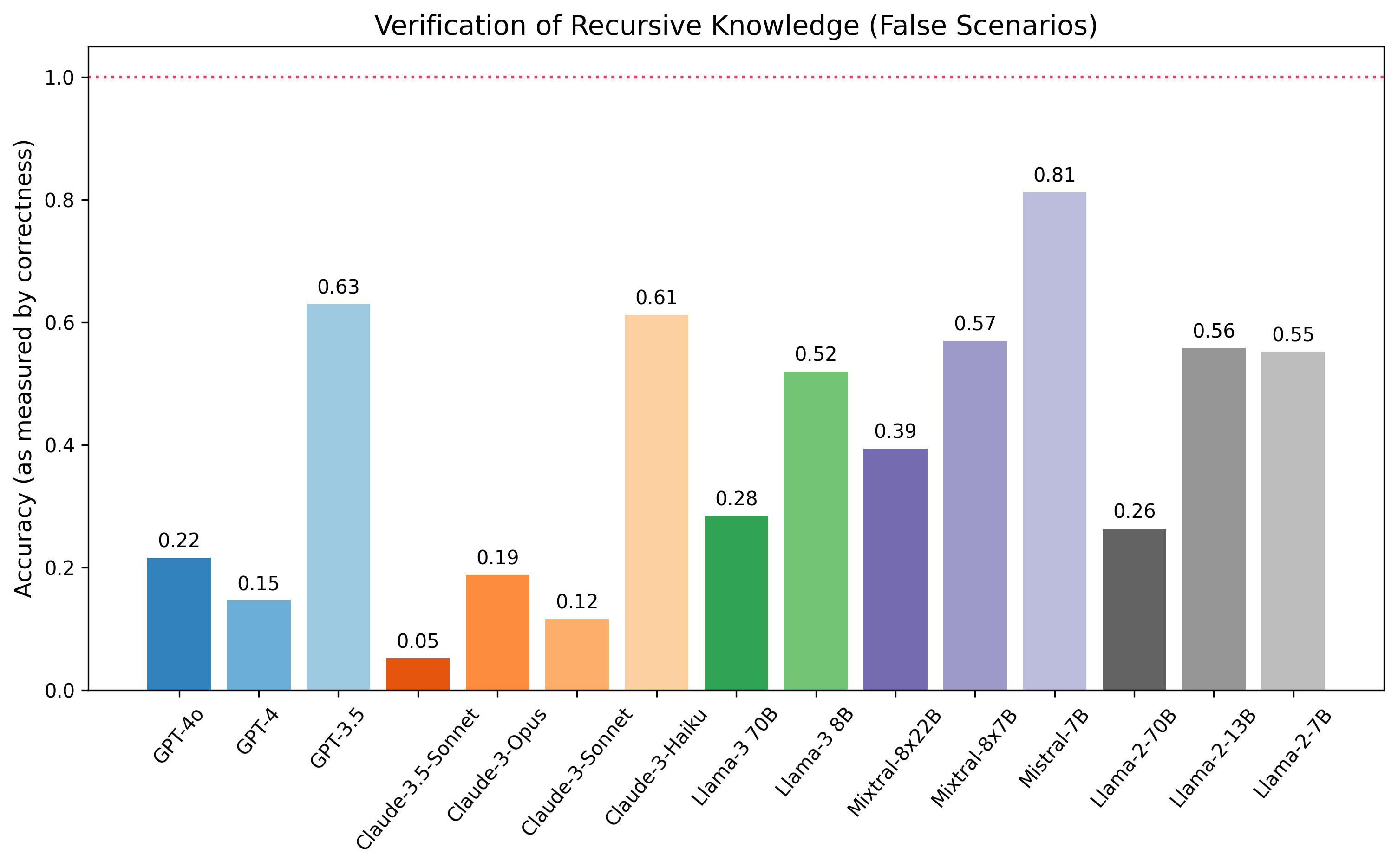}
    \caption{Accuracy of LMs on verification of recursive knowledge (\emph{left}: factual, \emph{right}: false scenarios).}
    \label{fig:Verification_of_Recursive_Knowledge_Accuracy}
\end{figure}

\begin{figure}[h]
    \centering
    \includegraphics[width=0.49\textwidth]{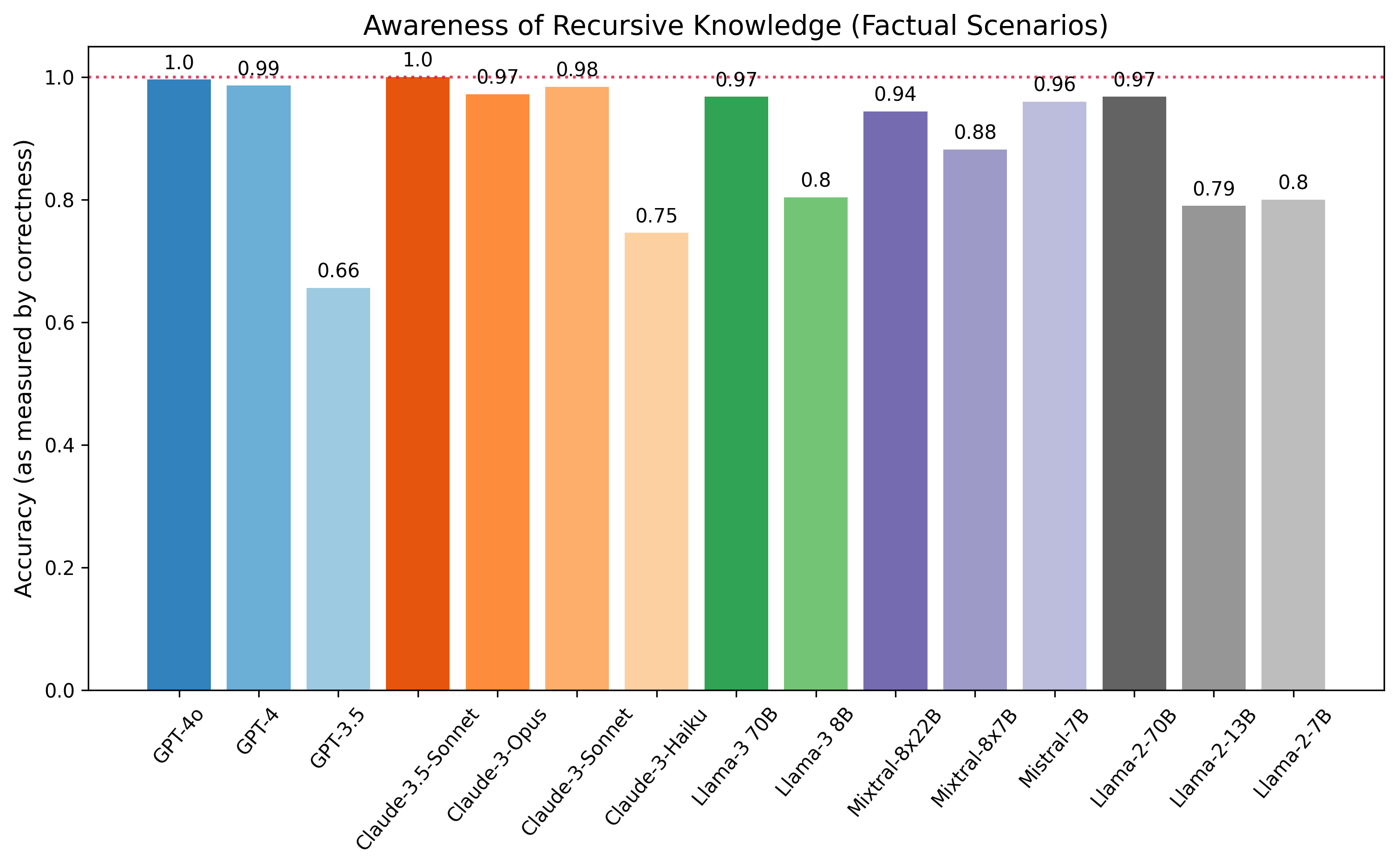}
    \includegraphics[width=0.49\textwidth]{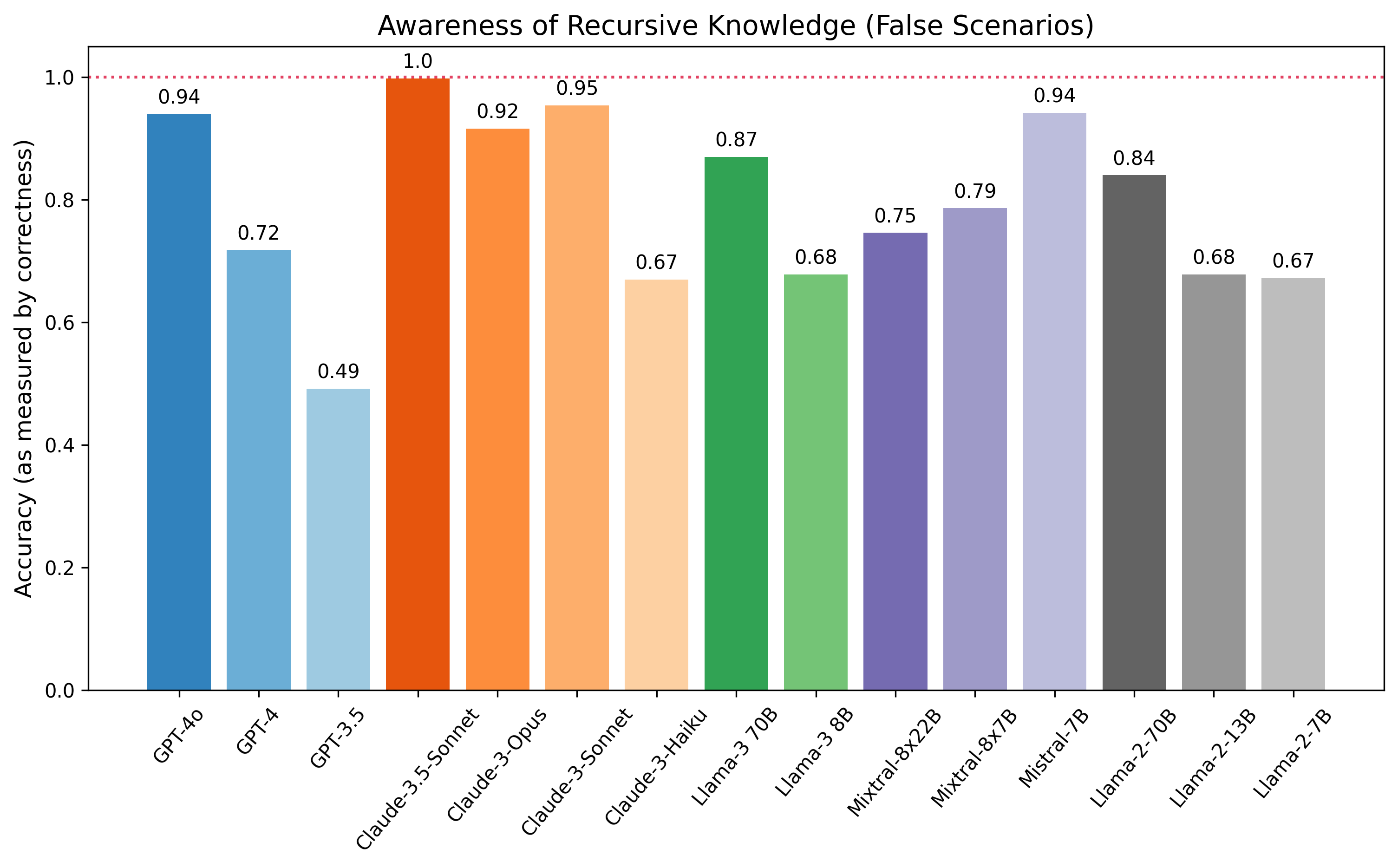}
    \caption{Accuracy of LMs on awareness of recursive knowledge (\emph{left}: factual, \emph{right}: false scenarios).}
    \label{fig:Awareness_of_Recursive_Knowledge_Accuracy}
\end{figure}

\begin{figure}[h]
    \centering
    \includegraphics[width=0.49\textwidth]{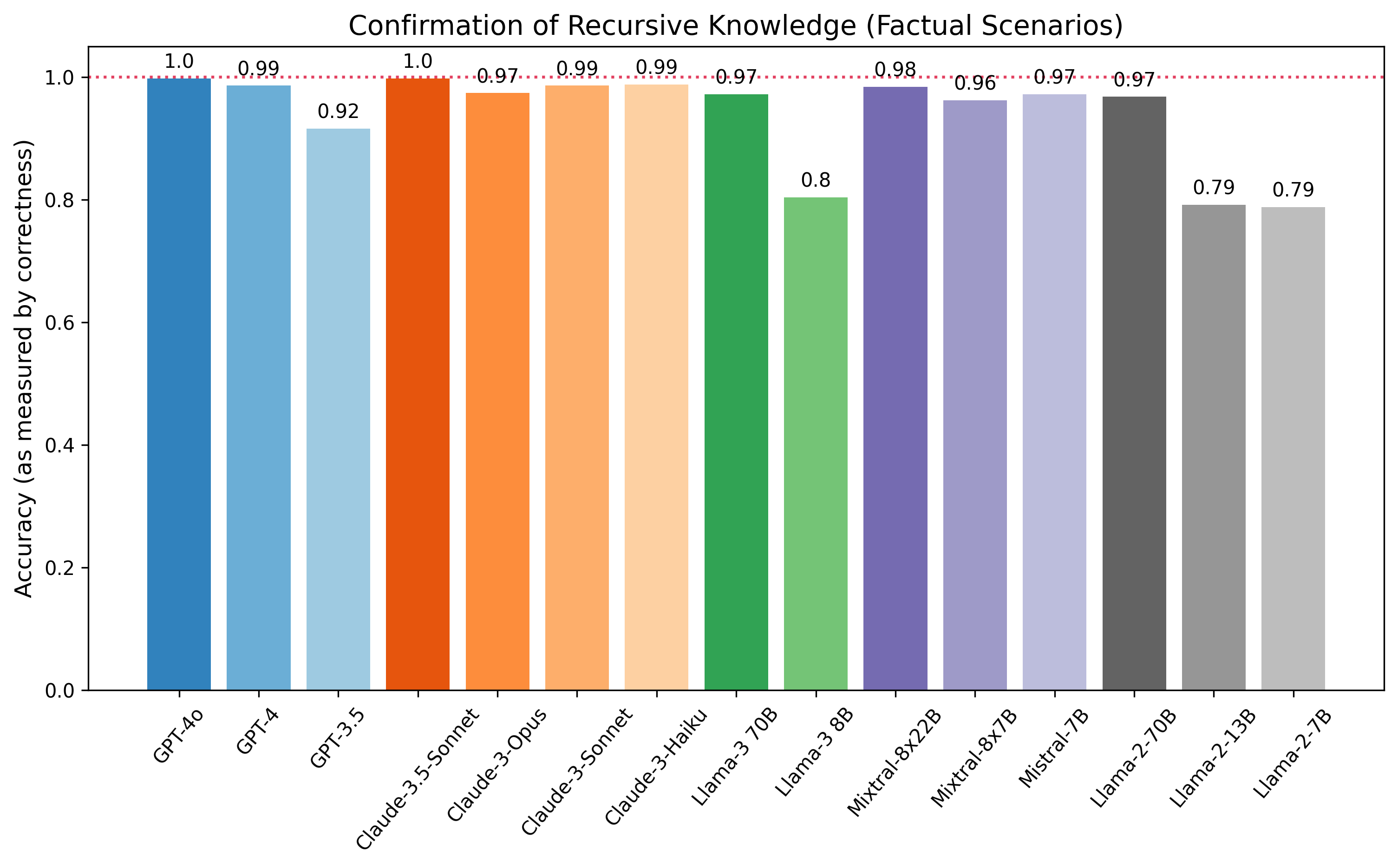}
    \includegraphics[width=0.49\textwidth]{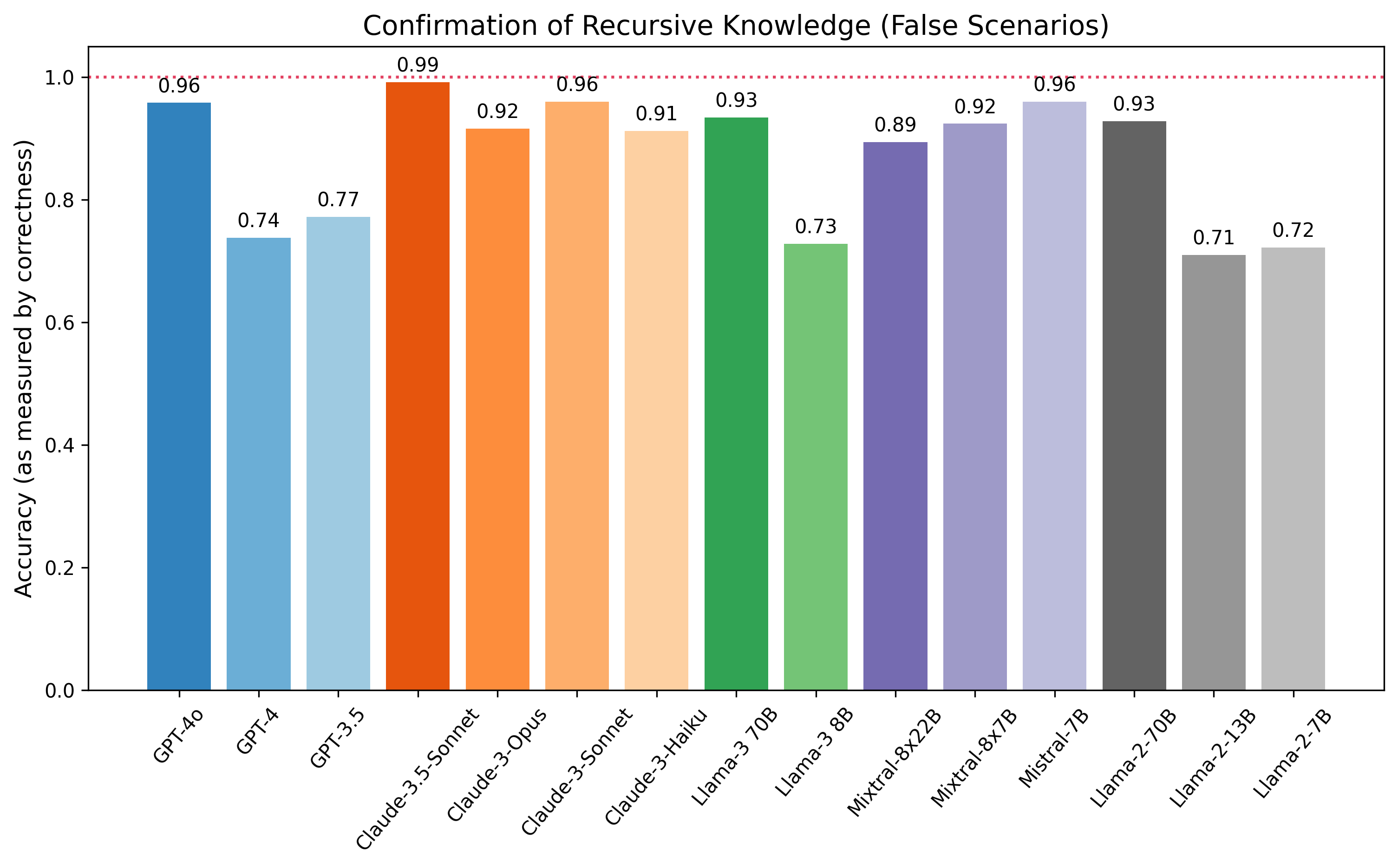}
    \caption{Accuracy of LMs on confirmation of recursive knowledge (\emph{left}: factual, \emph{right}: false scenarios).}
    \label{fig:Confirmation_of_Recursive_Knowledge_Accuracy}
\end{figure}

\clearpage

\clearpage

%% file: sections/prelims-and-background.tex
\section{Preliminaries and Background}
\label{sec:prelims}

\subsection{The Nature of Knowledge and Belief}

The capacity to represent knowledge and belief is fundamental to human cognition and social interaction. While the details of how these capacities and the corresponding verbs (``know'' and so on) are acquired are debated \citep{HacquardLidz}, it is well established that humans start developing the ability to attribute beliefs and knowledge to others from the earliest stages of ordinary cognitive development and socialization~\citep{phillips2021knowledge}.  %
This ability plays a crucial role in social life, allowing us to predict behavior, communicate intentions, and build shared understanding and trust. It has even been argued that, in order for an agent's words to mean anything at all, they must be able to represent what other agents believe, possibly even about what they themselves intend \citep{Grice1957}.

An adequate capacity to represent, process, and reason about belief and knowledge requires an ability to distinguish between their implications.
In everyday language, verbs such as ``believe'' and ``know'' signal different levels of certainty, confidence, and evidence, thereby shaping how messages are perceived and interpreted and how decisions are made. For AI systems---say, LMs such as GPT-4 or Llama-3---accurately processing and responding to these distinctions is therefore essential to avoid misinterpretations, misrepresentations, or misjudgements in high-stakes environments like healthcare, law, and scientific research.

As a rough first pass, we can take a \emph{belief} to be a mental state representing a proposition as true, regardless of whether it is actually true. In contrast, \emph{knowledge} requires a true belief which one has adequate evidence or justification for holding.\footnote{Since~\citet{gettier1963justified}, most philosophers agree that knowledge requires more than justified true belief, but there is no general agreement on what more is required. Since our focus is on the differences between belief and knowledge, where there is much wider agreement (in particular, concerning the status of truth and justification as necessary conditions for knowledge), we will not assume any particular view on the exact characterization of knowledge.} We can capture the formal relationship between belief and knowledge by borrowing some basic notation and principles from modal epistemic logic. Let us represent knowledge with the operator $K$ and belief with the operator $B$, where subscripts denote specific agents. For instance, $K_a p$ means ``agent $a$ knows that $p$,'' while $B_a p$ means ``agent $a$ believes that $p$.'' Throughout this work, we assume that knowledge and belief adhere to the following basic principles:

\textbf{Belief Axiom}: \emph{Knowledge entails belief.} If an agent $a$ knows that $p$, then $a$ believes that $p$.
\begin{equation}
    K_a p \rightarrow B_a p
\end{equation}

This axiom captures the principle that knowledge inherently includes belief~\citep{armstrong1973belief}. For instance, the statement ``I know that $p$, but I do not believe that $p$'' is not only awkward, but seems self-contradictory~\citep{armstrong1969does}. It would, therefore, be nonsensical to say, ``I know that the Earth orbits the Sun, but I do not believe it.''

\textbf{Truth (T) Axiom}: \emph{Knowledge requires truth.} If an agent $a$ knows that $p$, then $p$ must be true.
\begin{equation}
    K_a p \rightarrow p    
\end{equation}

Knowledge is ``factive''---it requires the truth of the proposition known~\citep{sep-knowledge-analysis}. Hence, one cannot know a falsehood such as ``Paris is the capital of Germany'', though one can believe it.

\textbf{Knowledge Distribution (K) Axiom}: \emph{Knowledge is closed under logical implication}. If an agent $a$ knows $q$ and knows that $q$ implies $p$, the agent must also know $p$~\citep{sep-closure-epistemic}:
\begin{equation}
(K_a q \wedge K_a (q \rightarrow p)) \rightarrow K_a p    
\end{equation}

For example, if a person knows that all mammals are warm-blooded and that whales are mammals, they must also know that whales are warm-blooded.

\textbf{Recursive Knowledge Axiom}: \emph{Agents can have knowledge about others' knowledge.} If agent $a$ knows that agent $b$ knows a fact $p$, and knows that $b$’s knowledge implies the truth of $p$, agent $a$ also knows $p$.
\begin{equation}
    (K_a (K_b p) \wedge K_a (K_b p \rightarrow p)) \rightarrow K_a p
\end{equation}

This principle governs cases where knowledge is nested, that is, when one agent knows that another agent knows something. For example, if Alice knows that Bob knows that a water molecule ($\text{H}_2 \text{O}$) consists of two hydrogen (H) atoms and one oxygen (O) atom, then Alice also knows this fact.

Altogether, these principles illustrate a fundamental asymmetry between belief and knowledge. While knowledge entails belief and truth, belief does not entail knowledge or truth. An individual can hold false beliefs, but false knowledge is a contradiction in terms.

In the context of AI, these distinctions become even more important and consequential. A model that fails to recognize the differences between belief and knowledge risks propagating errors and misunderstanding human communication. For instance, in healthcare, a system which fails to recognize that belief does not entail truth could misinterpret a patient's belief (``I believe I have a cancer'') as fact (``the patient has cancer''), leading not only to inaccurate diagnosis or faulty treatment but even death. Or a system which fails to recognize how people's behaviours can be shaped by their false beliefs, such as the child who writes letters to Santa and expects gifts during Christmas, may be unreliable when predicting the behaviour of human agents.

Our study investigates whether current LMs can grasp these fundamental distinctions. We evaluate their ability to distinguish between belief and knowledge, especially in scenarios that require understanding others' beliefs and reasoning about recursive knowledge. The results shed light on the limitations of current AI models and provide insights into how they might be improved to better reflect human epistemic reasoning before becoming even more integrated into society.

\subsection{Contextual Sensitivity and Factive Distinctions}

An additional potential layer of complexity is due to the fact that ordinary language usage of epistemic terms can be  ambiguous or pragmatically flexible. Most relevant to our discussion are apparently non-factive uses of the verb ``know.'' For instance, consider the following:

\begin{enumerate}[label=(\roman*)]
        \item ``I know I opened the door.'' \citep{Hazlett2010-HAZTMO} \\ (Reporting on what one remembers having done, where it turns out you are mistaken.)
        \item ``I knew Hillary Clinton would win the election.'' \citep{sep-knowledge-analysis}
        \item ``I knew I was going to die out there.'' \citep{Hazlett2010-HAZTMO} \\ (Reporting a  conviction or prediction at the time, which turns out to be mistaken.)
    \end{enumerate}

Adjudicating whether such cases invalidate the factivity of knowledge, or how the relation between ordinary language and the theory of knowledge should be conceived more generally, are beyond the scope of our discussion. Note, however, how the provision of context in both (i) and (iii), and the use of the past tense in (ii) and (iii), are crucial for motivating the non-factive readings. Since we provide our prompts in the absence of such context and always in the present tense, we assume it is most reasonable for the models, as well as human agents, to interpret them factively.

\clearpage

\section{Related Work: Commonsense Reasoning and Theory of Mind Evaluations}
\label{sec:related-work}

Our work is closely related to the following bodies of work.

\textbf{Logical and commonsense reasoning.} Endowing LMs with logical and commonsense reasoning abilities remains a central challenge in AI. The emergence of large, pre-trained models such as GPT-3 and GPT-4 \citep{brown2020language,GPT4} has intensified efforts to evaluate their capacity for human-like reasoning across domains such as arithmetic, logical deduction, probabilistic inference, and commonsense understanding. While LMs have demonstrated impressive performance on various generation tasks, questions persist about whether they truly reason or merely replicate patterns from training data \citep{zheng2023large,kandpal2023large}. Techniques like chain-of-thought (CoT) prompting \citep{wei2022chain}, which encourage models to generate intermediate reasoning steps, have improved performance on complex tasks \citep{wei2022emergent,suzgun2023challenging}. However, studies indicate that these models still struggle with multi-step logical deduction tasks and maintaining logical consistency \citep{valmeekam2022large,creswell2022selection}. Benchmarks like ReClor \citep{yu2020reclor} and LogiQA \citep{liu2021logiqa} have been used to show limitations in LMs' logical reasoning compared to humans.

Commonsense reasoning poses additional challenges for LMs~\citep{choi2022curious}. For instance, benchmarks such as CommonsenseQA~\citep{talmor2019commonsenseqa,talmor2022commonsenseqa} and Winograd Schema~\citep{levesque2012winograd} assess LMs' understanding of everyday scenarios. While progress has been made, LMs still find it difficult to perform well on tasks requiring subtle disambiguation or long, deep contextual comprehension \citep{bhargava2022commonsense,davis2023benchmarks}. Recent efforts to enhance reasoning in LMs include fine-tuning on data with high-quality natural-language explanations~\citep{chung2024scaling}, integrating explicit reasoning prompting techniques or modules \citep{kojima2022zero,zhou2023leasttomost,yao2023tree,suzgun2024meta}, and employing self-consistency techniques \citep{wang2022self,suzgun2023follow}, among others. Despite these advancements, however, important challenges still persist, since these models can be sensitive to prompt phrasing, may produce logically plausible but incorrect reasoning, and tend to overfit to specific task formats \citep{valmeekam2022large,webson2022prompt,webson2023language,shapira-etal-2024-clever,turpin2024language}.

Most related to our work, \citet{holliday2024conditional} tested a wide range of LMs on a suite of questions involving conditionals and epistemic modals to evaluate how much the reasoning abilities of LMs match those of theoretical frameworks in linguistic semantics and philosophy of language. Their study found that even the GPT-4 family exhibited logically inconsistent judgments across inference patterns involving epistemic modals. Nearly all models gave answers to certain complex conditional inferences---widely debated in the literature---that diverged from human judgments.  These results illustrate serious gaps in basic logical reasoning, revealing inconsistent and counterintuitive reasoning behaviors even in top-performing models with or without CoT. Overall, their work demonstrated that LMs' judgments may be unreliable when encountering novel inference patterns natural to humans and highlights the need for more out-of-distribution data to rigorously test LMs.

Recent studies~\citep{basmov2024llms,basmov2023simple} have provided useful insights into LMs’ linguistic capacities in simple reading comprehension contexts, yet our work advances the field by offering a more comprehensive analysis of epistemic reasoning. \citet{basmov2024llms}, for instance, examine LMs’ ability to process non-affirmative and hypothetical statements through a relatively small dataset of 50 triplets derived from WebQuestions, analyzing how models handle modal constructs. However, their work is constrained by its narrow scope as it focuses on ``imaginary data'' rather than real-world complexities. Our study addresses these limitations by introducing \dataset, a novel evaluation suite comprising 13,000 questions spread across multiple domains and tasks; \dataset allows us to assess LMs' ability to distinguish fact from belief and knowledge in both true and false contexts. This broader dataset allows for a deeper exploration of how LMs engage with real-world epistemic challenges, particularly when faced with contradictory beliefs and objective facts---a domain that \citet{basmov2024llms} do not readily address or focus on.

Moreover, while \citet{basmov2024llms} focus on hypothetical contexts, our research extends into more practical scenarios where recognizing false beliefs and understanding first-person epistemic statements are critical, particularly in fields like healthcare, counseling, and law. These real-world stakes necessitate a deeper understanding of subjective belief systems, an area where our study offers novel insights by demonstrating LMs’ struggles to process false first-person beliefs effectively.
\citet{basmov2023simple}, on the other hand, focus on elementary semantic inferences, revealing the models’ difficulties with basic linguistic tasks. However, their focus on simple entailments leaves unexamined more complex epistemological tasks, such as recursive knowledge assessment and belief attribution. Our research addresses this gap by evaluating LMs' layered epistemic reasoning abilities, uncovering challenges in handling recursive logic and distinguishing personal from external beliefs. Through this more nuanced and layered examination, our study offers a deeper understanding of the limitations of current language models in high-stakes applications.

\textbf{Theory of mind evaluations on modern LMs.} Theory of mind (ToM) refers to the capacity to process, explain, and predict people's observable actions in terms of their mental states, such as beliefs, desires, and intentions~\citep{baker2011bayesian}. ToM is considered to be a foundational aspect of human cognitive development and social interaction~\citep{carlson2013theory,meltzoff2002imitation,shatz1994theory}. Recently, as modern LMs have demonstrated increasingly sophisticated language understanding and generation abilities, researchers have begun to investigate whether these models can understand and reason about mental states similar to how humans do.  Several recent studies~\citep{sap-etal-2022-neural,gandhi2023understanding} found that models such as GPT-3 performed poorly on simple ToM tasks in zero-shot settings. Subsequent studies~\citep{bubeck2023sparks,kosinski2023theory}, however, have claimed that more recent and larger models such as GPT-4 display emergent ToM capabilities. \citeauthor{bubeck2023sparks} have even gone as far as to claim that GPT-4 has shown ``sparks of artificial general intelligence'' and that their ``findings suggest that GPT-4 has a very advanced level of theory of mind.'' 

These disparate findings and bold claims sparked a lively debate in the AI community and were quickly met with skepticism and critical examinations~\citep[][\emph{inter alia}]{trott2023large,aru2023mind,mahowald2024dissociating}. \citet{marcusdavis2023}, for instance, outlined many methodological issues with the experiments conducted by \citet{kosinski2023theory}, while \citet{ullman2023large} and \citet{shapira-etal-2024-clever} demonstrated that minor alterations to test samples could dramatically reduce model performance on elementary ToM tasks. \citet{shapira-etal-2024-clever} further argued that these large-scale LMs might be relying on shallow heuristics and superficial patterns rather than displaying genuine understanding of these ToM tasks.

\textbf{ToM and social reasoning benchmarks.} The recent challenges in accurately assessing ToM capabilities of LMs have spurred the development of more comprehensive benchmarks and evaluation methods. Evaluation suites such as 
ToMi~\citep{le2019revisiting}, 
ToMChallenges~\citep{Ma2023ToMChallengesAP}, 
BigToM~\citep{Gandhi2023UnderstandingSR},
HiToM~\citep{wu-etal-2023-hi-tom},
EPITOME~\citep{jones2023epitome},
Thinking for Doing~\citep[T4D;][]{zhou2024T4D},
OpenToM~\citep{xu-etal-2024-opentom},
\emph{inter alia}, 
have provided valuable tools for probing different aspects of ToM reasoning. However, as highlighted by~\citet{ma-etal-2023-towards-holistic} and others~\citep{le2019revisiting,aru2023mind,shapira-etal-2024-clever}, there is growing recognition in the field that existing benchmarks may still not be fully capturing the nuanced cognitive processes underlying ToM and that they are also prone to data contamination, spurious correlations, and implicit linguistic biases. This realization has led to calls for more holistic approaches to evaluation, including methods and benchmarks that can assess precursor abilities such as perception inference~\citep{rakoczy2022foundations} and perception-to-belief inference~\citep{pratt1990young,baron1994seeing}---as exemplified in~\citep{jung2024perceptions}---or provide a more comprehensive and conclusive picture of LM's capabilities and limitations in ToM and social reasoning tasks. Overall, as noted by~\citet{mahowald2024dissociating}, ``[t]o assess progress on the road toward building models that use language in human-like ways, it is important to develop benchmarks that evaluate both formal and functional linguistic competence.''

\textbf{\emph{Comparison with existing ToM evaluations}.} Our work distinguishes itself from existing ToM evaluations in several critical ways. First, it provides a more direct and principled approach to assessing LMs' comprehension of belief \emph{vs.} knowledge statements. By focusing on this fundamental epistemic distinction, we probe a crucial precursor to more complex ToM reasoning. This allows for a more granular analysis of LMs' capabilities, potentially revealing insights into their underlying mechanisms for processing epistemic concepts. Second, our work introduces a novel testing suite with an extensive quantity of original content, deliberately avoiding well-known and well-documented ToM tasks like the Sally-Anne test~\citep{baron1985does}, which may have been encountered during training. This design choice mitigates the risk of performance inflation due to memorization or overfitting to specific task structures. Instead, it tests models under fresh and atomic scenarios, providing a more robust and simple assessment of their generalization capabilities in epistemic and social reasoning. Third, our work's inclusion of both factual and false examples in the context of knowledge and belief adds a layer of complexity absent in many existing evaluations. This feature allows for a nuanced exploration of how LMs handle truth values in relation to epistemic states, shedding light on their capacity for reasoning in ``contrafactual'' scenarios, which is crucial in studying advanced ToM abilities~\citep[{see also}:][]{wu2024reasoning}. By addressing these aspects, our work thus seeks to provide a more comprehensive and illuminating picture of modern LMs' epistemic processing and social reasoning capabilities.

%% file: sections/limitations.tex
\section{Limitations and Future Directions}
\label{sec:limitations}
Despite the valuable insights offered by this study, several limitations must be acknowledged. These pertain primarily to the linguistic complexity of epistemic reasoning, the design of our experimental tasks, and the inherent challenges in evaluating LMs on epistemic distinctions. By recognizing these constraints, we can better understand the boundaries of our findings and identify areas for future research.

\textbf{Linguistic complexity and pragmatic considerations.}
One of the central limitations of our study is the oversimplification of the linguistic nuances surrounding epistemic phrases such as ``know'' and ``believe.'' While our experiments focused on a core set of epistemic terms, we did not sufficiently account for the flexibility and contextual fluidity with which these phrases are used in everyday language. In many colloquial contexts, the word ``know'' is used interchangeably with ``believe'' or ``think.'' People often say, ``I know'' when they mean ``I believe'' or ``I think,'' introducing ambiguity in the relationship between the speaker's epistemic stance and the literal truth of the statement.

Furthermore, epistemic utterances are not always straightforward in their intent. They can be performative, rhetorical, sarcastic, or satirical. For example, when someone says, ``I know the sky is green,'' the statement may not reflect their genuine belief but could instead be a satirical comment meant to challenge an interlocutor’s assumption. LMs, however, are often not equipped to detect these pragmatic subtleties, leading to erroneous conclusions about what constitutes belief, knowledge, or fact. This pragmatic gap limits the models' capacity to engage with language as it is used in real-world human interactions, and future studies should address this issue by incorporating more diverse linguistic contexts.

\textbf{Specific focus on select epistemic phrases.}
Our study primarily focused on the epistemic phrases ``know,'' ``believe,'' and ``think,'' overlooking other common epistemic expressions such as ``I know,'' ``I feel,'' ``I suppose,'' ``I gather,'' ``I imagine,'' and ``I presume.'' Each of these phrases conveys varying degrees of epistemic commitment and subjective belief, and their inclusion could provide a more comprehensive view of how LMs handle epistemic reasoning. By limiting our analysis to a small subset of epistemic phrases, we may have restricted our understanding of the models’ broader capabilities in processing and distinguishing between different levels of certainty and belief. A more expansive exploration of the full range of epistemic expressions could offer deeper insights into how LMs manage different shades of epistemological reasoning, particularly in more complex, context-sensitive scenarios.

\textbf{Influence of complementizer usage.}
Another linguistic issue relates to the presence or absence of the English complementizer ``that'' in the statements we presented to the models. Research has shown that the inclusion or omission of ``that'' can subtly influence the interpretation of epistemic authority~\citep{tollan2024does}. For instance, the phrase ``I know that the Earth orbits the Sun'' may be interpreted differently from ``I know the Earth orbits the Sun,'' with the latter potentially perceived as more assertive or informal. Although we attempted to standardize the structure of our statements, the decision to include the complementizer may have had some unintended effects on model performance. This variability in sentence structure could introduce noise into the results and confound the models’ ability to accurately process epistemic claims. Future studies should more rigorously examine how different syntactic structures influence model reasoning in epistemic contexts.

\textbf{Limited exploration of contextualism.}
Context plays a critical role in how epistemic statements are interpreted. In everyday conversations, the meaning of phrases like ``I know'' or ``I believe'' can change depending on the social, cultural, or conversational context in which they are uttered. However, our study did not fully account for the contextualism inherent in epistemic reasoning. LMs often rely heavily on the immediate textual input without fully integrating broader contextual cues that might inform the speaker's intent or epistemic state. For example, in legal or medical settings, ``knowing'' carries a far more stringent epistemic burden than in casual conversation. This lack of contextual sensitivity in our evaluation may have led to an oversimplified view of how LMs handle complex epistemic tasks. Addressing this issue would require more sophisticated, multi-modal analyses that incorporate not only linguistic inputs but also situational, social, and cultural cues that shape how epistemic phrases are understood.

\textbf{Task design and model generalization.}
Our task design, while comprehensive, may have limited the generalizability of our findings. The tasks we employed, although carefully crafted to evaluate models’ epistemic reasoning, were designed as isolated test cases, often lacking the rich, dynamic conversational flow that would be present in real-world scenarios. Consequently, the performance of LMs in these controlled settings may not directly translate to more complex, open-ended dialogues where epistemic distinctions are embedded in multi-turn conversations. This gap between task-based evaluation and real-world application remains a challenge for all current AI research and should be addressed in future studies by incorporating more interactive, dialog-based evaluations that better reflect the nuanced nature of human epistemic reasoning.

\textbf{Impact of zero-shot CoT on performance.}  To explore ways to enhance models' epistemic reasoning, we applied the zero-shot CoT prompting method~\citep{kojima2022zero} to GPT-4o. This technique, which involves prefacing responses with ``Let's think step by step,'' was designed to encourage the model to engage in more deliberate and structured reasoning. The outcomes of this intervention were mixed and somewhat inconclusive.  In tasks focused on direct fact verification and first-person belief verification, the model's performance dropped by 6\% and 4\%, respectively, indicating that CoT prompting did not enhance reasoning in these areas.  In contrast, this approach showed more promise in tasks involving complex reasoning. In the confirmation of first-person beliefs, especially in false scenarios, performance increased by 11\%, leading to an overall 6\% improvement in first-person belief reasoning. Additionally, a 10\% performance boost was observed in factual scenarios related to recursive knowledge awareness, although recursive knowledge confirmation only saw a marginal 0.4\% increase. 

Despite these gains, the model continued to display some of the earlier weaknesses, such as its difficulty recognizing that individuals can hold false beliefs. These findings suggest that while zero-shot CoT prompting can enhance performance in certain complex, non-verification tasks, it is still not a panacea for the deeper epistemic limitations observed in LMs. More advanced strategies may be needed to fully address these challenges and improve the model’s overall reasoning capabilities.

%% file: neurips_data_2023.bbl
\begin{thebibliography}{111}
\providecommand{\natexlab}[1]{#1}
\providecommand{\url}[1]{\texttt{#1}}
\expandafter\ifx\csname urlstyle\endcsname\relax
  \providecommand{\doi}[1]{doi: #1}\else
  \providecommand{\doi}{doi: \begingroup \urlstyle{rm}\Url}\fi

\bibitem[Chuey et~al.(2024)Chuey, Luo, and Markman]{chuey2024epistemic}
Aaron Chuey, Yiwei Luo, and Ellen~M Markman.
\newblock Epistemic language in news headlines shapes readers’ perceptions of objectivity.
\newblock \emph{Proceedings of the National Academy of Sciences}, 121\penalty0 (20):\penalty0 e2314091121, 2024.

\bibitem[Vukovic(2014)]{vukovic2014strong}
Milica Vukovic.
\newblock Strong epistemic modality in parliamentary discourse.
\newblock \emph{Open Linguistics}, 1\penalty0 (1), 2014.

\bibitem[Ekstr{\"o}m et~al.(2021)Ekstr{\"o}m, Rams{\"a}lv, and Westlund]{ekstrom2021epistemologies}
Mats Ekstr{\"o}m, Amanda Rams{\"a}lv, and Oscar Westlund.
\newblock The epistemologies of breaking news.
\newblock \emph{Journalism Studies}, 22\penalty0 (2):\penalty0 174--192, 2021.

\bibitem[Levine et~al.(2024)Levine, Tuwani, Kompa, Varma, Finlayson, Mehrotra, and Beam]{levine2024diagnostic}
David~M Levine, Rudraksh Tuwani, Benjamin Kompa, Amita Varma, Samuel~G Finlayson, Ateev Mehrotra, and Andrew Beam.
\newblock The diagnostic and triage accuracy of the gpt-3 artificial intelligence model: an observational study.
\newblock \emph{The Lancet Digital Health}, 6\penalty0 (8):\penalty0 e555--e561, 2024.

\bibitem[Kim et~al.(2024{\natexlab{a}})Kim, Leonte, Chen, Torous, Linos, Pinto, and Rodriguez]{kim2024large}
Jiyeong Kim, Kimberly~G Leonte, Michael~L Chen, John~B Torous, Eleni Linos, Anthony Pinto, and Carolyn~I Rodriguez.
\newblock Large language models outperform mental and medical health care professionals in identifying obsessive-compulsive disorder.
\newblock \emph{NPJ Digital Medicine}, 7\penalty0 (1):\penalty0 193, 2024{\natexlab{a}}.

\bibitem[Singhal et~al.(2023)Singhal, Azizi, Tu, Mahdavi, Wei, Chung, Scales, Tanwani, Cole-Lewis, Pfohl, et~al.]{singhal2023large}
Karan Singhal, Shekoofeh Azizi, Tao Tu, S~Sara Mahdavi, Jason Wei, Hyung~Won Chung, Nathan Scales, Ajay Tanwani, Heather Cole-Lewis, Stephen Pfohl, et~al.
\newblock Large language models encode clinical knowledge.
\newblock \emph{Nature}, 620\penalty0 (7972):\penalty0 172--180, 2023.

\bibitem[Xu et~al.(2024{\natexlab{a}})Xu, Yao, Dong, Gabriel, Yu, Hendler, Ghassemi, Dey, and Wang]{xu2024_acm}
Xuhai Xu, Bingsheng Yao, Yuanzhe Dong, Saadia Gabriel, Hong Yu, James Hendler, Marzyeh Ghassemi, Anind~K. Dey, and Dakuo Wang.
\newblock Mental-llm: Leveraging large language models for mental health prediction via online text data.
\newblock \emph{Proc. ACM Interact. Mob. Wearable Ubiquitous Technol.}, 8\penalty0 (1), mar 2024{\natexlab{a}}.
\newblock \doi{10.1145/3643540}.
\newblock URL \url{https://doi.org/10.1145/3643540}.

\bibitem[Demszky et~al.(2023{\natexlab{a}})Demszky, Yang, Yeager, Bryan, Clapper, Chandhok, Eichstaedt, Hecht, Jamieson, Johnson, et~al.]{demszky2023using}
Dorottya Demszky, Diyi Yang, David~S Yeager, Christopher~J Bryan, Margarett Clapper, Susannah Chandhok, Johannes~C Eichstaedt, Cameron Hecht, Jeremy Jamieson, Meghann Johnson, et~al.
\newblock Using large language models in psychology.
\newblock \emph{Nature Reviews Psychology}, 2\penalty0 (11):\penalty0 688--701, 2023{\natexlab{a}}.

\bibitem[Eliot(2023)]{eliot2023people}
Lance Eliot.
\newblock People are eagerly consulting generative ai chatgpt for mental health advice, stressing out ai ethics and ai law.
\newblock \emph{Forbes. Available online at: https://www. forbes. com/sites/deloitte/2024/02/16/making-the-leap-from-smart-to-themetaverse-in-operations}, 2023.

\bibitem[Wilhelm et~al.(2023)Wilhelm, Roos, and Kaczmarczyk]{wilhelm2023large}
Theresa~Isabelle Wilhelm, Jonas Roos, and Robert Kaczmarczyk.
\newblock Large language models for therapy recommendations across 3 clinical specialties: comparative study.
\newblock \emph{Journal of medical Internet research}, 25:\penalty0 e49324, 2023.

\bibitem[Salmi et~al.(2022)Salmi, M{\'e}relle, Gilissen, van~der Mei, and Bhulai]{salmi2022detecting}
Salim Salmi, Saskia M{\'e}relle, Renske Gilissen, Rob van~der Mei, and Sandjai Bhulai.
\newblock Detecting changes in help seeker conversations on a suicide prevention helpline during the covid- 19 pandemic: in-depth analysis using encoder representations from transformers.
\newblock \emph{BMC public health}, 22\penalty0 (1):\penalty0 530, 2022.

\bibitem[Magesh et~al.(2024)Magesh, Surani, Dahl, Suzgun, Manning, and Ho]{magesh2024hallucination}
Varun Magesh, Faiz Surani, Matthew Dahl, Mirac Suzgun, Christopher~D Manning, and Daniel~E Ho.
\newblock Hallucination-free? assessing the reliability of leading ai legal research tools.
\newblock \emph{arXiv preprint arXiv:2405.20362}, 2024.

\bibitem[Dahl et~al.(2024)Dahl, Magesh, Suzgun, and Ho]{dahl2024large}
Matthew Dahl, Varun Magesh, Mirac Suzgun, and Daniel~E Ho.
\newblock Large legal fictions: Profiling legal hallucinations in large language models.
\newblock \emph{arXiv preprint arXiv:2401.01301}, 2024.

\bibitem[Guha et~al.(2024)Guha, Nyarko, Ho, R{\'e}, Chilton, Chohlas-Wood, Peters, Waldon, Rockmore, Zambrano, et~al.]{guha2024legalbench}
Neel Guha, Julian Nyarko, Daniel Ho, Christopher R{\'e}, Adam Chilton, Alex Chohlas-Wood, Austin Peters, Brandon Waldon, Daniel Rockmore, Diego Zambrano, et~al.
\newblock Legalbench: A collaboratively built benchmark for measuring legal reasoning in large language models.
\newblock \emph{Advances in Neural Information Processing Systems}, 36, 2024.

\bibitem[Henry(2024)]{Henry2024}
Justin Henry.
\newblock We {{Asked Every Am Law}} 100 {{Law Firm How They}}'re {{Using Gen AI}}. {{Here}}'s {{What We Learned}}.
\newblock \emph{The American Lawyer}, January 2024.
\newblock URL \url{https://www.law.com/americanlawyer/2024/01/29/we-asked-every-am-law-100-firm-how-theyre-using-gen-ai-heres-what-we-learned/}.

\bibitem[Collens et~al.(2024)Collens, Reimer, Schifman, and Wilkinson]{Collens2024}
Jack Collens, Rachel Reimer, Gerald Schifman, and Pamela Wilkinson.
\newblock {{AI Survey}}: {{Where Artificial Intelligence Stands}} in the {{Legal Industry}}, April 2024.
\newblock URL \url{https://assets.law360news.com/1826000/1826128/law360_pulse-ai_survey.pdf}.

\bibitem[Caswell(2023)]{caswell2023}
David Caswell.
\newblock {AI and journalism: What's next?}, 2023.
\newblock URL \url{https://reutersinstitute.politics.ox.ac.uk/news/ai-and-journalism-whats-next}.

\bibitem[Hurst(2023)]{hurst2023}
Luke Hurst.
\newblock Robot reporters? here’s how news organisations are using ai in journalism.
\newblock \emph{Euronews}, 2023.
\newblock URL \url{https://www.euronews.com/next/2023/08/24/robot-reporters-heres-how-news-organisations-are-using-ai-in-journalism}.

\bibitem[Sweney(2023)]{sweney2023mirror}
M~Sweney.
\newblock Mirror and express owner publishes first articles written using ai.
\newblock \emph{The Guardian}, 2023.
\newblock URL \url{https://www.theguardian.com/business/2023/mar/07/mirror-and-express-owner-publishes-first-articles-written-using-ai}.

\bibitem[Wang et~al.(2024)Wang, Ribeiro, Robinson, Loeb, and Demszky]{wang2024tutorcopilothumanaiapproach}
Rose~E. Wang, Ana~T. Ribeiro, Carly~D. Robinson, Susanna Loeb, and Dora Demszky.
\newblock Tutor copilot: A human-ai approach for scaling real-time expertise.
\newblock \emph{arXiv preprint arXiv:2410.03017}, 2024.

\bibitem[Extance(2023)]{extance2023chatgpt}
Andy Extance.
\newblock Chatgpt has entered the classroom: how llms could transform education.
\newblock \emph{Nature}, 623\penalty0 (7987):\penalty0 474--477, 2023.

\bibitem[Demszky and Liu(2023)]{demszky2023m}
Dorottya Demszky and Jing Liu.
\newblock M-powering teachers: Natural language processing powered feedback improves 1: 1 instruction and student outcomes.
\newblock In \emph{Proceedings of the Tenth ACM Conference on Learning@ Scale}, pages 59--69, 2023.

\bibitem[Demszky et~al.(2023{\natexlab{b}})Demszky, Liu, Hill, Jurafsky, and Piech]{demszky2023can}
Dorottya Demszky, Jing Liu, Heather~C Hill, Dan Jurafsky, and Chris Piech.
\newblock Can automated feedback improve teachers’ uptake of student ideas? evidence from a randomized controlled trial in a large-scale online course.
\newblock \emph{Educational Evaluation and Policy Analysis}, page 01623737231169270, 2023{\natexlab{b}}.

\bibitem[Castelvecchi(2024)]{castelvecchi2024}
Davide Castelvecchi.
\newblock {Researchers built an 'AI Scientist' -- what can it do?}
\newblock \emph{Nature}, 2024.
\newblock URL \url{https://www.nature.com/articles/d41586-024-02842-3}.

\bibitem[Baek et~al.(2024)Baek, Jauhar, Cucerzan, and Hwang]{baek2024researchagent}
Jinheon Baek, Sujay~Kumar Jauhar, Silviu Cucerzan, and Sung~Ju Hwang.
\newblock Researchagent: Iterative research idea generation over scientific literature with large language models.
\newblock \emph{arXiv preprint arXiv:2404.07738}, 2024.

\bibitem[Lu et~al.(2024)Lu, Lu, Lange, Foerster, Clune, and Ha]{lu2024ai}
Chris Lu, Cong Lu, Robert~Tjarko Lange, Jakob Foerster, Jeff Clune, and David Ha.
\newblock The ai scientist: Towards fully automated open-ended scientific discovery.
\newblock \emph{arXiv preprint arXiv:2408.06292}, 2024.

\bibitem[Kim et~al.(2024{\natexlab{b}})Kim, Muhn, and Nikolaev]{kim2024financial}
Alex Kim, Maximilian Muhn, and Valeri Nikolaev.
\newblock Financial statement analysis with large language models.
\newblock \emph{arXiv preprint arXiv:2407.17866}, 2024{\natexlab{b}}.

\bibitem[Wu et~al.(2023{\natexlab{a}})Wu, Irsoy, Lu, Dabravolski, Dredze, Gehrmann, Kambadur, Rosenberg, and Mann]{wu2023bloomberggpt}
Shijie Wu, Ozan Irsoy, Steven Lu, Vadim Dabravolski, Mark Dredze, Sebastian Gehrmann, Prabhanjan Kambadur, David Rosenberg, and Gideon Mann.
\newblock Bloomberggpt: A large language model for finance.
\newblock \emph{arXiv preprint arXiv:2303.17564}, 2023{\natexlab{a}}.

\bibitem[Liu et~al.(2023{\natexlab{a}})Liu, Wang, Yang, and Zha]{liu2023fingpt}
Xiao-Yang Liu, Guoxuan Wang, Hongyang Yang, and Daochen Zha.
\newblock Fin{GPT}: Democratizing internet-scale data for financial large language models.
\newblock In \emph{NeurIPS 2023 Workshop on Instruction Tuning and Instruction Following}, 2023{\natexlab{a}}.
\newblock URL \url{https://openreview.net/forum?id=5BqWC1Fz8F}.

\bibitem[Lopez-Lira and Tang(2023)]{lopez2023can}
Alejandro Lopez-Lira and Yuehua Tang.
\newblock Can chatgpt forecast stock price movements? return predictability and large language models.
\newblock \emph{arXiv preprint arXiv:2304.07619}, 2023.

\bibitem[Robson(2024)]{bbc_ai_relationship2024}
David Robson.
\newblock Ai wisdom: What happens when you ask an algorithm for relationship advice.
\newblock 2024.
\newblock URL \url{https://www.bbc.com/future/article/20240515-ai-wisdom-what-happens-when-you-ask-an-algorithm-for-relationship-advice}.

\bibitem[Nguyen(2023)]{nguyen2023}
Britney Nguyen.
\newblock I asked chatgpt for online and in-person dating advice, here's what relationship coaches think of its answers.
\newblock 2023.
\newblock URL \url{https://www.businessinsider.com/what-real-life-relationship-coaches-think-chatgpts-dating-advice-2023-1}.

\bibitem[Liu et~al.(2023{\natexlab{b}})Liu, Yen, Marjieh, Griffiths, and Krishna]{liu2023improving}
Ryan Liu, Howard Yen, Raja Marjieh, Thomas~L Griffiths, and Ranjay Krishna.
\newblock Improving interpersonal communication by simulating audiences with language models.
\newblock \emph{arXiv preprint arXiv:2311.00687}, 2023{\natexlab{b}}.

\bibitem[Sap et~al.(2022)Sap, Le~Bras, Fried, and Choi]{sap-etal-2022-neural}
Maarten Sap, Ronan Le~Bras, Daniel Fried, and Yejin Choi.
\newblock Neural theory-of-mind? on the limits of social intelligence in large {LM}s.
\newblock In Yoav Goldberg, Zornitsa Kozareva, and Yue Zhang, editors, \emph{Proceedings of the 2022 Conference on Empirical Methods in Natural Language Processing}, pages 3762--3780, Abu Dhabi, United Arab Emirates, December 2022. Association for Computational Linguistics.
\newblock \doi{10.18653/v1/2022.emnlp-main.248}.
\newblock URL \url{https://aclanthology.org/2022.emnlp-main.248}.

\bibitem[Gandhi et~al.(2023{\natexlab{a}})Gandhi, Fr{\"a}nken, Gerstenberg, and Goodman]{gandhi2023understanding}
Kanishk Gandhi, Jan-Philipp Fr{\"a}nken, Tobias Gerstenberg, and Noah Goodman.
\newblock Understanding social reasoning in language models with language models.
\newblock In \emph{Thirty-seventh Conference on Neural Information Processing Systems Datasets and Benchmarks Track}, 2023{\natexlab{a}}.
\newblock URL \url{https://openreview.net/forum?id=8bqjirgxQM}.

\bibitem[Kosinski(2023)]{kosinski2023theory}
Michal Kosinski.
\newblock Theory of mind might have spontaneously emerged in large language models.
\newblock \emph{arXiv preprint arXiv:2302.02083}, 2023.

\bibitem[Ullman(2023)]{ullman2023large}
Tomer Ullman.
\newblock Large language models fail on trivial alterations to theory-of-mind tasks.
\newblock \emph{arXiv preprint arXiv:2302.08399}, 2023.

\bibitem[Shapira et~al.(2024)Shapira, Levy, Alavi, Zhou, Choi, Goldberg, Sap, and Shwartz]{shapira-etal-2024-clever}
Natalie Shapira, Mosh Levy, Seyed~Hossein Alavi, Xuhui Zhou, Yejin Choi, Yoav Goldberg, Maarten Sap, and Vered Shwartz.
\newblock Clever hans or neural theory of mind? stress testing social reasoning in large language models.
\newblock In Yvette Graham and Matthew Purver, editors, \emph{Proceedings of the 18th Conference of the European Chapter of the Association for Computational Linguistics (Volume 1: Long Papers)}, pages 2257--2273, St. Julian{'}s, Malta, March 2024. Association for Computational Linguistics.
\newblock URL \url{https://aclanthology.org/2024.eacl-long.138}.

\bibitem[Bubeck et~al.(2023)Bubeck, Chandrasekaran, Eldan, Gehrke, Horvitz, Kamar, Lee, Lee, Li, Lundberg, et~al.]{bubeck2023sparks}
S{\'e}bastien Bubeck, Varun Chandrasekaran, Ronen Eldan, Johannes Gehrke, Eric Horvitz, Ece Kamar, Peter Lee, Yin~Tat Lee, Yuanzhi Li, Scott Lundberg, et~al.
\newblock Sparks of artificial general intelligence: Early experiments with gpt-4.
\newblock \emph{arXiv preprint arXiv:2303.12712}, 2023.

\bibitem[Trott et~al.(2023)Trott, Jones, Chang, Michaelov, and Bergen]{trott2023large}
Sean Trott, Cameron Jones, Tyler Chang, James Michaelov, and Benjamin Bergen.
\newblock Do large language models know what humans know?
\newblock \emph{Cognitive Science}, 47\penalty0 (7):\penalty0 e13309, 2023.

\bibitem[Aru et~al.(2023)Aru, Labash, Corcoll, and Vicente]{aru2023mind}
Jaan Aru, Aqeel Labash, Oriol Corcoll, and Raul Vicente.
\newblock Mind the gap: Challenges of deep learning approaches to theory of mind.
\newblock \emph{Artificial Intelligence Review}, 56\penalty0 (9):\penalty0 9141--9156, 2023.

\bibitem[Mahowald et~al.(2024)Mahowald, Ivanova, Blank, Kanwisher, Tenenbaum, and Fedorenko]{mahowald2024dissociating}
Kyle Mahowald, Anna~A Ivanova, Idan~A Blank, Nancy Kanwisher, Joshua~B Tenenbaum, and Evelina Fedorenko.
\newblock Dissociating language and thought in large language models.
\newblock \emph{Trends in Cognitive Sciences}, 2024.
\newblock URL \url{https://web.mit.edu/bcs/nklab/media/pdfs/Mahowald.TICs2024.pdf}.

\bibitem[Le et~al.(2019)Le, Boureau, and Nickel]{le2019revisiting}
Matthew Le, Y-Lan Boureau, and Maximilian Nickel.
\newblock Revisiting the evaluation of theory of mind through question answering.
\newblock In \emph{Proceedings of the 2019 Conference on Empirical Methods in Natural Language Processing and the 9th International Joint Conference on Natural Language Processing (EMNLP-IJCNLP)}, pages 5872--5877, 2019.

\bibitem[Ma et~al.(2023{\natexlab{a}})Ma, Gao, and Xu]{Ma2023ToMChallengesAP}
Xiaomeng Ma, Lingyu Gao, and Qihui Xu.
\newblock Tomchallenges: A principle-guided dataset and diverse evaluation tasks for exploring theory of mind.
\newblock In \emph{Conference on Computational Natural Language Learning}, 2023{\natexlab{a}}.
\newblock URL \url{https://api.semanticscholar.org/CorpusID:258865295}.

\bibitem[Gandhi et~al.(2023{\natexlab{b}})Gandhi, Franken, Gerstenberg, and Goodman]{Gandhi2023UnderstandingSR}
Kanishk Gandhi, Jan-Philipp Franken, Tobias Gerstenberg, and Noah~D. Goodman.
\newblock Understanding social reasoning in language models with language models.
\newblock \emph{ArXiv}, abs/2306.15448, 2023{\natexlab{b}}.
\newblock URL \url{https://api.semanticscholar.org/CorpusID:259262573}.

\bibitem[Wu et~al.(2023{\natexlab{b}})Wu, He, Jia, Mihalcea, Chen, and Deng]{wu-etal-2023-hi-tom}
Yufan Wu, Yinghui He, Yilin Jia, Rada Mihalcea, Yulong Chen, and Naihao Deng.
\newblock Hi-{T}o{M}: A benchmark for evaluating higher-order theory of mind reasoning in large language models.
\newblock In Houda Bouamor, Juan Pino, and Kalika Bali, editors, \emph{Findings of the Association for Computational Linguistics: EMNLP 2023}, pages 10691--10706, Singapore, December 2023{\natexlab{b}}. Association for Computational Linguistics.
\newblock \doi{10.18653/v1/2023.findings-emnlp.717}.
\newblock URL \url{https://aclanthology.org/2023.findings-emnlp.717}.

\bibitem[Jones et~al.(2023)Jones, Trott, and Bergen]{jones2023epitome}
Cameron~Robert Jones, Sean Trott, and Ben Bergen.
\newblock {EPITOME}: Experimental protocol inventory for theory of mind evaluation.
\newblock In \emph{First Workshop on Theory of Mind in Communicating Agents}, 2023.
\newblock URL \url{https://openreview.net/forum?id=e5Yky8Fnvj}.

\bibitem[Zhou et~al.(2024)Zhou, Madaan, Potharaju, Gupta, McKee, Holtzman, Pujara, Ren, Mishra, Nematzadeh, Upadhyay, and Faruqui]{zhou2024T4D}
Pei Zhou, Aman Madaan, Srividya~Pranavi Potharaju, Aditya Gupta, Kevin~R. McKee, Ari Holtzman, Jay Pujara, Xiang Ren, Swaroop Mishra, Aida Nematzadeh, Shyam Upadhyay, and Manaal Faruqui.
\newblock How far are large language models from agents with theory-of-mind?, 2024.
\newblock URL \url{https://openreview.net/forum?id=xnUIMz5u2s}.

\bibitem[Xu et~al.(2024{\natexlab{b}})Xu, Zhao, Zhu, Du, and He]{xu-etal-2024-opentom}
Hainiu Xu, Runcong Zhao, Lixing Zhu, Jinhua Du, and Yulan He.
\newblock {O}pen{T}o{M}: A comprehensive benchmark for evaluating theory-of-mind reasoning capabilities of large language models.
\newblock In Lun-Wei Ku, Andre Martins, and Vivek Srikumar, editors, \emph{Proceedings of the 62nd Annual Meeting of the Association for Computational Linguistics (Volume 1: Long Papers)}, pages 8593--8623, Bangkok, Thailand, August 2024{\natexlab{b}}. Association for Computational Linguistics.
\newblock URL \url{https://aclanthology.org/2024.acl-long.466}.

\bibitem[Wu et~al.(2024)Wu, Qiu, Ross, Aky{\"u}rek, Chen, Wang, Kim, Andreas, and Kim]{wu2024reasoning}
Zhaofeng Wu, Linlu Qiu, Alexis Ross, Ekin Aky{\"u}rek, Boyuan Chen, Bailin Wang, Najoung Kim, Jacob Andreas, and Yoon Kim.
\newblock Reasoning or reciting? exploring the capabilities and limitations of language models through counterfactual tasks.
\newblock In \emph{Proceedings of the 2024 Conference of the North American Chapter of the Association for Computational Linguistics: Human Language Technologies (Volume 1: Long Papers)}, pages 1819--1862, 2024.

\bibitem[Basmov et~al.(2024)Basmov, Goldberg, and Tsarfaty]{basmov2024llms}
Victoria Basmov, Yoav Goldberg, and Reut Tsarfaty.
\newblock Llms' reading comprehension is affected by parametric knowledge and struggles with hypothetical statements.
\newblock \emph{arXiv preprint arXiv:2404.06283}, 2024.

\bibitem[Basmov et~al.(2023)Basmov, Goldberg, and Tsarfaty]{basmov2023simple}
Victoria Basmov, Yoav Goldberg, and Reut Tsarfaty.
\newblock Simple linguistic inferences of large language models (llms): Blind spots and blinds.
\newblock \emph{arXiv preprint arXiv:2305.14785}, 2023.

\bibitem[Holliday and Mandelkern(2024)]{holliday2024conditional}
Wesley~H Holliday and Matthew Mandelkern.
\newblock Conditional and modal reasoning in large language models.
\newblock \emph{arXiv preprint arXiv:2401.17169}, 2024.

\bibitem[OpenAI(2024)]{gpt4o}
OpenAI.
\newblock {Hello GPT-4o}, 2024.
\newblock URL \url{https://openai.com/index/hello-gpt-4o/}.

\bibitem[{OpenAI}(2023{\natexlab{a}})]{GPT4TechnicalReport}
{OpenAI}.
\newblock {{GPT-4 Technical Report}}, March 2023{\natexlab{a}}.
\newblock URL \url{https://arxiv.org/abs/2303.08774}.

\bibitem[{OpenAI}(2023{\natexlab{b}})]{ChatGPT}
{OpenAI}.
\newblock Introducing {{ChatGPT}}, November 2023{\natexlab{b}}.
\newblock URL \url{https://openai.com/blog/chatgpt}.

\bibitem[Anthropic(2024)]{AnthropicClaude3}
Anthropic.
\newblock The {C}laude 3 {M}odel {F}amily: {O}pus, {S}onnet, {H}aiku, March 2024.
\newblock URL \url{https://www-cdn.anthropic.com/de8ba9b01c9ab7cbabf5c33b80b7bbc618857627/Model_Card_Claude_3.pdf}.

\bibitem[AI@Meta(2024)]{llama3modelcard}
AI@Meta.
\newblock {Llama 3 Model Card}, 2024.
\newblock URL \url{https://github.com/meta-llama/llama3/blob/main/MODEL_CARD.md}.

\bibitem[Touvron et~al.(2023)Touvron, Martin, Stone, Albert, Almahairi, Babaei, Bashlykov, Batra, Bhargava, Bhosale, et~al.]{touvron2023llama}
Hugo Touvron, Louis Martin, Kevin Stone, Peter Albert, Amjad Almahairi, Yasmine Babaei, Nikolay Bashlykov, Soumya Batra, Prajjwal Bhargava, Shruti Bhosale, et~al.
\newblock Llama 2: Open foundation and fine-tuned chat models.
\newblock \emph{arXiv preprint arXiv:2307.09288}, 2023.

\bibitem[Jiang et~al.(2023)Jiang, Sablayrolles, Mensch, Bamford, Chaplot, Casas, Bressand, Lengyel, Lample, Saulnier, et~al.]{jiang2023mistral}
Albert~Q Jiang, Alexandre Sablayrolles, Arthur Mensch, Chris Bamford, Devendra~Singh Chaplot, Diego de~las Casas, Florian Bressand, Gianna Lengyel, Guillaume Lample, Lucile Saulnier, et~al.
\newblock Mistral 7b.
\newblock \emph{arXiv preprint arXiv:2310.06825}, 2023.

\bibitem[Jiang et~al.(2024)Jiang, Sablayrolles, Roux, Mensch, Savary, Bamford, Chaplot, Casas, Hanna, Bressand, et~al.]{jiang2024mixtral}
Albert~Q Jiang, Alexandre Sablayrolles, Antoine Roux, Arthur Mensch, Blanche Savary, Chris Bamford, Devendra~Singh Chaplot, Diego de~las Casas, Emma~Bou Hanna, Florian Bressand, et~al.
\newblock Mixtral of experts.
\newblock \emph{arXiv preprint arXiv:2401.04088}, 2024.

\bibitem[Suzgun et~al.(2024)Suzgun, Shieber, and Jurafsky]{suzgun-etal-2024-string2string}
Mirac Suzgun, Stuart Shieber, and Dan Jurafsky.
\newblock string2string: A modern python library for string-to-string algorithms.
\newblock In Yixin Cao, Yang Feng, and Deyi Xiong, editors, \emph{Proceedings of the 62nd Annual Meeting of the Association for Computational Linguistics (Volume 3: System Demonstrations)}, pages 278--285, Bangkok, Thailand, August 2024. Association for Computational Linguistics.
\newblock URL \url{https://aclanthology.org/2024.acl-demos.26}.

\bibitem[Hu et~al.(2024)Hu, Chen, Li, Guo, Wen, Yu, and Guo]{hu2024towards}
Xuming Hu, Junzhe Chen, Xiaochuan Li, Yufei Guo, Lijie Wen, Philip~S. Yu, and Zhijiang Guo.
\newblock Towards understanding factual knowledge of large language models.
\newblock In \emph{The Twelfth International Conference on Learning Representations}, 2024.
\newblock URL \url{https://openreview.net/forum?id=9OevMUdods}.

\bibitem[Hacquard and Lidz(2022)]{HacquardLidz}
Valentine Hacquard and Jeffrey Lidz.
\newblock On the acquisition of attitude verbs.
\newblock \emph{Annual Review of Linguistics}, 8:\penalty0 193–212, 2022.

\bibitem[Phillips et~al.(2021)Phillips, Buckwalter, Cushman, Friedman, Martin, Turri, Santos, and Knobe]{phillips2021knowledge}
Jonathan Phillips, Wesley Buckwalter, Fiery Cushman, Ori Friedman, Alia Martin, John Turri, Laurie Santos, and Joshua Knobe.
\newblock Knowledge before belief.
\newblock \emph{Behavioral and Brain Sciences}, 44:\penalty0 e140, 2021.

\bibitem[Grice(1957)]{Grice1957}
H.~P. Grice.
\newblock Meaning.
\newblock \emph{The Philosophical Review}, 66\penalty0 (3):\penalty0 377--388, 1957.

\bibitem[Gettier(1963)]{gettier1963justified}
Edmund~L Gettier.
\newblock Is justified true belief knowledge?
\newblock \emph{Analysis}, 23\penalty0 (6):\penalty0 121--123, 1963.

\bibitem[Armstrong(1973)]{armstrong1973belief}
David~Malet Armstrong.
\newblock \emph{Belief, truth and knowledge}.
\newblock Cambridge University Press, 1973.

\bibitem[Armstrong(1969)]{armstrong1969does}
David~M Armstrong.
\newblock Does knowledge entail belief?
\newblock In \emph{Proceedings of the Aristotelian Society}, volume~70, pages 21--36. JSTOR, 1969.

\bibitem[Ichikawa and Steup(2018)]{sep-knowledge-analysis}
Jonathan~Jenkins Ichikawa and Matthias Steup.
\newblock {The Analysis of Knowledge}.
\newblock In Edward~N. Zalta, editor, \emph{The {Stanford} Encyclopedia of Philosophy}. Metaphysics Research Lab, Stanford University, {S}ummer 2018 edition, 2018.

\bibitem[Luper(2020)]{sep-closure-epistemic}
Steven Luper.
\newblock {Epistemic Closure}.
\newblock In Edward~N. Zalta, editor, \emph{The {Stanford} Encyclopedia of Philosophy}. Metaphysics Research Lab, Stanford University, {S}ummer 2020 edition, 2020.

\bibitem[Hazlett(2010)]{Hazlett2010-HAZTMO}
Allan Hazlett.
\newblock The myth of factive verbs.
\newblock \emph{Philosophy and Phenomenological Research}, 80\penalty0 (3):\penalty0 497--522, 2010.
\newblock \doi{10.1111/j.1933-1592.2010.00338.x}.

\bibitem[Brown et~al.(2020)Brown, Mann, Ryder, Subbiah, Kaplan, Dhariwal, Neelakantan, Shyam, Sastry, Askell, Agarwal, Herbert{-}Voss, Krueger, Henighan, Child, Ramesh, Ziegler, Wu, Winter, Hesse, Chen, Sigler, Litwin, Gray, Chess, Clark, Berner, McCandlish, Radford, Sutskever, and Amodei]{brown2020language}
Tom~B. Brown, Benjamin Mann, Nick Ryder, Melanie Subbiah, Jared Kaplan, Prafulla Dhariwal, Arvind Neelakantan, Pranav Shyam, Girish Sastry, Amanda Askell, Sandhini Agarwal, Ariel Herbert{-}Voss, Gretchen Krueger, Tom Henighan, Rewon Child, Aditya Ramesh, Daniel~M. Ziegler, Jeffrey Wu, Clemens Winter, Christopher Hesse, Mark Chen, Eric Sigler, Mateusz Litwin, Scott Gray, Benjamin Chess, Jack Clark, Christopher Berner, Sam McCandlish, Alec Radford, Ilya Sutskever, and Dario Amodei.
\newblock {Language Models are Few-Shot Learners}.
\newblock In Hugo Larochelle, Marc'Aurelio Ranzato, Raia Hadsell, Maria{-}Florina Balcan, and Hsuan{-}Tien Lin, editors, \emph{Advances in Neural Information Processing Systems 33: Annual Conference on Neural Information Processing Systems 2020, NeurIPS 2020, December 6-12, 2020, virtual}, 2020.
\newblock URL \url{https://proceedings.neurips.cc/paper/2020/hash/1457c0d6bfcb4967418bfb8ac142f64a-Abstract.html}.

\bibitem[OpenAI(2023)]{GPT4}
OpenAI.
\newblock {GPT-4 Technical Report}.
\newblock \emph{arXiv preprint arXiv:2303.08774}, 2023.
\newblock URL \url{https://arxiv.org/abs/2303.08774}.

\bibitem[Zheng et~al.(2023)Zheng, Zhou, Meng, Zhou, and Huang]{zheng2023large}
Chujie Zheng, Hao Zhou, Fandong Meng, Jie Zhou, and Minlie Huang.
\newblock Large language models are not robust multiple choice selectors.
\newblock In \emph{The Twelfth International Conference on Learning Representations}, 2023.

\bibitem[Kandpal et~al.(2023)Kandpal, Deng, Roberts, Wallace, and Raffel]{kandpal2023large}
Nikhil Kandpal, Haikang Deng, Adam Roberts, Eric Wallace, and Colin Raffel.
\newblock Large language models struggle to learn long-tail knowledge.
\newblock In \emph{International Conference on Machine Learning}, pages 15696--15707. PMLR, 2023.

\bibitem[Wei et~al.(2022{\natexlab{a}})Wei, Wang, Schuurmans, Bosma, Xia, Chi, Le, Zhou, et~al.]{wei2022chain}
Jason Wei, Xuezhi Wang, Dale Schuurmans, Maarten Bosma, Fei Xia, Ed~Chi, Quoc~V Le, Denny Zhou, et~al.
\newblock Chain-of-thought prompting elicits reasoning in large language models.
\newblock \emph{Advances in neural information processing systems}, 35:\penalty0 24824--24837, 2022{\natexlab{a}}.

\bibitem[Wei et~al.(2022{\natexlab{b}})Wei, Tay, Bommasani, Raffel, Zoph, Borgeaud, Yogatama, Bosma, Zhou, Metzler, et~al.]{wei2022emergent}
Jason Wei, Yi~Tay, Rishi Bommasani, Colin Raffel, Barret Zoph, Sebastian Borgeaud, Dani Yogatama, Maarten Bosma, Denny Zhou, Donald Metzler, et~al.
\newblock Emergent abilities of large language models.
\newblock \emph{Transactions on Machine Learning Research}, 2022{\natexlab{b}}.

\bibitem[Suzgun et~al.(2023{\natexlab{a}})Suzgun, Scales, Sch{\"a}rli, Gehrmann, Tay, Chung, Chowdhery, Le, Chi, Zhou, et~al.]{suzgun2023challenging}
Mirac Suzgun, Nathan Scales, Nathanael Sch{\"a}rli, Sebastian Gehrmann, Yi~Tay, Hyung~Won Chung, Aakanksha Chowdhery, Quoc Le, Ed~Chi, Denny Zhou, et~al.
\newblock Challenging big-bench tasks and whether chain-of-thought can solve them.
\newblock In \emph{Findings of the Association for Computational Linguistics: ACL 2023}, pages 13003--13051, 2023{\natexlab{a}}.

\bibitem[Valmeekam et~al.(2022)Valmeekam, Olmo, Sreedharan, and Kambhampati]{valmeekam2022large}
Karthik Valmeekam, Alberto Olmo, Sarath Sreedharan, and Subbarao Kambhampati.
\newblock Large language models still can't plan (a benchmark for llms on planning and reasoning about change).
\newblock In \emph{NeurIPS 2022 Foundation Models for Decision Making Workshop}, 2022.

\bibitem[Creswell et~al.(2022)Creswell, Shanahan, and Higgins]{creswell2022selection}
Antonia Creswell, Murray Shanahan, and Irina Higgins.
\newblock Selection-inference: Exploiting large language models for interpretable logical reasoning.
\newblock \emph{arXiv preprint arXiv:2205.09712}, 2022.

\bibitem[Yu et~al.(2020)Yu, Jiang, Dong, and Feng]{yu2020reclor}
Weihao Yu, Zihang Jiang, Yanfei Dong, and Jiashi Feng.
\newblock Reclor: A reading comprehension dataset requiring logical reasoning.
\newblock \emph{arXiv preprint arXiv:2002.04326}, 2020.

\bibitem[Liu et~al.(2021)Liu, Cui, Liu, Huang, Wang, and Zhang]{liu2021logiqa}
Jian Liu, Leyang Cui, Hanmeng Liu, Dandan Huang, Yile Wang, and Yue Zhang.
\newblock Logiqa: a challenge dataset for machine reading comprehension with logical reasoning.
\newblock In \emph{Proceedings of the Twenty-Ninth International Conference on International Joint Conferences on Artificial Intelligence}, pages 3622--3628, 2021.

\bibitem[Choi(2022)]{choi2022curious}
Yejin Choi.
\newblock The curious case of commonsense intelligence.
\newblock \emph{Daedalus}, 151\penalty0 (2):\penalty0 139--155, 2022.

\bibitem[Talmor et~al.(2019)Talmor, Herzig, Lourie, and Berant]{talmor2019commonsenseqa}
Alon Talmor, Jonathan Herzig, Nicholas Lourie, and Jonathan Berant.
\newblock Commonsenseqa: A question answering challenge targeting commonsense knowledge.
\newblock In \emph{Proceedings of the 2019 Conference of the North American Chapter of the Association for Computational Linguistics: Human Language Technologies, Volume 1 (Long and Short Papers)}, pages 4149--4158, 2019.

\bibitem[Talmor et~al.(2022)Talmor, Yoran, Bras, Bhagavatula, Goldberg, Choi, and Berant]{talmor2022commonsenseqa}
Alon Talmor, Ori Yoran, Ronan~Le Bras, Chandra Bhagavatula, Yoav Goldberg, Yejin Choi, and Jonathan Berant.
\newblock Commonsenseqa 2.0: Exposing the limits of ai through gamification.
\newblock \emph{arXiv preprint arXiv:2201.05320}, 2022.

\bibitem[Levesque et~al.(2012)Levesque, Davis, and Morgenstern]{levesque2012winograd}
Hector Levesque, Ernest Davis, and Leora Morgenstern.
\newblock The winograd schema challenge.
\newblock In \emph{Thirteenth international conference on the principles of knowledge representation and reasoning}, 2012.

\bibitem[Bhargava and Ng(2022)]{bhargava2022commonsense}
Prajjwal Bhargava and Vincent Ng.
\newblock Commonsense knowledge reasoning and generation with pre-trained language models: A survey.
\newblock In \emph{Proceedings of the AAAI Conference on Artificial Intelligence}, volume~36, pages 12317--12325, 2022.

\bibitem[Davis(2023)]{davis2023benchmarks}
Ernest Davis.
\newblock Benchmarks for automated commonsense reasoning: A survey.
\newblock \emph{ACM Computing Surveys}, 56\penalty0 (4):\penalty0 1--41, 2023.

\bibitem[Chung et~al.(2024)Chung, Hou, Longpre, Zoph, Tay, Fedus, Li, Wang, Dehghani, Brahma, et~al.]{chung2024scaling}
Hyung~Won Chung, Le~Hou, Shayne Longpre, Barret Zoph, Yi~Tay, William Fedus, Yunxuan Li, Xuezhi Wang, Mostafa Dehghani, Siddhartha Brahma, et~al.
\newblock Scaling instruction-finetuned language models.
\newblock \emph{Journal of Machine Learning Research}, 25\penalty0 (70):\penalty0 1--53, 2024.

\bibitem[Kojima et~al.(2022)Kojima, Gu, Reid, Matsuo, and Iwasawa]{kojima2022zero}
Takeshi Kojima, Shixiang~(Shane) Gu, Machel Reid, Yutaka Matsuo, and Yusuke Iwasawa.
\newblock Large language models are zero-shot reasoners.
\newblock In \emph{Advances in Neural Information Processing Systems}, volume~35, pages 22199--22213, 2022.

\bibitem[Zhou et~al.(2023)Zhou, Sch{\"a}rli, Hou, Wei, Scales, Wang, Schuurmans, Cui, Bousquet, Le, and Chi]{zhou2023leasttomost}
Denny Zhou, Nathanael Sch{\"a}rli, Le~Hou, Jason Wei, Nathan Scales, Xuezhi Wang, Dale Schuurmans, Claire Cui, Olivier Bousquet, Quoc~V Le, and Ed~H. Chi.
\newblock Least-to-most prompting enables complex reasoning in large language models.
\newblock In \emph{The Eleventh International Conference on Learning Representations}, 2023.
\newblock URL \url{https://openreview.net/forum?id=WZH7099tgfM}.

\bibitem[Yao et~al.(2023)Yao, Yu, Zhao, Shafran, Griffiths, Cao, and Narasimhan]{yao2023tree}
Shunyu Yao, Dian Yu, Jeffrey Zhao, Izhak Shafran, Thomas~L Griffiths, Yuan Cao, and Karthik Narasimhan.
\newblock Tree of thoughts: deliberate problem solving with large language models.
\newblock In \emph{Proceedings of the 37th International Conference on Neural Information Processing Systems}, pages 11809--11822, 2023.

\bibitem[Suzgun and Kalai(2024)]{suzgun2024meta}
Mirac Suzgun and Adam~Tauman Kalai.
\newblock Meta-prompting: Enhancing language models with task-agnostic scaffolding.
\newblock \emph{arXiv preprint arXiv:2401.12954}, 2024.

\bibitem[Wang et~al.(2022)Wang, Wei, Schuurmans, Le, Chi, and Zhou]{wang2022self}
Xuezhi Wang, Jason Wei, Dale Schuurmans, Quoc Le, Ed~Chi, and Denny Zhou.
\newblock {Self-Consistency Improves Chain of Thought Reasoning in Language Models}.
\newblock \emph{arXiv preprint arXiv:2203.11171}, 2022.
\newblock URL \url{https://arxiv.org/abs/2203.11171}.

\bibitem[Suzgun et~al.(2023{\natexlab{b}})Suzgun, Melas-Kyriazi, and Jurafsky]{suzgun2023follow}
Mirac Suzgun, Luke Melas-Kyriazi, and Dan Jurafsky.
\newblock Follow the wisdom of the crowd: Effective text generation via minimum bayes risk decoding.
\newblock In \emph{Findings of the Association for Computational Linguistics: ACL 2023}, pages 4265--4293, 2023{\natexlab{b}}.

\bibitem[Webson and Pavlick(2022)]{webson2022prompt}
Albert Webson and Ellie Pavlick.
\newblock Do prompt-based models really understand the meaning of their prompts?
\newblock In \emph{Proceedings of the 2022 Conference of the North American Chapter of the Association for Computational Linguistics: Human Language Technologies}, pages 2300--2344, 2022.

\bibitem[Webson et~al.(2023)Webson, Loo, Yu, and Pavlick]{webson2023language}
Albert Webson, Alyssa Loo, Qinan Yu, and Ellie Pavlick.
\newblock Are language models worse than humans at following prompts? it’s complicated.
\newblock In \emph{Findings of the Association for Computational Linguistics: EMNLP 2023}, pages 7662--7686, 2023.

\bibitem[Turpin et~al.(2024)Turpin, Michael, Perez, and Bowman]{turpin2024language}
Miles Turpin, Julian Michael, Ethan Perez, and Samuel Bowman.
\newblock Language models don't always say what they think: unfaithful explanations in chain-of-thought prompting.
\newblock \emph{Advances in Neural Information Processing Systems}, 36, 2024.

\bibitem[Baker et~al.(2011)Baker, Saxe, and Tenenbaum]{baker2011bayesian}
Chris Baker, Rebecca Saxe, and Joshua Tenenbaum.
\newblock Bayesian theory of mind: Modeling joint belief-desire attribution.
\newblock In \emph{Proceedings of the annual meeting of the cognitive science society}, volume~33, 2011.

\bibitem[Carlson et~al.(2013)Carlson, Koenig, and Harms]{carlson2013theory}
Stephanie~M Carlson, Melissa~A Koenig, and Madeline~B Harms.
\newblock Theory of mind.
\newblock \emph{Wiley Interdisciplinary Reviews: Cognitive Science}, 4\penalty0 (4):\penalty0 391--402, 2013.

\bibitem[Meltzoff(2002)]{meltzoff2002imitation}
Andrew~N Meltzoff.
\newblock Imitation as a mechanism of social cognition: Origins of empathy, theory of mind, and the representation of action.
\newblock \emph{Blackwell Handbook of Childhood Cognitive Development}, pages 6--25, 2002.

\bibitem[Shatz(1994)]{shatz1994theory}
Marilyn Shatz.
\newblock Theory of mind and the development of social-linguistic intelligence in early childhood.
\newblock In \emph{This chapter is based on a paper presented at the symposium" Early Theory of Mind Competencies," at the bienniel meeting of the Society for Research in Child Development, New Orleans, LA, Mar 1993.} Lawrence Erlbaum Associates, Inc, 1994.

\bibitem[Marcus and Davis(2023)]{marcusdavis2023}
Gary Marcus and Ernest Davis.
\newblock {How Not to Test GPT-3}.
\newblock 2023.
\newblock URL \url{https://cacm.acm.org/blogs/blog-cacm/270142-how-not-to-test-gpt-3/fulltext}.

\bibitem[Ma et~al.(2023{\natexlab{b}})Ma, Sansom, Peng, and Chai]{ma-etal-2023-towards-holistic}
Ziqiao Ma, Jacob Sansom, Run Peng, and Joyce Chai.
\newblock Towards a holistic landscape of situated theory of mind in large language models.
\newblock In Houda Bouamor, Juan Pino, and Kalika Bali, editors, \emph{Findings of the Association for Computational Linguistics: EMNLP 2023}, pages 1011--1031, Singapore, December 2023{\natexlab{b}}. Association for Computational Linguistics.
\newblock \doi{10.18653/v1/2023.findings-emnlp.72}.
\newblock URL \url{https://aclanthology.org/2023.findings-emnlp.72}.

\bibitem[Rakoczy(2022)]{rakoczy2022foundations}
Hannes Rakoczy.
\newblock Foundations of theory of mind and its development in early childhood.
\newblock \emph{Nature Reviews Psychology}, 1\penalty0 (4):\penalty0 223--235, 2022.

\bibitem[Pratt and Bryant(1990)]{pratt1990young}
Chris Pratt and Peter Bryant.
\newblock Young children understand that looking leads to knowing (so long as they are looking into a single barrel).
\newblock \emph{Child development}, 61\penalty0 (4):\penalty0 973--982, 1990.

\bibitem[Baron-Cohen and Goodhart(1994)]{baron1994seeing}
Simon Baron-Cohen and Frances Goodhart.
\newblock The ‘seeing-leads-to-knowing’deficit in autism: The pratt and bryant probe.
\newblock \emph{British Journal of Developmental Psychology}, 12\penalty0 (3):\penalty0 397--401, 1994.

\bibitem[Jung et~al.(2024)Jung, Kim, Jin, Kim, Seonwoo, Choi, Oh, and Kim]{jung2024perceptions}
Chani Jung, Dongkwan Kim, Jiho Jin, Jiseon Kim, Yeon Seonwoo, Yejin Choi, Alice Oh, and Hyunwoo Kim.
\newblock Perceptions to beliefs: Exploring precursory inferences for theory of mind in large language models.
\newblock \emph{arXiv preprint arXiv:2407.06004}, 2024.

\bibitem[Baron-Cohen et~al.(1985)Baron-Cohen, Leslie, and Frith]{baron1985does}
Simon Baron-Cohen, Alan~M Leslie, and Uta Frith.
\newblock Does the autistic child have a “theory of mind”?
\newblock \emph{Cognition}, 21\penalty0 (1):\penalty0 37--46, 1985.

\bibitem[Tollan and Palaz(2024)]{tollan2024does}
Rebecca Tollan and Bilge Palaz.
\newblock What does that mean? complementizers and epistemic authority.
\newblock \emph{Open Mind}, 8:\penalty0 366--394, 2024.

\end{thebibliography}
